\documentclass{article}

\usepackage[final]{neurips_2021}

\usepackage[utf8]{inputenc} 
\usepackage[T1]{fontenc}    
\usepackage{hyperref}       
\usepackage{url}            
\usepackage{booktabs}       
\usepackage{amsfonts}       
\usepackage{nicefrac}       
\usepackage{microtype}      
\usepackage{xcolor}         
\usepackage{amsmath}
\usepackage{amsthm}
\usepackage{amssymb}
\usepackage{tikz}
\usepackage{arydshln}
\usepackage{wrapfig}

\newtheorem{definition}{Definition}

\newtheorem{assumption}{Assumption}
\usepackage{multirow}
\newtheorem{theorem}{Theorem}

\newtheorem{lemma}{Lemma}
\usepackage{tikz}
\usetikzlibrary{positioning}
\usepackage{subfigure}
\usepackage{adjustbox}
\usepackage{booktabs}

\usepackage{amssymb}
\usepackage{pifont}
\newcommand{\cmark}{\ding{51}}%
\newcommand{\xmark}{\ding{55}}%
\usepackage{xcolor}
\def\boxit#1{%
  \smash{\color{blue}\fboxrule=0.25pt\relax\fboxsep=1pt\relax%
  \llap{\rlap{\fbox{\vphantom{0}\makebox[#1]{}}}~}}\ignorespaces
}

\usepackage{colortbl}
\definecolor{mygray}{gray}{.9}
\definecolor{mypink}{rgb}{.99,.91,.95}
\definecolor{mycyan}{cmyk}{.3,0,0,0}

\definecolor{best}{gray}{1}
\definecolor{second}{gray}{1}

\title{Invariance Principle Meets Information Bottleneck for Out-of-Distribution Generalization}

\author{%
  Kartik Ahuja$^{\dagger}$ \And Ethan Caballero\thanks{Equal contribution.} $^{}$ $^{ \dagger}$ \And Dinghuai Zhang$^{*}$ $^{\dagger}$ \And Jean-Christophe Gagnon-Audet $^{\dagger}$ \And 
 \And Yoshua Bengio $^{\dagger}$ \And  Ioannis Mitliagkas$^{\dagger}$ \And  Irina Rish\thanks{Mila - Quebec AI Institute, Université de Montréal. Correspondence to: kartik.ahuja@mila.quebec.}
}

\begin{document}
\maketitle

\begin{abstract}
The invariance principle from causality is at the heart of notable approaches such as invariant risk minimization (IRM) that seek to address out-of-distribution (OOD) generalization failures. Despite the promising theory, invariance principle-based approaches fail in common classification tasks,  where invariant (causal) features capture all the information about  the label.  Are these failures due to the methods failing to capture the invariance? Or is the invariance principle itself insufficient? To answer these questions, we revisit the fundamental assumptions in linear regression tasks, where invariance-based approaches were shown to provably generalize OOD. In contrast to the linear regression tasks, we show that for linear classification tasks we need much stronger restrictions on the distribution shifts, or otherwise OOD generalization is impossible.  Furthermore, even with appropriate restrictions on distribution shifts in place, we show that the invariance principle alone is insufficient. We prove that a form of the \textit{information bottleneck} constraint along with invariance  helps address  key failures when invariant features capture all the information about the label and also retains the existing success when they do not.  We propose an approach that incorporates both of these principles and demonstrate its effectiveness in several experiments.
\end{abstract}

\section{Introduction}
Recent years have witnessed an explosion of  examples showing deep learning models are prone to exploiting shortcuts (spurious features) \citep{geirhos2020shortcut,pezeshki2020gradient} which make them fail to generalize out-of-distribution (OOD). In  \cite{beery2018recognition}, a convolutional neural network  was trained to classify camels from cows; however, it was found that the model relied on the background color (e.g., green pastures for cows) and not on  the properties of the animals (e.g., shape). These examples become very concerning when they occur in real-life applications (e.g., COVID-19 detection  \citep{degrave2020ai}). 

To address these out-of-distribution generalization failures, invariant risk minimization \citep{arjovsky2019invariant} and several other works were proposed  \citep{ahuja2020invariant,  pezeshki2020gradient, krueger2020out, robey2021model,Zhang2021CanSS}. The invariance principle from causality \citep{peters2015causal,pearl1995causal} is at the heart of these works. The principle distinguishes predictors that only rely on the causes of the label from those that do not. The optimal predictor that only focuses on the causes is invariant and min-max optimal \citep{rojas2018invariant,  koyama2020out,ahuja2020empirical} under many distribution shifts but the same is not true for other predictors.

\textbf{Our contributions.} Despite the promising theory,  invariance principle-based approaches fail in settings \citep{aubin2021linear} where  invariant features capture all information about the label contained in the input.
A particular example is image classification  (e.g., cow vs. camel) \citep{beery2018recognition}  where the label is a deterministic function of the invariant features (e.g., shape of the animal), and does not depend on the spurious features (e.g., background).
To understand such failures, we revisit the fundamental assumptions in linear regression tasks, where invariance-based approaches were shown to provably generalize OOD. We show that, in contrast to the linear regression tasks, OOD generalization is significantly harder for linear classification tasks; we need much stronger restrictions in the form of support overlap assumptions\footnote{Support is the region where the probability density for continuous random variables (probability mass function for discrete random variables) is positive. Support overlap refers to the setting where train and test distribution maybe different but share the same support. We formally define this later in Assumption \ref{assumption 6_new}.} on the distribution shifts, or otherwise it is not possible to guarantee OOD generalization under interventions on variables other than the target class.
We then proceed to show that, even under the right assumptions on distribution shifts, the invariance principle is insufficient. However, we establish that \textit{information bottleneck} (IB) constraints \citep{tishby2000information}, together with the invariance principle, provably works in both settings -- when invariant features completely capture the information about the label and also when they do not.  (Table \ref{table1_summary} summarizes our theoretical results presented later).  We propose an approach that combines both these principles and demonstrate its effectiveness  on linear unit tests \citep{aubin2021linear} and on different real datasets.

\begin{table}[h]
\label{table1_summary}
\begin{center}
\adjustbox{max width=\textwidth}{%
\begin{tabular}{cccccccc}
\toprule
Task                & Invariant features             & Support   overlap       & Support overlap        &    \multicolumn{4}{ c }{\textbf{OOD generalization guarantee} $(\mathcal{E}_{tr}\rightarrow \mathcal{E}_{all}$)}      \\
                    &  capture label info             &  invariant  features               &    spurious features            &  ERM  & IRM & IB-ERM & IB-IRM  \\ 
\midrule

\multirow{3}{5em}{Linear Classification}                  &  Full/Partial            &  No              &    Yes/No       &   \multicolumn{4}{ c }{Impossible for any algorithm to generalize OOD [Thm2]} \\ 
                                                   &  Full                &  Yes             &    No            & \boxit{2.3in}\xmark  & \xmark & \cmark &  \;\;\;\;\;\;\;\;\;\;\;\;\;\;\;\;\cmark\;\; [Thm3,4] \\ 
                                                   
 & Partial                                                                    & Yes              & No               & \xmark  & \xmark  & \xmark & \;\;\;\;\;\; \;\;\;\;\;\;\;\;\;\;\;\;\cmark\;\;[Appendix] \\
 
 &  Full                                                                  &  Yes             &    Yes           & \boxit{2.3in}\cmark  & \cmark  & \cmark &\;\;\;\;\;\;\;\;\;\;\;\;\;\;\;\cmark\;\;[Thm3,4]\\
 &  Partial                                                                   &  Yes             &    Yes           & \xmark  & \cmark  & \xmark & \cmark\\
 \hline
 \multirow{2}{5em}{Linear Regression}                  &  Full                   &  No              &    No            & \cmark  & \cmark & \cmark & \cmark \\ 
 &  Partial                                                                   &  No              &    No          & \boxit{2.3in}\xmark  & \cmark & \xmark & \;\;\;\;\;\;\;\;\;\;\;\;\;\;\cmark\;\;\;[Thm4] \\ 
\bottomrule
\end{tabular}}
\end{center}
\caption{ Summary of the new and existing results \citep{arjovsky2019invariant, rosenfeld2020risks}. IB-ERM (IRM): information bottleneck - empirical (invariant) risk minimization ERM (IRM).} 
\end{table}

\vspace{-0.5em}
\section{OOD generalization and invariance: background \& failures}
 \vspace{-0.5em}
\noindent{\bf Background.} 
 We consider a supervised training data $D$ gathered from a set of training  environments $\mathcal{E}_{tr}$:  $D = \{D^e\}_{e\in \mathcal{E}_{tr}}$, where $D^e=\{x_{i}^e, y_{i}^e\}_{i=1}^{n^e}$ is the dataset from environment $e\in \mathcal{E}_{tr}$ and $n^e$ is the number of instances in environment $e$.  $x^e_{i} \in  \mathbb{R}^{d}$ and $y^e_{i}\in \mathcal{Y} \subseteq \mathbb{R}^k$ correspond to the input feature value  and the  label for $i^{th}$ instance respectively. Each $(x^e_{i},y^e_{i})$ is an i.i.d. draw from $\mathbb{P}^e$, where $\mathbb{P}^e$ is the joint distribution of the input feature and the label in environment $e$. Let $\mathcal{X}^e$ be the support of the input feature values in the environment $e$.   The goal of OOD generalization is to use training data $D$ to construct a predictor  $f: \mathbb{R}^{d} \rightarrow \mathbb{R}^k$ that performs well across many unseen environments in $\mathcal{E}_{all}$, where $\mathcal{E}_{all} \supset \mathcal{E}_{tr}$. Define the risk of $f$ in environment $e$ as $R^e(f) = \mathbb{E}\big[\ell(f(X^e), Y^e) \big]$, where for example $\ell$ can be $0$-$1$ loss, logistic loss, square loss, $(X^e, Y^e) \sim \mathbb{P}^e$, and the expectation $\mathbb{E}$ is w.r.t. $\mathbb{P}^e$.
 Formally stated, our goal is to use the data from training environments $\mathcal{E}_{tr}$  to find $f: \mathbb{R}^{d}  \rightarrow \mathcal{Y}$ to minimize
 \vspace{-0.05in}
 \begin{equation}
     \min_{f} \max_{e \in \mathcal{E}_{all}} R^e(f).
     \label{eqn1: min_max_ood}
 \end{equation}
So far we did not state any restrictions on $\mathcal{E}_{all}$. Consider binary classification: without any restrictions on $\mathcal{E}_{all}$, no method can reduce the above objective ($\ell$ is $0$-$1$ loss) to below one. Suppose a method outputs $f^{*}$; if  $\exists \; e\in \mathcal{E}_{all} \setminus  \mathcal{E}_{tr}$ with labels based on $1-f^{*}$, then it achieves an error of one. Some assumptions on $\mathcal{E}_{all}$ are thus necessary. Consider how  $\mathcal{E}_{all}$ is restricted using invariance for linear regressions   \citep{arjovsky2019invariant}. 
\begin{assumption}
\label{assumption 1_new} \textbf{Linear regression structural equation model (SEM).} In each $ e \in \mathcal{E}_{all}$ 
\begin{equation}
    \begin{split}
    & Y^e \leftarrow  w_{\mathsf{inv}}^{*}\cdot Z_{\mathsf{inv}}^e+ \epsilon^e,  \;\;\;\; Z_{\mathsf{inv}}^{e} \perp \epsilon^e, \;\;\;\;\mathbb{E}[\epsilon^e]=0, \mathbb{E}\big[|\epsilon^e|^2\big] \leq \sigma_{\mathsf{sup}}^2 \\ 
    & X^e \leftarrow S(Z_{\mathsf{inv}}^e, Z_{\mathsf{spu}}^e)
    \end{split}
\end{equation}
where $w_{\mathsf{inv}}^{*}\in \mathbb{R}^m$, $Z_{\mathsf{inv}}^{e}\in \mathbb{R}^m$, $Z_{\mathsf{spu}}\in \mathbb{R}^{o}$, $S\in \mathbb{R}^{d \times (m+o) }$, $S$ is invertible ($m+o=d$). We focus on invertible $S$ but several results extend to non-invertible $S$ as well (see Appendix).
\end{assumption}
Assumption \ref{assumption 1_new} states  how $Y^e$ and $X^e$ are generated from latent invariant features $Z_{\mathsf{inv}}^{e}$\footnote{In many examples in the literature, invariant features are causal, but not always \citep{rosenfeld2020risks}.}, latent spurious features $Z_{\mathsf{spu}}^{e}$  and noise $\epsilon^e$. The \textit{relationship between label and invariant features is invariant, i.e., $w_{\mathsf{inv}}^{*}$ is fixed} across all environments. However, the distributions of   $Z_{\mathsf{inv}}^{e}$, $Z_{\mathsf{spu}}^{e}$, and $\epsilon^e$ are allowed to change arbitrarily across all the environments. Suppose $S$ is identity. If we regress only on the invariant features $Z^{e}_{\mathsf{inv}}$, then the optimal solution is $w_{\mathsf{inv}}^{*}$, which is independent of the environment, and the error it achieves is bounded above by the variance of $\epsilon^e$ ($\sigma_{\mathsf{sup}}^2$). If we regress on the entire $Z^e$ and the optimal predictor places a non-zero weight on $Z_{\mathsf{spu}}^e$ (e.g.,  $Z_{\mathsf{spu}}^e\leftarrow Y^e + \zeta^e$), then this predictor fails to solve equation \eqref{eqn1: min_max_ood} ($\exists\; e\in \mathcal{E}_{all}$, $Z_{\mathsf{spu}}^e \rightarrow \infty$, $\text{error}\rightarrow   \infty$, see Appendix for details). Also, not only regressing on $Z^{e}_{\mathsf{inv}}$ is better than on $Z^e$, it can be shown that it is optimal, i.e., it solves equation \eqref{eqn1: min_max_ood} under Assumption \ref{assumption 1_new} and achieves a value of $\sigma_{\mathsf{sup}}^2$ for the objective in equation \eqref{eqn1: min_max_ood}.

\textbf{Invariant predictor.} Define a linear representation  map $\Phi:\mathbb{R}^{r \times d}$ (that transforms $X^e$ as $\Phi(X^e)$) and define a linear classifier $w:\mathbb{R}^{k\times r}$ (that operates on the representation $w\cdot \Phi(X^e))$.  
We want to search for representations $\Phi$ such that $\mathbb{E}[Y^e|\Phi(X^e)]$ is invariant (in Assumption \ref{assumption 1_new} if $\Phi(X^e)=Z_{\mathsf{inv}}^e$, then $\mathbb{E}[Y^e|\Phi(X^e)]$ is invariant).  We say that a data representation $\Phi$ elicits an invariant predictor $w \cdot \Phi$ across the set of training environments $\mathcal{E}_{tr}$ if there is a predictor $w$ that simultaneously achieves the minimum risk, i.e.,
$w \in \arg\min_{\tilde{w}} R^{e}(\tilde{w}\cdot \Phi), \; \forall e \in \mathcal{E}_{tr}$.  The main objective of IRM is stated as 
\begin{equation}
        \min_{w \in \mathbb{R}^{k \times r}, \Phi \in \mathbb{R}^{r \times d}} \frac{1}{|\mathcal{E}_{tr}|}\sum_{e \in \mathcal{E}_{tr}}R^{e}(w \cdot \Phi) 
   \quad  \text{s.t.}\;w \in \arg\min_{\tilde{w}\in \mathbb{R}^{k \times r}} R^{e}(\tilde{w} \cdot \Phi),\;\forall e \in \mathcal{E}_{tr}.
    \label{eqn: IRM}
\end{equation}
Observe that if we drop the constraints in the above which search only over invariant predictors, then we get the standard empirical risk minimization (ERM) \citep{vapnik1992principles} (assuming all the training environments occur with equal probability). In all our theorems, we use $0$-$1$ loss for binary classification $\mathcal{Y}= \{0,1\}$ and square loss for regression $\mathcal{Y}= \mathbb{R}$. For binary classification, the output of the predictor is given as $\mathsf{I}(w\cdot \Phi(X^e))$, where $\mathsf{I}(\cdot)$ is the indicator function that takes $1$ if the input is $\geq 0$ and $0$ otherwise, and the risk is $R^{e}(w\cdot \Phi) = \mathbb{E}\big[|\mathsf{I}(w\cdot\Phi (X^e)) - Y^e|\big]$. For regression, the output of the predictor is $w\cdot \Phi (X^e)$ and the corresponding risk is $R^{e}(w\cdot \Phi) = \mathbb{E}\big[(w\cdot\Phi(X^e) - Y^e)^2\big]$.  We now present  the main OOD generalization result from \cite{arjovsky2019invariant} for linear regressions.

\begin{theorem}
\label{thm9: arjovsky} (Informal) If Assumption \ref{assumption 1_new} is satisfied, $\mathsf{Rank}[\Phi]>0$, $|\mathcal{E}_{tr}|>2d$, and $\mathcal{E}_{tr}$ lie in a linear general position (a mild condition on the data in $\mathcal{E}_{tr}$, defined in the Appendix), then each solution to equation \eqref{eqn: IRM}  achieves OOD generalization (solves equation \eqref{eqn1: min_max_ood}, $\nexists \; e \in \mathcal{E}_{all}$ with risk $>\sigma_{\mathsf{sup}}^2$). 
\end{theorem}

Despite the above guarantees, IRM has been shown to fail in several cases including linear SEMs in \citep{aubin2021linear}. We take a closer look at these failures next.

\noindent{\bf Understanding the failures: fully informative invariant features vs. partially informative invariant features (FIIF vs. PIIF).} We define properties salient to the datasets/SEMs used in the  OOD generalization literature. Each $e\in \mathcal{E}_{all}$, the distribution $(X^e,Y^e)\sim \mathbb{P}^e$  satisfies the following properties. a) $\exists$ a map $\Phi^{*}$ (linear or not), which we call an \textit{invariant feature map}, such that $\mathbb{E}\big[Y^e\big|\Phi^{*}\big(X^e\big)\big]$ is the same for all $e\in \mathcal{E}_{all}$ and $Y^e \not\perp \Phi^{*}(X^e)$. These conditions ensure $\Phi^{*}$ maps to features that have a finite predictive power and have the same optimal predictor across $\mathcal{E}_{all}$. For the SEM in Assumption \ref{assumption 1_new}, $\Phi^{*}$ maps to $Z_{\mathsf{inv}}^e$. b) $\exists$ a map  $\Psi^{*}$ (linear or not), which we call \textit{spurious feature map}, such that $\mathbb{E}\big[Y^e\big|\Psi^{*}\big(X^e\big)\big]$ is not the same for all $e\in \mathcal{E}_{all}$ and $Y^e \not\perp \Psi^{*}(X^e)$ for some environments.  $\Psi^{*}$ often creates a hindrance in learning predictors that only rely on $\Phi^{*}$. Note that $\Psi^{*}$ should not be a transformation of some $\Phi^{*}$. For the SEM in Assumption \ref{assumption 1_new}, suppose $Z_{\mathsf{spu}}^e$ is anti-causally related to $Y^e$, then $\Psi^{*}$ maps to $Z_{\mathsf{spu}}^e$ (See Appendix for an example). 

In the colored MNIST (CMNIST) dataset \citep{arjovsky2019invariant}, the digits are colored in such a way that in the training domain, color is highly predictive of the digit label but this correlation being spurious breaks down at test time. Suppose the invariant feature map $\Phi^{*}$ extracts the uncolored digit and the spurious feature map $\Psi^{*}$ extracts the background color.  \cite{ahuja2020empirical} studied two variations of the colored MNIST  dataset, which differed in the way final labels are generated from original MNIST labels (corrupted with noise or not). They showed that the IRM exhibits good OOD generalization ($50 \% $ improvement over ERM) in anti-causal-CMNIST (AC-CMNIST, original data from \cite{arjovsky2019invariant})  but is no different from ERM and fails in covariate shift-CMNIST (CS-CMNIST).
In AC-CMNIST, the invariant features $\Phi^{*}(X^e)$ (uncolored digit) are \textit{partially informative} about the label, i.e., $Y \not \perp X^e | \Phi^{*}(X^e)$, and color contains information about label not contained in the uncolored digit. On the other hand in CS-CMNIST, invariant features are \textit{fully informative} about the label, i.e., $Y  \perp X^e | \Phi^{*}(X^e)$, i.e., they contains all the information about the label that is contained in input $X^e$. Most human labelled  datasets have fully informative invariant features; the labels (digit value) only depend on the invariant features (uncolored digit) and spurious features (color of the digit) do not affect the label. \footnote{The deterministic labelling case was referred as realizable problems in \citep{arjovsky2019invariant}.}  In the rare case, when the humans are asked to label images in which the object being labelled itself is blurred, humans can rely on spurious features such as the background making such a data representative of PIIF setting. In Table \ref{table1}, we divide the different datasets used in the literature based on informativeness of the invariant features. We observe that when the invariant features are fully informative, both IRM and ERM fail but only in classification tasks and not in regression tasks \citep{ahuja2020empirical}; this is consistent with the linear regression result in Theorem \ref{thm9: arjovsky}, where IRM succeeds regardless of whether $Y^{e}\perp X^{e}|Z_{\mathsf{inv}}^e$ holds or not. Motivated by this observation, we take a closer look at the classification tasks where invariant features are fully informative.


\begin{table}
 \renewcommand{\arraystretch}{1.25}
\begin{center}
 \adjustbox{max width=\textwidth}{\begin{tabular}{||l | l |} 
 \hline
 \textbf{Fully informative invariant features (FIIF) }   &  \textbf{Partially informative invariant features (PIIF) }  \\
 $\forall e \in \mathcal{E}_{all}, Y^e \perp X^e |  \Phi^{*}(X^e)$ &  $\exists \;e \in \mathcal{E}_{all} \;Y^e \not\perp X^e | \Phi^{*}(X^e)$\\
 \hline
 \textbf{Task: classification} & \textbf{Task: classification or regression} \\
 Example 2/2S, CS-CMNIST    & Example 1/1S, Example 3/3S, AC-CMNIST  \\     
SEM in Assumption \ref{assumption 3_new}  & SEM in  \cite{rosenfeld2020risks}  \\
   \textbf{ERM and IRM fail} & \textbf{ERM fails, IRM succeeds sometimes  } \\ 
  Theorem 3,4 (This paper) & Theorem 9, 5.1 \citep{arjovsky2019invariant, rosenfeld2020risks} \\
   \hline
\end{tabular}}
\end{center}
\caption{Categorization of OOD evaluation datasets and SEMs. Example 1/1S, 2/2S, 3/3S from \citep{aubin2021linear}, AC-CMNIST\citep{arjovsky2019invariant}, CS-CMNIST\citep{ahuja2020empirical}.
}
\label{table1}
\end{table}
\vspace{-0.05in}
\section{OOD generalization theory for linear classification tasks}
\begin{figure}
    \centering
    \includegraphics[trim=0 2in 0 0, width=2.5in]{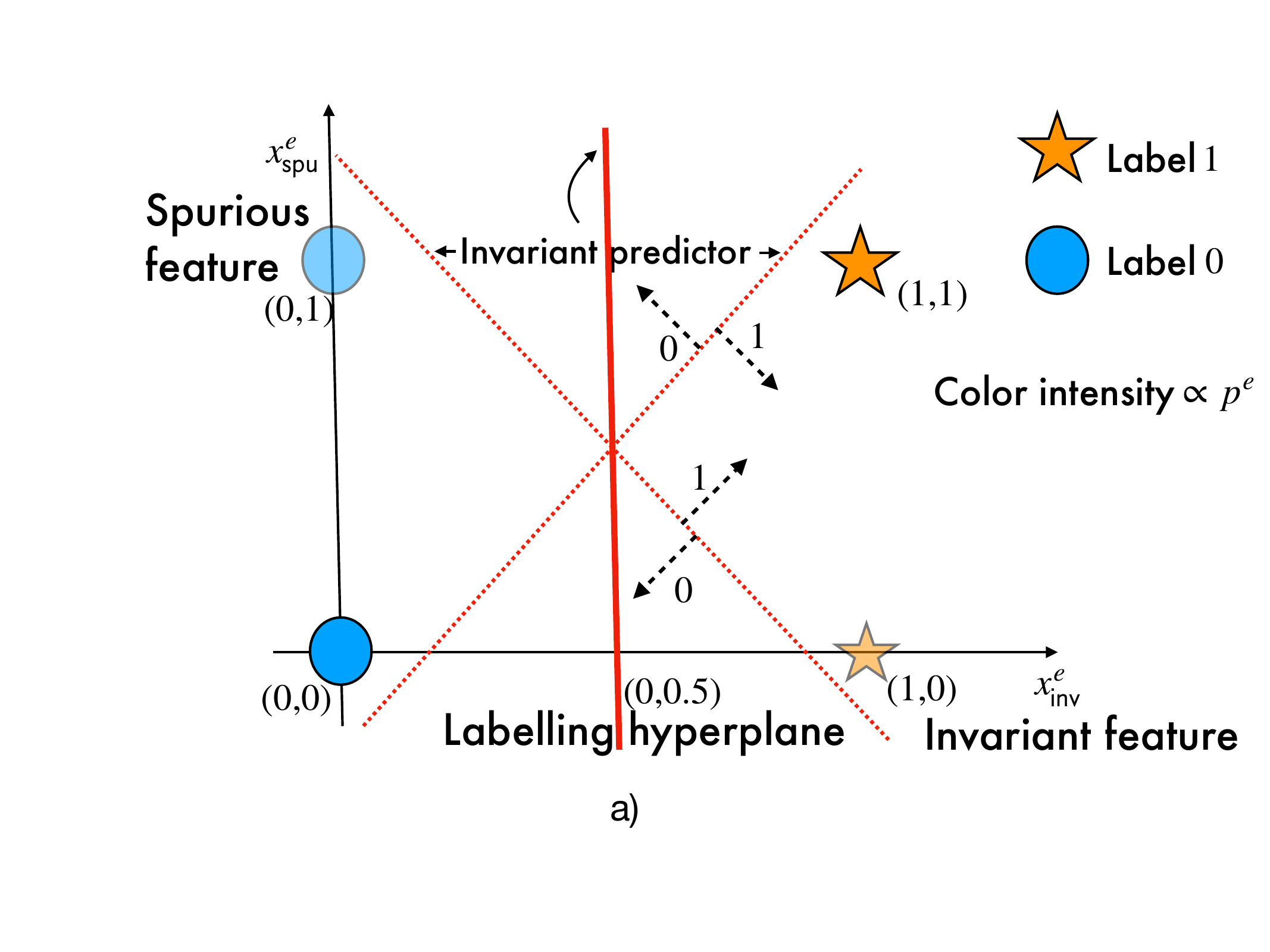}      \includegraphics[trim=0 2in 0 0, width=2.5in]{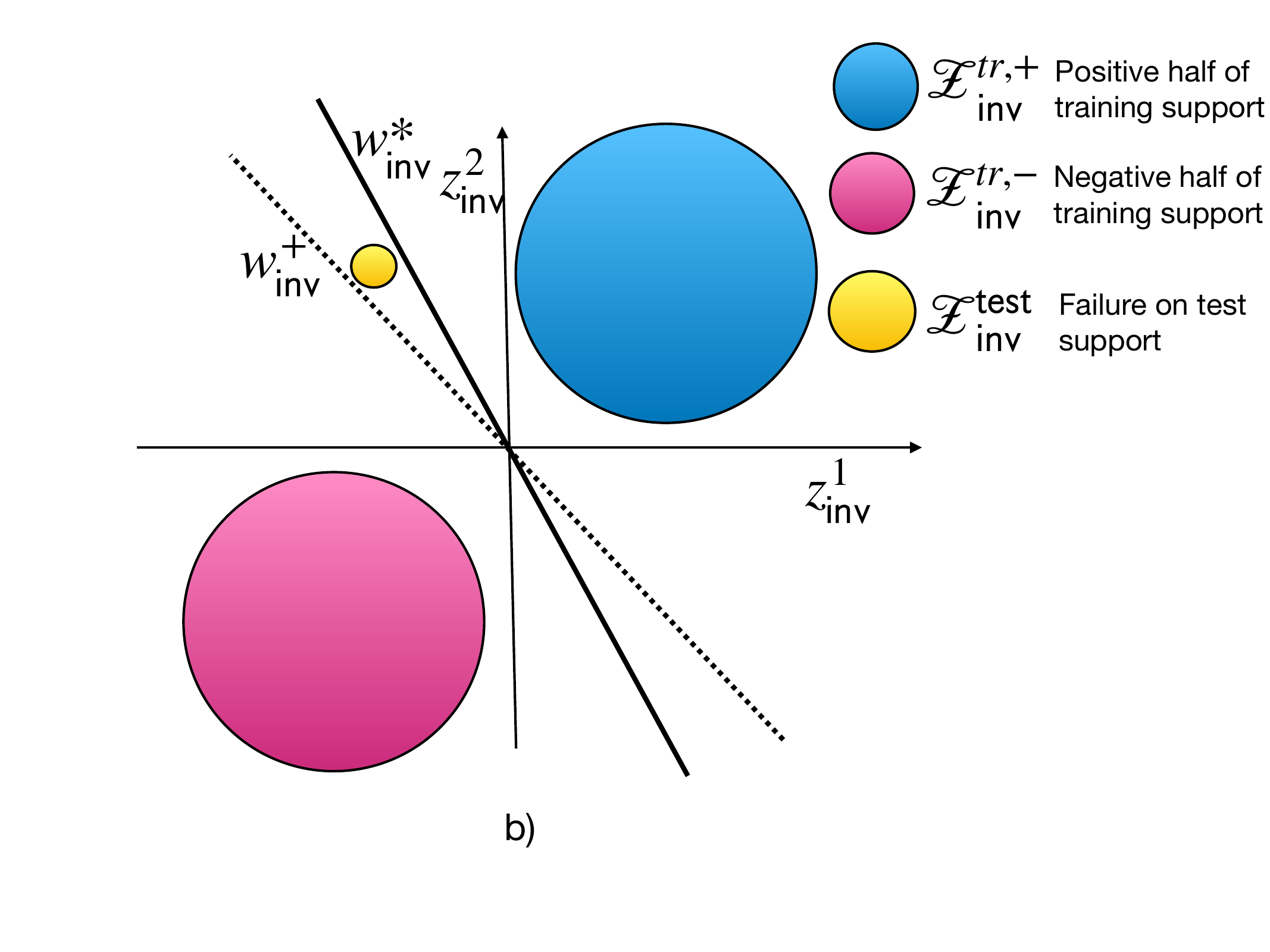}
    \caption{ a) 2D classification example illustrating multiple invariant predictors: Most of these predictors rely on spurious features and each of them achieve zero error across all $\mathcal{E}_{tr}$, b) illustration of the impossibility result. If latent invariant features in the training environments are separable, then there are multiple equally good candidates that could have generated the data, and the algorithm cannot distinguish between these.}
    \label{fig:2d_impossible}
\end{figure}

\vspace{-0.05in}
\textbf{A two-dimensional example with fully informative invariant features.}  We start with a 2D classification example (based on \cite{nagarajan2020understanding}), which can be understood as a simplified version of the CS-CMNIST dataset \citep{ahuja2020empirical}, Example 2/2S of \citet{aubin2021linear}, where both IRM and ERM fail. The example goes as follows. In each training environment $e \in \mathcal{E}_{tr}$
\begin{equation}
    \begin{split}
    &  Y^e \leftarrow \mathsf{I}\Big(X_{\mathsf{inv}}^e-\frac{1}{2}\Big), \; \text{where}\; X_{\mathsf{inv}}^e \in\{0,1\}\; \text{is}\; \mathsf{Bernoulli}\Big(\frac{1}{2}\Big), \\
    & X_{\mathsf{spu}}^e \leftarrow X_{\mathsf{inv}}^e\oplus W^e,\; \text{where}\; W^e \in \{0,1\} \; \text{is} \; \mathsf{Bernoulli}\big(1-p^e\big)\; \text{with selection bias} \; p^{e}>\frac{1}{2},
    \end{split}
    \label{eqn:2d_toy_example}
\end{equation}
where $\mathsf{Bernoulli}(a)$ takes value $1$ with probability $a$ and $0$ otherwise. Each training environment is characterized by the probability $p^e$. Following   Assumption \ref{assumption 1_new}, we assume that the labelling function does not change from $\mathcal{E}_{tr}$ to $\mathcal{E}_{all}$, thus the relation between the label and the invariant features does not change. Assume that the distribution of $X_{\mathsf{inv}}^{e}$ and $X_{\mathsf{spu}}^{e}$ can change arbitrarily.  See Figure \ref{fig:2d_impossible}a) for a pictorial representation of this example illustrating the gist of the problem: there are many classifiers with the same error on $\mathcal{E}_{tr}$ while only the one identical to the labelling function $\mathsf{I}(X_{\mathsf{inv}}^e-\frac{1}{2})$ generalizes correctly OOD. Define a classifier $\mathsf{I}(w_{\mathsf{inv}}x_{\mathsf{inv}} + w_{\mathsf{spu}}x_{\mathsf{spu}}-\frac{1}{2}(w_{\mathsf{inv}} + w_{\mathsf{spu}}))$. Define a set of classifiers $\mathcal{S} = \{(w_{\mathsf{inv}}, w_{\mathsf{spu}})\;\text{s.t.}\;w_{\mathsf{inv}} >|w_{\mathsf{spu}}| \}$. Observe that all the classifiers in $\mathcal{S}$ achieve a zero classification error on the training environments. However, only classifiers for which $w_{\mathsf{spu}}=0$ solve the OOD generalization (eq. \eqref{eqn1: min_max_ood}).  With $\Phi$ as the identity, it can be shown that all the classifiers $\mathcal{S}$ form an invariant predictor (satisfy the constraint in equation \eqref{eqn: IRM} over all the training environments when $\ell$ is the $0$-$1$ loss).  Observe that increasing the number of training environments to infinity does not address the problem, unlike with the linear regression result discussed 
in Theorem \ref{thm9: arjovsky} \citep{arjovsky2019invariant}, where it was shown that if the number of environments increases linearly in the dimension of the data, then the solution to IRM also solves the OOD generalization (eq. \eqref{eqn1: min_max_ood}). \footnote{Please note that this example illustrates certain important facets in a very simple fashion; only in this example a max-margin classifier can solve the problem but not in general. (Further explanation in the Appendix).} We use the above example to construct general SEMs for linear classification when the invariant features are fully informative. We follow the structure of the SEM from Assumption \ref{assumption 1_new} in our construction.

\begin{assumption}
\label{assumption 3_new} \textbf{Linear classification structural equation model (FIIF).} In each $e\in \mathcal{E}_{all}$ 
\begin{equation}
\begin{split}
   & Y^e \leftarrow \mathsf{I}\big(w_{\mathsf{inv}}^{*} \cdot Z_{\mathsf{inv}}^e\big) \oplus N^e, \;\;\;\;\ N^e \sim \mathsf{Bernoulli}(q), q<\frac{1}{2},\;\;\;\;\; N^{e} \perp (Z_{\mathsf{inv}}^{e},    Z_{\mathsf{spu}}^{e}),\\
  &  X^{e} \leftarrow S\big(Z_{\mathsf{inv}}^{e}, Z_{\mathsf{spu}}^{e}\big),
\end{split}
\end{equation}
where $w_{\mathsf{inv}}^{*}\in \mathbb{R}^{m}$ with $\|w_{\mathsf{inv}}^{*}\|=1$ is the labelling hyperplane, $Z_{\mathsf{inv}}^e\in \mathbb{R}^m$, $Z_{\mathsf{spu}}^e\in \mathbb{R}^o$, $N^e$ is binary noise with identical distribution across environments, $\oplus$ is the XOR operator, $S$ is invertible. 
\end{assumption}
If noise level $q$ is zero, then the above SEM covers linearly separable problems. See Figure \ref{figure:dag_comparison}a) for the directed acyclic graph (DAG) corresponding to this SEM. From the DAG observe that  $Y^{e} \perp  X^{e}|Z_{\mathsf{inv}}^{e}$, which implies that the invariant features are fully informative.  Contrast this with a DAG that follows Assumption \ref{assumption 1_new} shown in Figure \ref{figure:dag_comparison}b), where $Y^{e} \not \perp  X^{e}|Z_{\mathsf{inv}}^{e}$ and thus the invariant features are not fully informative. If $\mathcal{E}_{all}$ follows the SEM in Assumption \ref{assumption 3_new} and suppose the distribution of $Z_{\mathsf{inv}}^e$, $Z_{\mathsf{spu}}^e$ can change arbitrarily, then it can be shown that only a classifier identical to the labelling function $\mathsf{I}(w_{\mathsf{inv}}^{*}\cdot Z_{\mathsf{inv}}^e)$ can solve the OOD generalization (eq. \eqref{eqn1: min_max_ood}); such a classifier achieves an error of $q$ (noise level) in all the environments. As a result, if for a classifier we can find $e\in \mathcal{E}_{all}$ that follows Assumption \ref{assumption 3_new} where the error is greater than $q$, then such a classifier does not solve equation \eqref{eqn1: min_max_ood}. Now we ask -- what are the minimal conditions on training environments $\mathcal{E}_{tr}$ to achieve OOD generalization when $\mathcal{E}_{all}$ follow Assumption \ref{assumption 3_new}? %
 To achieve OOD generalization for linear regressions,  in Theorem \ref{thm9: arjovsky}, it was required that the number of training environments grows linearly in the dimension of the data.  However, there was no restriction on the support of the latent invariant and latent spurious features, and they were allowed to change arbitrarily from train to test (for further discussion on this, see the Appendix). Can we continue to work with similar assumptions for the  SEM in Assumption \ref{assumption 3_new} and solve the OOD generalization (eq. \eqref{eqn1: min_max_ood})? We state some assumptions and notations to answer that. Define the support of the invariant (spurious) features $Z_{\mathsf{inv}}^{e}$ ($Z_{\mathsf{spu}}^{e}$) in environment $e$ as  $\mathcal{Z}^{e}_{\mathsf{inv}}$ ($\mathcal{Z}^{e}_{\mathsf{spu}}$), and the support of the joint distribution over invariant and spurious feature $Z=[Z_{\mathsf{inv}}^{e}, Z_{\mathsf{spu}}^{e}]$ in environment $e$ as $\mathcal{Z}^{e}$. 

\begin{assumption}
\label{assumption 4_new} \textbf{Bounded invariant features.}
$ \cup_{e \in \mathcal{E}_{tr}} \mathcal{Z}^{e}_{\mathsf{inv}}$  is a bounded set.\footnote{A set $\mathcal{Z}$ is bounded if $\exists M<\infty$ such that $\forall z \in \mathcal{Z}, \|z\|\leq M$.}
\end{assumption}

\begin{assumption}
\label{assumption 5_new} \textbf{Bounded spurious features.}
$ \cup_{e \in \mathcal{E}_{tr}} \mathcal{Z}^{e}_{\mathsf{spu}}$  is a bounded set. 
\end{assumption}


\begin{assumption}
\label{assumption 6_new}
\textbf{Invariant feature support overlap.}
$\forall e \in \mathcal{E}_{all}, \mathcal{Z}^{e}_{\mathsf{inv}} \subseteq \cup_{e'\in \mathcal{E}_{tr}}\mathcal{Z}_{\mathsf{inv}}^{e'}$
\end{assumption}

\begin{assumption}
\label{assumption 7_new}
\textbf{Spurious feature support overlap.}
$\forall e \in \mathcal{E}_{all}, \mathcal{Z}^{e}_{\mathsf{spu}} \subseteq \cup_{e'\in \mathcal{E}_{tr}}\mathcal{Z}_{\mathsf{spu}}^{e'}$
\end{assumption}

\begin{assumption}
\label{assumption joint_new}
\textbf{Joint feature support overlap.}
$\forall e \in \mathcal{E}_{all}, \mathcal{Z}^{e} \subseteq \cup_{e'\in \mathcal{E}_{tr}}\mathcal{Z}^{e'}$
\end{assumption}

Assumption \ref{assumption 6_new} (\ref{assumption 7_new}) states that the support of the invariant (spurious) features for unseen environments is the same as the union of the support over the training environments. It is important to note that support overlap does not imply that the distribution over the invariant features does not change.  We now define a margin that measures how much the is training support of invariant features $Z_{\mathsf{inv}}^e$ separated by the labelling hyperplane $w_{\mathsf{inv}}^{*}$.  Define $\mathsf{Inv}$-$\mathsf{Margin} = \min_{z \in \cup_{e\in \mathcal{E}_{tr}} \mathcal{Z}_{\mathsf{inv}}^{e}} \mathsf{sgn}\big(w_{\mathsf{inv}}^{*} \cdot z\big) \big(w_{\mathsf{inv}}^{*} \cdot z\big)$. This margin only coincides with the standard margin in support vector machines  when the noise level $q$ is 0 (linearly separable) and $S$ is identity. If $\mathsf{Inv}$-$\mathsf{Margin}>0$, then the labelling hyperplane $w_{\mathsf{inv}}^{*}$ separates the support into two halves (see Figure \ref{fig:2d_impossible}b)). 
\begin{assumption} 
\label{assumption 8_new}
\textbf{Strictly separable invariant features.} $\mathsf{Inv}$-$\mathsf{Margin}>0$. 
\end{assumption}

Next, we show the importance of support overlap for invariant features.

\begin{theorem}
\label{theorem 2}
\textbf{Impossibility of guaranteed OOD generalization for linear classification.} Suppose  each $e \in \mathcal{E}_{all}$ follows Assumption \ref{assumption 3_new}. If for all the training environments $\mathcal{E}_{tr}$, the latent invariant features are bounded and strictly separable, i.e., Assumption \ref{assumption 4_new} and \ref{assumption 8_new} hold, then  every deterministic algorithm fails to solve the OOD generalization (eq. \eqref{eqn1: min_max_ood}), i.e.,  for the output of every algorithm $\exists \; e\in \mathcal{E}_{all}$  in which the error exceeds the minimum required value $q$ (noise level).
\end{theorem}
The proofs to all the theorems are in the Appendix.  We provide a high-level intuiton as to why invariant feature support overlap is crucial to the impossibility result.  In Figure  \ref{fig:2d_impossible}b), we show that if the support of latent invariant features are strictly separated by the labelling hyperplane $w_{\mathsf{inv}}^{*}$, then we can find another valid hyperplane $w_{\mathsf{inv}}^{+}$ that is equally likely to have generated the same data. There is no algorithm that can distinguish between  $w_{\mathsf{inv}}^{*}$ and $w_{\mathsf{inv}}^{+}$. As a result, if we use data from the region where the hyperplanes disagree (yellow region Figure  \ref{fig:2d_impossible}b)), then the algorithm fails.

\textbf{Significance of Theorem \ref{theorem 2}.}
We showed that without the support overlap assumption on the invariant features,  OOD generalization is impossible for linear classification tasks. This is in contrast to linear regression in Theorem \ref{thm9: arjovsky} \citep{arjovsky2019invariant}, where even in the absence of the support overlap assumption, guaranteed OOD generalization was possible. Applying the above Theorem \ref{theorem 2} to the 2D case (eq. \eqref{eqn:2d_toy_example}) implies that we cannot assume that the support of invariant latent features can change, or else that case is also impossible to solve. 

Next, we ask what further assumptions are minimally needed to be able to solve the OOD generalization (eq. \eqref{eqn1: min_max_ood}). Each  classifier can be written as $\bar{w} \cdot X^e = \bar{w} \cdot S (Z_{\mathsf{inv}}^e, Z_{\mathsf{spu}}^e) =\tilde{w}_{\mathsf{inv}}\cdot Z_{\mathsf{inv}}^e + \tilde{w}_{\mathsf{spu}} Z_{\mathsf{spu}}^e  $. If $\tilde{w}_{\mathsf{spu}}\not=0$, then the classifier $\bar{w}$ is said to rely on spurious features. 

\tikzset{ 
roundnode/.style={circle, draw = black,
very thick, minimum size=1cm},
}

\begin{theorem}
\label{theorem 3}
\textbf{Sufficiency and Insufficiency of ERM and IRM.} Suppose each $e \in \mathcal{E}_{all}$ follows Assumption \ref{assumption 3_new}. Assume that a) the invariant features are strictly separable, bounded, and satisfy support overlap, b) the spurious features are bounded (Assumptions \ref{assumption 4_new}-\ref{assumption 6_new}, \ref{assumption 8_new} hold).

$\bullet$ \textbf{Sufficiency:} If the joint features satisfy support overlap (Assumption \ref{assumption joint_new} holds), then both ERM and IRM  solve the OOD generalization problem (eq. \eqref{eqn1: min_max_ood}). Also,  there exist solutions to  ERM and IRM solutions that rely on the spurious features and still achieve OOD generalization. 

$\bullet$ \textbf{Insufficiency:} If spurious features do not satisfy support overlap (Assumption \ref{assumption 7_new} is violated), then both ERM and IRM  fail at solving the OOD generalization problem (eq. \eqref{eqn1: min_max_ood}). Also,  there exist no such classifiers that rely on spurious features and also achieve OOD generalization.

\end{theorem}

\textbf{Significance of Theorem \ref{theorem 3}.} From the first part, we learn that if the support overlap is satisfied jointly for both the invariant features and the spurious features (Assumption \ref{assumption joint_new}), then either ERM or IRM can solve the OOD generalization (eq. \eqref{eqn1: min_max_ood}). Interestingly, in this case we can have classifiers that rely on the spurious features and yet solve the OOD generalization (eq. \eqref{eqn1: min_max_ood}). For the 2D case (eq. \eqref{eqn:2d_toy_example}) this case implies that the entire set $\mathcal{S}$ solves the OOD generalization (eq. \eqref{eqn1: min_max_ood}).
From the second part, we learn that if support overlap holds for invariant features but not for spurious features, then the ideal OOD optimal predictors rely only on the invariant features. In this case, methods like ERM and IRM continue to rely on spurious features and fail at OOD generalization. For the above 2D case (eq. \eqref{eqn:2d_toy_example}) this implies that only the predictors that rely only on $X_{\mathsf{inv}}^e$ in the set $\mathcal{S}$ solve the OOD generalization (eq. \eqref{eqn1: min_max_ood}). 


To summarize, we looked at SEMs for classification tasks when invariant features are fully informative, and find that the support overlap assumption over invariant features is necessary. Even in the presence of support overlap for invariant features, we showed that ERM and IRM can easily fail if  the support overlap is violated for spurious features. This raises a natural question -- Can we even solve the case with the support overlap assumption only on the invariant features? We will now show that the information bottleneck principle can help tackle these cases.

 \vspace{-0.8em}
\section{Information bottleneck principle meets invariance principle}
 \vspace{-0.8em}

\textbf{Why the information bottleneck?} The information bottleneck principle prescribes to learn a representation that compresses the input $X$ as much as possible while preserving all the relevant information about the target label $Y$ \citep{tishby2000information}. Mutual information $I(X;\Phi(X))$ is used to measure information compression. If representation $\Phi(X)$ is a deterministic transformation of $X$, then in principle we can use the entropy of $\Phi(X)$ to measure compression \citep{kirsch2020unpacking}. Let us revisit the 2D case (eq. \eqref{eqn:2d_toy_example}) and apply this principle to it. Following the second part of Theorem \ref{theorem 3}, where ERM and IRM failed,  assume that invariant features satisfy the support overlap assumption, but make no such assumption for the spurious features. Consider three choices for $\Phi$: identity (selects both features), selects invariant feature only, selects spurious feature only. The entropy of $H(\Phi(X^e))$ when $\Phi$ is the identity is $H(p^e) + \log(2)$, where $H(p^e)$ is the Shannon entropy in $\mathsf{Bernoulli}(p^e)$. If $\Phi$ selects the invariant/spurious features only, then $H(\Phi(X^e)) = \log(2)$.  Among all three choices, the one that has the least entropy and also achieves zero error is the representation that focuses on the invariant feature. We could find the OOD optimal predictor in this example just by using information bottleneck. Does it mean the invariance principle isn't needed? We answer this next.

\textbf{Why invariance?}  Consider a simple classification SEM. In each $e\in \mathcal{E}_{tr}$,   $Y^e \leftarrow X_{\mathsf{inv}}^{1,e} \oplus X_{\mathsf{inv}}^{2,e} \oplus N^e$ and $X_{\mathsf{spu}}^{e} \leftarrow Y^e \oplus V^e$, where all the random variables involved are binary valued, noise $N^e, V^e$ are Bernoulli with parameters $q$ (identical across $\mathcal{E}_{tr}$), $c^e$ (varies across $\mathcal{E}_{tr}$) respectively. If $c^e<q$, then in $\mathcal{E}_{tr}$ predictions based on  $X_{\mathsf{spu}}^e$ are better than predictions based on $X_{\mathsf{inv}}^{1,e},X_{\mathsf{inv}}^{2,e}$. If both $X_{\mathsf{inv}}^{1,e}, X_{\mathsf{inv}}^{2,e} $ are uniform Bernoulli, then these features have a higher entropy than $ X_{\mathsf{spu}}^{e} $. In this case, the information bottleneck would bar using $X_{\mathsf{inv}}^{1,e}, X_{\mathsf{inv}}^{2,e}$. Instead, we want the model to focus on $X_{\mathsf{inv}}^{1,e}$, $X_{\mathsf{inv}}^{2,e}$ and not on $X_{\mathsf{spu}}^{e}$. Invariance constraints encourage the model to focus on $X_{\mathsf{inv}}^{1,e}$, $X_{\mathsf{inv}}^{2,e}$. In this example, observe that invariant features are partially informative unlike the 2D case (eq. \eqref{eqn:2d_toy_example}).

\textbf{Why invariance and information bottleneck?} We have illustrated through simple examples when the information bottleneck is needed but not invariance and vice-versa. We now provide a simple example where both these constraints are needed at the same time. This example combines the 2D case (eq. \eqref{eqn:2d_toy_example}) and the example we highlighted in the paragraph above: $Y^e \leftarrow X_{\mathsf{inv}}^{e}\oplus N^e$, $X_{\mathsf{spu}}^{1,e} \leftarrow X_{\mathsf{inv}}^{e}  \oplus W^e$, and $X_{\mathsf{spu}}^{2,e} \leftarrow Y^e \oplus V^e$. In this case, the invariance constraint does not allow representations that use $X_{\mathsf{spu}}^{2,e}$ but does not prohibit representations that rely on $X_{\mathsf{spu}}^{1,e}$. However, information bottleneck constraints on top ensure that representations that only use $X_{\mathsf{inv}}^{e}$ are used.  We now describe an objective \footnote{Results extend to alternate objective with information bottleneck constraints and average risk as objective.
} that combines both these principles:


\tikzset{ 
roundnode/.style={circle, draw = black,
thick,  minimum size=1.cm},
}
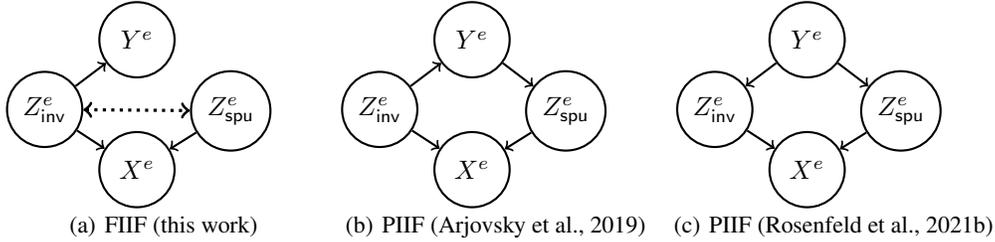
\begin{figure}[t]
    \centering
        \subfigure[FIIF (this work)]{
    \begin{minipage}{0.3\columnwidth}
    \begin{tikzpicture}[]
        \node[roundnode] (Y) {$Y^e$};
        \node[roundnode] (Z_inv)[below left=.2cm and 0.5cm of Y] {$Z^e_{\mathsf{inv}}$};
        \node[roundnode] (Z_spu)[below right=.2cm and 0.5cm of Y] {$Z^e_{\mathsf{spu}}$};
        \node[roundnode] (X)[below =.7cm of Y]{$X^e$};
        \draw[<-][thick] (Y) -- (Z_inv);
        \draw[dotted, <->][very thick] (Z_inv) -- (Z_spu);
        \draw[<-][thick] (X) -- (Z_spu);
        \draw[<-][thick] (X) -- (Z_inv);
    \end{tikzpicture}
    \vspace{0.07cm}
    \end{minipage}
    }
    \subfigure[PIIF \citep{arjovsky2019invariant} ]{
    \begin{minipage}{0.3\columnwidth}
    \begin{tikzpicture}[]
        \node[roundnode] (Y) {$Y^e$};
        \node[roundnode] (Z_inv)[below left=0.2cm and 0.5cm of Y] {$Z^e_{\mathsf{inv}}$};
        \node[roundnode] (Z_spu)[below right=0.2cm and 0.5cm of Y] {$Z^e_{\mathsf{spu}}$};
        \node[roundnode] (X)[below =.7cm of Y]{$X^e$};
        \draw[<-][thick] (Y) -- (Z_inv);
        \draw[->][thick] (Y) -- (Z_spu);
        \draw[<-][thick] (X) -- (Z_spu);
        \draw[<-][thick] (X) -- (Z_inv);
    \end{tikzpicture}
    \vspace{0.07cm}
    \end{minipage}
    }
    \subfigure[PIIF \citep{rosenfeld2020risks}]{
    \begin{minipage}{0.3\columnwidth}
    \begin{tikzpicture}[]
        \node[roundnode] (Y) {$Y^e$};
        \node[roundnode] (Z_inv)[below left=.2cm and 0.5cm of Y] {$Z^e_{\mathsf{inv}}$};
        \node[roundnode] (Z_spu)[below right=.2cm and 0.5cm of Y] {$Z^e_{\mathsf{spu}}$};
        \node[roundnode] (X)[below =.7cm of Y]{$X^e$};
        \draw[->][thick] (Y) -- (Z_inv);
        \draw[->][thick] (Y) -- (Z_spu);
        \draw[<-][thick] (X) -- (Z_spu);
        \draw[<-][thick] (X) -- (Z_inv);
    \end{tikzpicture}
    \vspace{0.07cm}
    \end{minipage}
    }
    \vspace{-0.2cm}
    \caption{Comparison of the DAG from Assumption \ref{assumption 3_new} (fully informative invariant features) vs. DAGs from \citet{rosenfeld2020risks,arjovsky2019invariant} (partially informative invariant features).
    }
    \label{figure:dag_comparison}
    \vspace{-0.5cm}
\end{figure}


\vspace{-0.25in}
\begin{equation}
\min_{
w, \Phi 
} 
\sum_{e\in \mathcal{E}_{tr}}h^e\big(w\cdot \Phi\big)  
 \quad \text{s.t.} \; \frac{1}{|\mathcal{E}_{tr}|}\sum_{e\in \mathcal{E}_{tr}}R^{e}\big(w\cdot \Phi \big) \leq r^{\mathsf{th}},\; w \in \arg\min_{\tilde{w}\in \mathbb{R}^{k\times r}} R^e(\tilde{w}\cdot \Phi), \forall e \in
 \mathcal{E}_{tr},
\label{eqn: entropy_risk_min_1}
\end{equation}
where $h^e$ in the above is a lower bounded differential entropy defined below and $r^{\mathsf{th}}$ is the threshold on the average risk. Typical information bottleneck based optimization in neural networks involves minimization of the entropy of the representation output from a certain hidden layer. For both analytical convenience and also because the above setup is a linear model, we work with the simplest form of bottleneck which directly minimizes the entropy of the output layer. Recall the definition of differential entropy of a random variable $X$,  $h(X)= -\mathbb{E}_X[\log d\mathbb{P}_X]$  and $d\mathbb{P}_X$ is the Radon-Nikodym derivative of $\mathbb{P}_X$ with respect to Lebesgue measure. Because in general differential entropy has no lower bound, we add a small independent noise term $\zeta$ \citep{kirsch2020unpacking} to the classifier to ensure that the entropy is bounded below.  We call the above optimization information bottleneck based invariant risk minimization (IB-IRM).  In summary, {\em among all the highly predictive invariant predictors we pick the ones that have the least entropy}.  If we drop the invariance constraint from the above optimization, we get information bottleneck based empirical risk minimization (IB-ERM).
In the above formulation and following result, we assume that $X^e$ are continuous random variables; the results continue to hold for discrete $X^e$ as well (See Appendix for details).



\begin{theorem}
\label{theorem4}
 \textbf{IB-IRM and IB-ERM vs. IRM and ERM}
 
 $\bullet$ \textbf{Fully informative invariant features (FIIF).} Suppose each $e \in \mathcal{E}_{all}$ follows 
Assumption \ref{assumption 3_new}. Assume that the invariant features are strictly separable, bounded, and satisfy support overlap (Assumptions \ref{assumption 4_new},\ref{assumption 6_new} and \ref{assumption 8_new} hold). Also, for each $e\in \mathcal{E}_{tr}$ $Z_{\mathsf{spu}}^e \leftarrow AZ_{\mathsf{inv}}^e + W^e$, where $A \in \mathbb{R}^{o\times m}$, $W^e\in \mathbb{R}^{o}$ is continuous, bounded, and zero mean noise.  Each solution to IB-IRM (eq. \eqref{eqn: entropy_risk_min_1}, with $\ell$ as $0$-$1$ loss, and $r^{\mathsf{th}}=q$), and IB-ERM  solves the OOD generalization (eq. \eqref{eqn1: min_max_ood}) but ERM and IRM (eq.\eqref{eqn: IRM}) fail.

$\bullet$ \textbf{Partially informative invariant features (PIIF).} Suppose each $e \in \mathcal{E}_{all}$ follows 
Assumption \ref{assumption 1_new} and $\exists\; e \in \mathcal{E}_{tr}$ such that  $ \mathbb{E}[\epsilon^eZ_{\mathsf{spu}}^e]\not=0$. If $|\mathcal{E}_{tr}|>2d$ and the set $\mathcal{E}_{tr}$ lies in a linear general position (a mild condition defined in the Appendix), then each solution to IB-IRM (eq. \eqref{eqn: entropy_risk_min_1}, with $\ell$ as square loss, $\sigma_{\epsilon}^{2} <r^{\mathsf{th}} \leq \sigma_{Y}^{2}$, where $\sigma_{Y}^{2}$ and $\sigma_{\epsilon}^2$ are the variance in the label and noise across $\mathcal{E}_{tr}$) and IRM (eq.\eqref{eqn: IRM}) solves  OOD generalization  (eq. \eqref{eqn1: min_max_ood}) but IB-ERM and ERM fail. 

\end{theorem}
 
\textbf{Significance of Theorem \ref{theorem4} and remarks.}  In the first part (FIIF), IB-ERM and IB-IRM succeed  without assuming support overlap for the spurious features, which was crucial for success of ERM and IRM in Theorem \ref{theorem 3}. This establishes that support overlap of spurious features is not a necessary condition. 
Observe that when invariant features are fully informative, IB-ERM and IB-IRM succeed, but when invariant features are partially informative IB-IRM and IRM succeed. In real data settings, we do not  know if the invariant features are fully or partially informative. Since IB-IRM is the only common winner in both the settings, it would be pragmatic to use it in the absence of domain knowledge about the informativeness of the invariant features.  In the paragraph preceding the objective in equation \eqref{eqn: entropy_risk_min_1}, we discussed examples where both the IB and IRM constraints were needed at the same time. In the Appendix, we generalize that example and show that if we change the assumptions in linear classification SEM in Assumption \ref{assumption 3_new} such that the invariant features are partially informative, then we see the  joint benefit of IB and IRM constraints.  At this point, it is also worth pointing to a result in \cite{rosenfeld2020risks}, which focused on linear classification SEMs (DAG shown in Figure \ref{figure:dag_comparison}c) with partially informative invariant features. Under the assumption of complete support overlap for spurious and invariant features, authors showed IRM succeeds.





\vspace{-0.8em}
\subsection{Proposed approach}
\vspace{-0.4em}
 We take the three terms from the optimization in equation \eqref{eqn: entropy_risk_min_1} and create a weighted combination as 
 $$\sum_{e}\Big(R^{e}(\Phi) + \lambda \|\nabla_{w, w=1.0} R^{e}(w \cdot \Phi)\|^2 + \nu h^{e}(\Phi) \Big)\leq
    \sum_{e}\Big(R^{e}(\Phi) + \lambda \|\nabla_{w, w=1.0} R^{e}(w \cdot \Phi)\|^2 + \nu h(\Phi) \Big).$$
\vspace{-0.3cm}

In the LHS above, the first term corresponds to the risks across environments, the second term approximates invariance constraint (follows the IRMv1 objective \citep{arjovsky2019invariant}), and the third term is the entropy of the classifier in each environment. 
\begin{wrapfigure}{r}{0.3\textwidth}
    \vspace{0.1cm}
    \begin{center}
    \includegraphics[trim=0 0in 0 1in, width=1.8in]{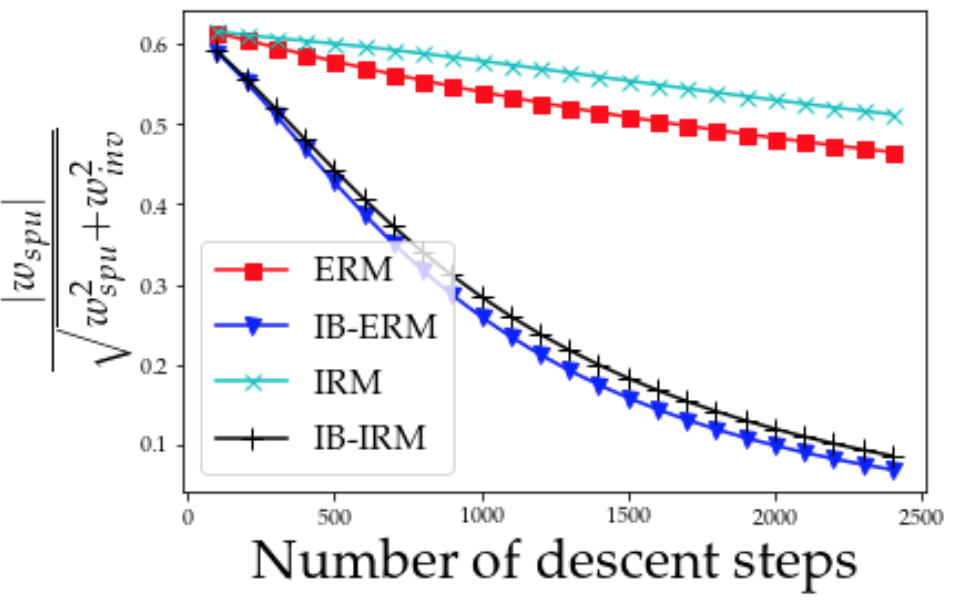}
     \caption{\small{Comparing convergence of $\frac{|w_{\mathsf{spu}}|}{\sqrt{w_{\mathsf{spu}}^2 + w_{\mathsf{inv}}^2}}$ (metric from \cite{nagarajan2020understanding}) for average selection bias $p=0.9$.}}
    \end{center}
    \label{iberm_comparison1}
    \vspace{-0.3cm}
\end{wrapfigure}
In the RHS, $h(\Phi)$ is the entropy of $\Phi$ unconditional on the environment (the entropy on the left-hand side is entropy conditional on the environment assuming all the environments are equally likely). Optimizing over differential entropy is not easy, and thus we resort to minimizing an upper bound of it \citep{kirsch2020unpacking}. We use the standard result that among all continuous random variables with the same variance, Gaussian has the maximum differential entropy. Since the entropy of Gaussian increases with its variance, we use the variance of $\Phi$ instead of the differential entropy (For further details, see the Appendix). Our final objective is given as  
\begin{equation}
    \sum_{e}\Big(R^{e}(\Phi) + \lambda\|\nabla_{w, w=1.0} R^{e}(w \cdot \Phi)\|^2 + \gamma \mathsf{Var}(\Phi) \Big). \label{eqn:finalobjective}
\end{equation}

\vspace{-.3cm}
\textbf{On the behavior of gradient descent with and without information bottleneck.} In the entire discussion so far, we have focused on ensuring that the set of optimal solutions to the desired objective (IB-IRM, IB-ERM, etc.) correspond to the solutions of the OOD generalization problem (eq. \eqref{eqn1: min_max_ood}). In some simple cases, such as the 2D case (eq. \eqref{eqn:2d_toy_example}), it can be shown that gradient descent is biased towards selecting the ideal classifier \citep{soudry2018implicit, nagarajan2020understanding}. Even though gradient descent can eventually learn the ideal classifier that only relies on the invariant features, training is frustratingly slow as was shown by \cite{nagarajan2020understanding}. In the next theorem, we characterize the impact of using IB penalty ($\mathsf{Var}(\Phi)$) in the 2D example (eq. \eqref{eqn:2d_toy_example}). We compare the methods in terms of $|\frac{w_{\mathsf{spu}}(t)}{w_{\mathsf{inv}}(t)}|$, which was the metric 
used in \cite{nagarajan2020understanding}; $w_{\mathsf{spu}}(t)$ and $w_{\mathsf{inv}}(t)$ are the weights for the spurious feature  and the invariant feature at time $t$ of training (assuming training happens with continuous time gradient descent). 

\begin{theorem}\label{theorem_ib_ls} \textbf{Impact of IB on learning speed.} Suppose each $e\in \mathcal{E}_{tr}$ follows the 2D case from equation \eqref{eqn:2d_toy_example}. Set $\lambda=0$, $\gamma>0$ in equation \eqref{eqn:finalobjective} to get the IB-ERM objective with $\ell$ as exponential loss.  Continuous-time gradient descent on this IB-ERM  objective achieves $|\frac{w_{\mathsf{spu}}(t)}{w_{\mathsf{inv}}(t)}|\leq \epsilon$ in time less than $\frac{W_{0}(\frac{1}{2\gamma})}{2(1-p)\epsilon}$ ($W_0(\cdot)$ denotes the principal branch of the Lambert $W$ function), while in the same time the ratio for ERM  $|\frac{w_{\mathsf{spu}}(t)}{w_{\mathsf{inv}}(t)}| \geq 
{\ln(\frac{1+2p}{3-2p})}
/{\ln\big(1+\frac{W_{0}(\frac{1}{2\gamma})}{2(1-p)\epsilon}\big)}
$, where $p=\frac{1}{|\mathcal{E}_{tr}|}\sum_{e\in \mathcal{E}_{tr}} p^e$ .
\end{theorem}

$|\frac{w_{\mathsf{spu}}(t)}{w_{\mathsf{inv}}(t)}|$ converges to zero for both methods, but it converges much faster for IB-ERM (for $p=0.9, \epsilon=0.001, \gamma=0.58$,  the ratio for IB-ERM is $|\frac{w_{\mathsf{spu}}(t)}{w_{\mathsf{inv}}(t)}|\leq 0.001$ and ratio for ERM is $|\frac{w_{\mathsf{spu}}(t)}{w_{\mathsf{inv}}(t)}|\geq 0.09$). In the above theorem, we  analyzed the impact of information bottleneck only. The convergence analysis for both the penalties jointly comes with its own challenges, and we hope to explore this in future work. However,  we carried out experiments with gradient descent on all the objectives for the 2D example (eq. \eqref{eqn:2d_toy_example}). See Figure 3  for the comparisons. 


\vspace{-0.5em}
\section{Experiments}
\vspace{-0.5em}

\textbf{Methods, datasets \& metrics.} We compare our approaches --  information bottleneck based ERM (IB-ERM) and information bottleneck based IRM (IB-IRM) with ERM and IRM. We also compare with an Oracle model trained on data where spurious features are permuted to remove spurious correlations. We use all the datasets in Table \ref{table1}, Terra Incognita dataset \citep{beery2018recognition}, and COCO \citep{ahmed2021systematic}. We follow the same protocol for tuning hyperparameters from \cite{aubin2021linear, arjovsky2019invariant} for their respective datasets (see the Appendix for more details). As is reported in literature, for Example 2/2S, Example 3/3S we use classification error and for AC-CMNIST, CS-CMNIST, Terra Incognita,  and COCO we use accuracy. For Example 1/1S, we use mean square error (MSE). The code for experiments can be found at \url{https://github.com/ahujak/IB-IRM}.  

\textbf{Summary of results.}
In Table \ref{table_lin_unit_test_3_env}, we provide a comparison of methods for different examples in linear unit tests \citep{aubin2021linear} for three and six training environments. In Table \ref{table_cmniste}, we provide a comparison of the methods for different CMNIST datasets, Terra Incognita and COCO dataset.
Based on our Theorem \ref{theorem4}, we do not expect ERM and IB-ERM to do well  on Example 1/1S, Example 3/3S and AC-CMNIST as  these datasets fall in the PIIF category, i.e, the invariant features are partially informative. On these examples, we find that IRM and IB-IRM do better than ERM and IB-ERM (for Example 3/3S when there are three environments all methods perform poorly). 
Based on our Theorem \ref{theorem4}, we do not expect IRM and ERM to do well  on Example 2/2S, CS-CMNIST, Terra Incognita and COCO dataset,\footnote{We place Terra Incognita and COCO dataset in the FIIF assuming that the humans who labeled the images did not need to rely on unreliable/spurious features such as background to generate the labels.} as  these datasets fall in the FIIF category, i.e., the invariant features are fully informative. On these FIIF examples, we find that IB-ERM always performs well (close to oracle), and in some cases IB-IRM also performs well.   Our experiments confirm that IB penalty has a crucial role to play in FIIF settings and IRMv1 penalty has a crucial role to play in PIIF settings (to further this claim, we provide an ablation study in the Appendix). On Example 1/1S, AC-CMNIST, we find that IB-IRM is able to extract the benefit of IRMv1 penalty. On CS-CMNIST and Example 2/2S we find that IB-IRM is able to extract the benefit of IB penalty. In settings such as COCO dataset, where IB-IRM does not perform as well as IB-ERM, better hyperparameter tuning strategies should be able to help IB-IRM adapt and put a higher weight on IB penalty.  Overall, we can conclude that IB-ERM improves over ERM (significantly in FIIF  and marginally in PIIF settings), and IB-IRM improves over IRM (improves in FIIF settings and retains advantages in PIIF settings).

\textbf{Remark.} As we move from three to six environments, we observe that MSE in Example 1/1S exhibits a larger variance. This is because of the way data is generated, the new environments that are sampled have labels that have a higher noise level (we follow the same procedure as in \cite{aubin2021linear}). 

{\small
\begin{table}[th]
\begin{center}
\adjustbox{max width=\textwidth}{%
\begin{tabular}{lcccccc}
\toprule
      & \#Envs &    ERM  & IB-ERM    & IRM  & IB-IRM & Oracle \\
      \midrule
Example1    & 3   & 13.36 $\pm$ 1.49 & 12.96 $\pm$ 1.30  & \multicolumn{1}{>{\columncolor{best}}c}{11.15$\pm$ 0.71}   & \multicolumn{1}{>{\columncolor{second}}c}{11.68 $\pm$ 0.90} & 10.42$\pm$0.16\\
Example1s   & 3   & 13.33 $\pm$ 1.49 & 12.92 $\pm$ 1.30  & 11.07 $\pm$ 0.68 & 11.74 $\pm$ 1.03 & 10.45$\pm$0.19\\ 
Example2    & 3   & 0.42 $\pm$ 0.01 & 0.00 $\pm$ 0.00  & 0.45 $\pm$ 0.00 & 0.00 $\pm$ 0.00 & 0.00 $\pm$ 0.00\\
Example2s   & 3   & 0.45 $\pm$ 0.01 & 0.00 $\pm$ 0.01  & 0.45 $\pm$ 0.01 & 0.06 $\pm$ 0.12 & 0.00 $\pm$ 0.00\\ 
Example3    & 3   & 0.48 $\pm$ 0.07 & 0.49 $\pm$ 0.06  & 0.48 $\pm$ 0.07 & 0.48 $\pm$ 0.07 & 0.01 $\pm$ 0.00\\
Example3s   & 3   & 0.49 $\pm$ 0.06 & 0.49 $\pm$ 0.06 & 0.49 $\pm$ 0.07  & 0.49 $\pm$ 0.07 & 0.01 $\pm$ 0.00\\
\midrule
Example1    &  6 & 33.74 $\pm$ 60.18 & 32.03 $\pm$ 57.05  & 23.04 $\pm$ 40.64 & 25.66 $\pm$ 45.96 & 22.21$\pm$39.25 \\
Example1s   &  6  & 33.62 $\pm$ 59.80 & 31.92 $\pm$ 56.70  & 22.92 $\pm$ 40.60 & 25.60 $\pm$ 45.62 & 22.13$\pm$38.93\\ 
Example2    &  6  & 0.37 $\pm$ 0.06 & 0.02 $\pm$ 0.05  & 0.46 $\pm$ 0.01 & 0.43 $\pm$ 0.11 & 0.00$\pm$0.00\\
Example2s   &  6  & 0.46 $\pm$ 0.01 & 0.02 $\pm$ 0.06  & 0.46 $\pm$ 0.01 & 0.45 $\pm$ 0.10 & 0.00$\pm$0.00\\ 
Example3    &  6  & 0.33 $\pm$ 0.18 & 0.26 $\pm$ 0.20  & 0.14 $\pm$ 0.18 & 0.19 $\pm$ 0.19 & 0.01$\pm$0.00\\
Example3s   &  6  & 0.36 $\pm$ 0.19 & 0.27 $\pm$ 0.20  & 0.14 $\pm$ 0.18 & 0.19 $\pm$ 0.19 & 0.01$\pm$0.00\\
\bottomrule
\end{tabular}}
\end{center}
\caption{Comparisons on linear unit tests 
in terms of mean square error (regression) and classification error (classification). 
``\#Envs'' means the number of training environments.}
\label{table_lin_unit_test_3_env}
\end{table}
}

{\small
\begin{table}[t]
\begin{center}
\adjustbox{max width=\textwidth}{%
\begin{tabular}{lcccc}
\toprule
                & ERM             & IB-ERM          & IRM            &   IB-IRM      \\
\midrule
CS-CMNIST       & 60.27 $\pm$ 1.21  & 71.80 $\pm$ 0.69 & 61.49 $\pm$ 1.45 &  71.79 $\pm$ 0.70\\
AC-CMNIST      & 16.84 $\pm$ 0.82 & 50.24 $\pm$ 0.47 & 66.98 $\pm$ 1.65 & 67.67 $\pm$ 1.78 \\ 
Terra Incognita  & 49.80 $\pm$ 4.40 & 56.40 $\pm$ 2.10 & 54.60 $\pm$ 1.30 & 54.10 $\pm$ 2.00  \\ 
COCO             & 22.70 $\pm$ 1.04 & 31.66 $\pm$ 2.39 & 18.47 $\pm$ 10.20 & 25.10 $\pm$ 1.03 \\ 
\bottomrule
\end{tabular}}
\end{center}
\caption{Classification accuracy percentage on colored MNISTs, Terra Incognita and COCO dataset.}
\label{table_cmniste}
\vspace{-.7cm}
\end{table}
}


\vspace{-0.5em}
\section{Extensions, limitations, and future work}
\vspace{-0.5em}
 
 \textbf{Extension to  non-linear models and multi-class classification.}  In this work our theoretical analysis focused on linear models. Consider the map  $X \leftarrow S(Z_{\mathsf{inv}}, Z_{\mathsf{spu}})$ in Assumption \ref{assumption 3_new}. Suppose $S$ is non-linear and bijective. We can divide the learning task into two parts a) invert $S$ to obtain $Z_{\mathsf{inv}}, Z_{\mathsf{spu}}$ and b) learn a linear model that only relies on the invariant features $Z_{\mathsf{inv}}$ to predict the label $Y$. For part b), we can rely on the approaches proposed in this work. For part a), we need to leverage advancements in the field of non-linear ICA \citep{khemakhem2020variational}. The current state-of-the-art to solve part a) requires strong structural assumptions on the dependence between all the components of $Z_{\mathsf{inv}}, Z_{\mathsf{spu}}$ \citep{lu2021nonlinear}. Therefore, solving part a) and part b) in conjunction with minimal assumptions forms an exciting future work. In the entire work, the discussion was focused on binary classification tasks and regression tasks. For multi-class classification settings, we consider natural extension of the SEM in Assumption \ref{assumption 3_new} (See the Appendix) and our main results continue to hold.

 \textbf{On the choice for IB penalty and IRMv1 penalty.} We use the approximation for entropy (in equation \eqref{eqn:finalobjective}) described in \cite{kirsch2020unpacking}. The approximation (even though an upper bound) serves as an effective proxy for the true information bottleneck as shown in the experiments in \cite{kirsch2020unpacking} (e.g., see their experiment on Imagenette dataset). Also, our experiments validate this approximation even in moderately high dimensions, as an example in CS-CMNIST, the dimension of the layer at which bottleneck constraints are applied is 256. Developing tighter approximations for information bottleneck in high dimensions and  analyzing their impact on OOD generalization is an important future work.  In recent works \citep{rosenfeld2020risks, kamath2021does, gulrajani2020search}, there has been criticism of different aspects of IRM, e.g., failure of IRMv1 penalty in non-linear models, the tuning of IRMv1 penalty, etc.  Since we use IRMv1 penalty in our proposed loss, these criticisms apply to our objective as well.  Other approximations of invariance have been proposed in the literature \citep{koyama2020out, ahuja2020invariant, chang2020invariant}. Exploring their benefits together with information bottleneck is a fruitful future work.  Before concluding, we want to remark that we have already discussed the closest related works. However, we also provide a detailed discussion of the broader related literature in the Appendix. 
 

\vspace{-0.15in}
\section{Conclusion}
\vspace{-0.15in}
In this work, we revisited the fundamental assumptions for OOD generalization for settings when invariant features capture all the information about the label. We showed how linear classification tasks are different and need much stronger assumptions than linear regression tasks. We provide a sharp characterization of performance of ERM and IRM under different assumptions on support overlap of invariant and spurious features.  We showed that support overlap of invariant features is necessary or otherwise OOD generalization is impossible. However, ERM and IRM seem to fail even in the absence of support overlap of spurious features.  We prove that a form of the information bottleneck constraint 
along with invariance goes a long way in overcoming the failures while retaining the existing provable guarantees.

 


\section*{Acknowledgements}
We thank Reyhane Askari Hemmat, Adam Ibrahim, Alexia Jolicoeur-Martineau, Divyat Mahajan, Ryan D'Orazio, Nicolas Loizou, Manuela Girotti, and Charles Guille-Escuret for the feedback.
Kartik Ahuja would also like to thank Karthikeyan Shanmugam for discussions pertaining to the related works.

\section*{Funding disclosure}
We would like to thank Samsung Electronics Co., Ldt. for funding this research. Kartik Ahuja acknowledges the support provided by IVADO postdoctoral fellowship funding program. 
Yoshua Bengio acknowledges the support from CIFAR and IBM. Ioannis Mitliagkas acknowledges support from an NSERC Discovery grant (RGPIN-2019-06512), a Samsung grant, Canada CIFAR AI chair and MSR collaborative research grant. Irina Rish acknowledges the support from Canada CIFAR AI Chair Program and from the Canada Excellence Research Chairs Program.  We thank Compute Canada for providing computational resources. 

\bibliographystyle{apalike}
\bibliography{neurips_ib_irm_jmtd}


\section*{Checklist}


\begin{enumerate}

\item For all authors...
\begin{enumerate}
  \item Do the main claims made in the abstract and introduction accurately reflect the paper's contributions and scope?
    \answerYes{See Section 2-5 and the additional details such as the proofs in the supplementary material.}
  \item Did you describe the limitations of your work?
    \answerYes{See Section 4.1 and Section 6.}
  \item Did you discuss any potential negative societal impacts of your work?
    \answerYes{See Section A.1 in the Appendix in the supplementary material.}
  \item Have you read the ethics review guidelines and ensured that your paper conforms to them?
    \answerYes{}
\end{enumerate}

\item If you are including theoretical results...
\begin{enumerate}
  \item Did you state the full set of assumptions of all theoretical results?
    \answerYes{See Section 2-4.} 
	\item Did you include complete proofs of all theoretical results?
    \answerYes{See the Appendix in the Supplementary Material.}
\end{enumerate}

\item If you ran experiments...
\begin{enumerate}
  \item Did you include the code, data, and instructions needed to reproduce the main experimental results (either in the supplemental material or as a URL)?
    \answerYes{See \url{https://github.com/ahujak/IB-IRM}}
  \item Did you specify all the training details (e.g., data splits, hyperparameters, how they were chosen)?
    \answerYes{See Section A.2 in the Appendix in the supplementary material.}
	\item Did you report error bars (e.g., with respect to the random seed after running experiments multiple times)?
    \answerYes{See Section A.2 in the Appendix in the supplementary material.}
	\item Did you include the total amount of compute and the type of resources used (e.g., type of GPUs, internal cluster, or cloud provider)?
    \answerYes{See Section A.2 in the Appendix in the supplementary material.}
\end{enumerate}

\item If you are using existing assets (e.g., code, data, models) or curating/releasing new assets...
\begin{enumerate}
  \item If your work uses existing assets, did you cite the creators?
    \answerYes{We use the codes from following github repositories \url{https://github.com/facebookresearch/DomainBed}, \url{https://github.com/facebookresearch/InvariantRiskMinimization} and \url{https://github.com/facebookresearch/InvarianceUnitTests} and we have cited the creators in the Section A.2 in the Appendix in the supplementary material.}
  \item Did you mention the license of the assets?
    \answerYes{All the repositories mentioned above use MIT license. We have mentioned this in Section A.2 in the Appendix in the supplementary material.}
  \item Did you include any new assets either in the supplemental material or as a URL?
    \answerYes{We have included code for our experiments in the supplementary material.}
  \item Did you discuss whether and how consent was obtained from people whose data you're using/curating?
    \answerNA{}
  \item Did you discuss whether the data you are using/curating contains personally identifiable information or offensive content?
    \answerNA{}
\end{enumerate}

\item If you used crowdsourcing or conducted research with human subjects...
\begin{enumerate}
  \item Did you include the full text of instructions given to participants and screenshots, if applicable?
    \answerNA{}
  \item Did you describe any potential participant risks, with links to Institutional Review Board (IRB) approvals, if applicable?
    \answerNA{}
  \item Did you include the estimated hourly wage paid to participants and the total amount spent on participant compensation?
    \answerNA{}
\end{enumerate}

\end{enumerate}


\clearpage 
\appendix

\section{Appendix}

\textbf{Organization.} In Section \ref{secn:broader_impact}, we discuss the societal impact of this work. In Section \ref{secn:experiments_details}, we provide further details on the experiments. In Section \ref{secn:structural_equation_models}, we provide a detailed discussion on structural equation models and the linear general position assumption used to prove Theorem \ref{thm9: arjovsky}. In Section \ref{secn:proof_theorem2}, we first cover the notations used in the proofs, followed by some technical remarks to be kept in mind for all the proofs, and then we provide the proof of the impossibility result in Theorem \ref{theorem 2}. In Section \ref{secn:proof_theorem3}, we provide the proof for sufficiency and insufficiency characterization of ERM and IRM
discussed in Theorem \ref{theorem 3}. In Section \ref{secn:proof_theorem4}, we provide the proof for Theorem \ref{theorem4}, which compares IB-IRM, IB-ERM with IRM and ERM. In Section \ref{secn:loss_eqn7}, we discuss the step-by-step derivation of the final objective in equation \eqref{eqn:finalobjective}. In Section \ref{secn:2d_example_details}, we provide the proof for Theorem \ref{theorem_ib_ls}, which compares the impact of information bottleneck penalty on the learning speed. In Section \ref{secn:ib_irm_conjunction}, we provide an analysis of settings when both IRM and IB penalty work together in conjunction.   Also, at the end of each section describing a proof, we provide remarks on various aspects, including some simple extensions that our results already cover. Although in the main manuscript we covered the relevant related works, in Section \ref{secn:related_works}, we provide a more detailed discussion on other related works.
\subsection{Societal impact}
\label{secn:broader_impact}
When machine learning models are deployed to assist in making decisions in safety-critical applications (e.g., self-driving cars, healthcare, etc.), we want to ensure that they make decisions that can be trusted well beyond the regime of the training data that they are exposed to. The models used in current practice are prone to exploiting spurious correlations/shortcuts in arriving at decisions and are thus not always reliable. In this work, we took some steps towards building a well-founded theory and proposing methods based on the same that can eventually help us build machines that work well beyond the training data regime. At this point, we do not anticipate a negative impact specifically of this work.


\subsection{Experiments details}
\label{secn:experiments_details}

In this section, we provide further details on the experiments. The codes to reproduce the experiments is provided at \url{https://github.com/ahujak/IB-IRM}. We have also added the codes to DomainBed (\url{https://github.com/facebookresearch/DomainBed}).

\subsubsection{Datasets} 
We first describe the datasets (Example 1/1S, Example 2/2S, Example 3/3S) introduced in \cite{aubin2021linear}; these datasets are referred to as the linear unit tests. The results for linear unit tests are presented in Table \ref{table_lin_unit_test_3_env}.

\textbf{Example 1/1S (PIIF).} This example follows the linear regression SEM from Assumption \ref{assumption 1_new}. 
The dataset in environment $e\in \mathcal{E}_{all}$ is sampled from the following 
\begin{equation*}
\begin{split}
    & Z_{\mathsf{inv}}^{e} \sim \mathcal{N}_{m}(0, (\sigma^e)^2), \;\;\;\;\; \;\;\tilde{Y}^e \sim \mathcal{N}_{m}(W_{yz}Z_{\mathsf{inv}}^{e}, (\sigma^e)^2), \\
    & Z_{\mathsf{spu}}^{e} \sim \mathcal{N}_{o}(W_{zy}\tilde{Y}^e, 1),\;\;\;\; Z^{e} \leftarrow (Z_{\mathsf{inv}}^{e},Z_{\mathsf{spu}}^{e}), \\
    & Y^{e} \leftarrow \frac{2}{(m+o)} \boldsymbol{1}_{m}^{\mathsf{T}}\tilde{Y}^{e}, \;\;\;\;\;\;\;\;\;\;\;\;\;\;\; X^e \leftarrow S(Z^e),
\end{split}
\end{equation*}
where $W_{yz} \in \mathbb{R}^{m\times m} $, $W_{zy} \in \mathbb{R}^{o\times m} $ are matrices drawn i.i.d. from the standard normal distribution, $\boldsymbol{1}_{m} \in \mathbb{R}^m$ is a vector of ones, $\mathcal{N}_{k}$ is a $k$ dimensional vector from the normal distribution. For the first three environments ($e_0, e_1,e_2$), the variances are fixed as $(\sigma^{e_0})^2 = 0.1$, $(\sigma^{e_1})^2 = 1.5$, and $(\sigma^{e_2})^2 = 2.0$.
When the number of environments is greater than three, then $(\sigma^{e_j})^2\sim \mathsf{Uniform}(10^{-2}, 10)$. The scrambling matrix $S$ is set to identity in Example 1 and a random unitary matrix is selected to rotate the latents in Example 1S. In the above dataset, the invariant features are causal and partially informative about the label. The spurious features are anti-causally related to the label and carry extra information about the label not contained in the invariant features.

\textbf{Example 2/2S (FIIF).} This example follows the linear classification SEM from Assumption \ref{assumption 3_new} with zero noise. The dataset generalizes the 2D cow versus camel classification task in equation \eqref{eqn:2d_toy_example}. Let

\begin{equation*}
\begin{split}
  & \theta_{\mathsf{cow}} = \boldsymbol{1}_{\mathsf{m}}, \;\;\;\;\;\;\;  \theta_{\mathsf{camel}} = -\theta_{\mathsf{cow}}, \;\;\;\;\;\;\; \nu_{\mathsf{animal}} = 10^{-2}, \\   
    &\theta_{\mathsf{grass}} = \boldsymbol{1}_{\mathsf{o}}, \;\;\;\;\;\;\;  \theta_{\mathsf{sand}} = -\theta_{\mathsf{grass}}, \;\;\;\;\;\;\; \nu_{\mathsf{background}} = 1. \\  
\end{split}
\end{equation*}

The dataset in environment $e\in \mathcal{E}_{all}$ is sampled from the following distribution

\begin{equation*}
\begin{split}
    &U^{e} \sim \mathsf{Categorical}\big(p^es^e, (1-p^e)s^{e}, p^{e}(1-s^{e}) , (1-p^{e})(1-s^{e}) \big),  \\ 
&Z_{\mathsf{inv}}^{e}   \sim  \begin{cases}
  (\mathcal{N}_{m}(0, 0.1) + \theta_{\mathsf{cow}})\nu_{\mathsf{animal}}\;\;\; \;\;\text{if} \; U^{e}\in\{1,2\}, \; \;\;\;\; \\
  (\mathcal{N}_{m}(0, 0.1) + \theta_{\mathsf{camel}})\nu_{\mathsf{animal}}\;\;\; \text{if} \; U^{e}\in\{3,4\}, \; \;\;\;\;  
    \end{cases} \\ 
&Z_{\mathsf{spu}}^{e}   \sim  \begin{cases}
  (\mathcal{N}_{o}(0, 0.1) + \theta_{\mathsf{grass}})\nu_{\mathsf{background}}\;\;\; \;\;\text{if} \; U^{e}\in\{1,4\}, \; \;\;\;\; \\
  (\mathcal{N}_{o}(0, 0.1) + \theta_{\mathsf{sand}})\nu_{\mathsf{background}}\;\;\;\;\; \;\text{if} \; U^{e}\in\{2,3\}, \; \;\;\;\; 
    \end{cases}\\
& Z^{e} \leftarrow (Z_{\mathsf{inv}}^{e},Z_{\mathsf{spu}}^{e}), \; \;\;\;  X^e \leftarrow S(Z^e), \\
& Y^{e} \leftarrow \mathsf{I}(\boldsymbol{1}_{m}^{\mathsf{T}} Z_{\mathsf{inv}}^{e}),
\end{split}
\end{equation*}
where for the first three environments the background parameters are $p^{e_0}=0.95$, $p^{e_1}=0.97$, $p^{e_2}=0.99$ and the animal parameters are $s^{e_0}=0.3$, $s^{e_1}=0.5$, $s^{e_2}=0.7$. When the number of environments are greater than three, then $p^{e_j} \sim \mathsf{Uniform}(0.9,1)$, and $s^{e_{j}}\sim \mathsf{Uniform}(0.3, 0.7)$. The scrambling matrix $S$ is set to identity in Example 2 and a random unitary matrix is selected to rotate the latents in Example 2S.  In the above dataset, the invariant features are causal and carry full information about the label. The spurious features are correlated with the invariant features through a confounding selection bias $U^e$.

\textbf{Example 3/3S (PIIF).} This example is a classification problem following the SEM assumed in \citep{rosenfeld2020risks}. The example is meant to 
present a linear version of the spiral classification problem in \citep{parascandolo2020learning}. Let $\theta_{\mathsf{inv}} = 0.1\cdot \boldsymbol{1}_{m}$, and $\theta_{\mathsf{spu}}^e \sim \mathcal{N}_{o}(0,1)$ for all the environments. The dataset in environment $e\in \mathcal{E}_{all}$ is sampled from the following distribution

\begin{equation}
\begin{split}
  & Y^e \sim \mathsf{Bernoulli}\Big(\frac{1}{2}\Big), \\
    &Z_{\mathsf{inv}}^{e}   \sim  \begin{cases}
  \mathcal{N}_{m}(+\theta_{\mathsf{inv}}, 0.1)\;\text{if} \; Y^{e}=0, \; \;\;\;\; \\
  \mathcal{N}_{m}(-\theta_{\mathsf{inv}}, 0.1)\;\text{if} \; Y^e=1, \; \;\;\;\;  
    \end{cases} \\ 
    &Z_{\mathsf{spu}}^{e}   \sim  \begin{cases}
  \mathcal{N}_{o}(+\theta_{\mathsf{spu}}^e, 0.1)\;\text{if} \; Y^{e}=0, \; \;\;\;\; \\
  \mathcal{N}_{o}(-\theta_{\mathsf{spu}}^e, 0.1)\;\text{if} \; Y^e=1, \; \;\;\;\;  
    \end{cases}, \\ 
    & Z^{e} \leftarrow (Z_{\mathsf{inv}}^{e},Z_{\mathsf{spu}}^{e}), \; \;\;\;  X^e \leftarrow S(Z^e). \\
\end{split} 
\end{equation}
The scrambling matrix $S$ is set to identity in Example 3 and a random unitary matrix is selected to rotate the latents in Example 3S. 
In the above dataset, the invariant features are anti-causally related to the label $Y^e$. The spurious features carry extra information about the label not contained in the invariant features. 

\textbf{AC-CMNIST dataset (PIIF).} 
We follow the same construction as was proposed in \cite{arjovsky2019invariant}.  We set up a binary classification task-- identify whether the digit is less than 5 (not including 5) or more than 5. There are three environments -- two training environments containing 25,000 data points each, one test environment containing 10,000 points. Define a preliminary label $\tilde{Y}=0$ if the digit is between 0-4 and $\tilde{Y}=1$ if the digit is between 5-9. We add noise to this preliminary label by flipping it with a 25 percent probability to construct the final label. We flip the final labels to obtain the color id $Z_{\mathsf{spu}}^e$, where the flipping probabilities are environment-dependent. The flipping probabilities are $0.2$, $0.1$, and $0.9$, in the first, second, and third environment respectively. The third environment is the testing environment. If $Z_{\mathsf{spu}}^e=1$, we color the digit red, otherwise we color it to be green. In this dataset, the color (spurious feature) carries extra information about the label not contained in the uncolored image. 

\textbf{CS-CMNIST dataset (FIIF).} We follow the same construction based on \cite{ahuja2020empirical}, except instead of a binary classification task, we set up a ten-class classification task, where the ten classes are the ten digits. For each digit class, we have an associated color.\footnote{The list of the RGB values for the ten colors are: [0, 100, 0], [188, 143, 143], [255, 0, 0], [255, 215, 0], [0, 255, 0], [65, 105, 225], [0, 225, 225], [0, 0, 255], [255, 20, 147], [160, 160, 160].}  There are also three environments -- two training environments containing 20,000 data points each, one test containing 20,000 points.  In the two training environments, the $p^e$ is set to $1.0$ and $0.9$, i.e., given the digit label the image is colored with the associated color with probability $p^e$ and with a random color with probability $1-p^e$. In the testing environment, the $p^e$ is set to $0$, i.e., all the images are colored completely at random. In this dataset, the color (spurious feature) does not carry any extra information about the label that is not already contained in the uncolored image. 

\textbf{Terra Incognita dataset (FIIF).} This dataset is a subset of the Caltech Camera Traps dataset~\citep{beery2018recognition} as formulated in \cite{gulrajani2020search}. We set up a ten-class classification task for $3\times 224\times 224$ images - identifying between 9 different species of wild animal and no animal (\{ bird, bobcat, cat, coyote, dog, empty, opossum, rabbit, raccoon, squirrel\}). There are four domains - \{L100, L38, L43, L46\} - which represents different locations of the cameras in the American Southwest. For a given location the background never change, except for illumination difference across the time of day and vegetation changes across seasons. The data is unbalanced in the number of images per location, distribution of species per location, and distribution of species overall.

\textbf{COCO dataset (FIIF).} We use COCO on colours dataset described in \cite{ahmed2021systematic} (See the details in Appendix A.2 of \cite{ahmed2021systematic}). There are ten object classes and for each object class there is a majority color associated with it, i.e., an object class assumes the background color assigned to it with $0.8$ probability. At test time, the object backgrounds are colored randomly with colors different from the ones seen in training. 


\subsubsection{Training and evaluation procedure}

\textbf{Example 1/1S, 2/2S, 3/3S.} We follow the same protocol as was prescribed in \cite{aubin2021linear} for the model selection, hyperparameter selection, training, and evaluation. For all three examples, the models used are linear. The training loss is the square error for the regression setting (Example 1/1S), and binary cross-entropy for the classification setting (Example 2/2S, 3/3S). For the two new approaches,  IB-IRM, and IB-ERM, there is a new hyperparameter $\gamma$ associated with the $\mathsf{Var}(\Phi)$ term in the final objective in equation \eqref{eqn:finalobjective}.  We use random hyperparameter search and use $20$ hyperparameter queries and average over $50$ data seeds; these numbers are the same as what was used in \cite{aubin2021linear}. We sample the $\gamma$ from $1 - 10^{\mathsf{Uniform}(-2, 0)}$ following the  practice in unit test experiments \citep{aubin2021linear}. Note that the hyperparameters are trained using training environment distribution data, which is called the train-domain validation set evaluation procedure in \cite{gulrajani2020search}. For the evaluation of performance on Example 1/1s, we reported mean square errors and standard deviations. For the evaluation of performance on Example 2/2S, Example 3/3s, we reported classification errors and standard deviations.

\textbf{AC-CMNIST dataset.} We use the default MLP architecture from  \url{https://github.com/facebookresearch/InvariantRiskMinimization}. There are two fully connected layers each with output size $256$, ReLU activation, and $\ell_2$-regularizer coefficient of $1\mathrm{e}-3$. These layers are followed by the output layer of size two.
 We use  Adam optimizer for training with a learning rate set to $1\mathrm{e}-3$. We optimize the cross-entropy loss function. We set the batch size to $256$. The total number of steps is set to 500. 
 We use grid search to search the following hyperparameters, $\lambda$ for IRMv1 penalty, and $\gamma$ for the IB penalty.  For IRM, we need to select the IRMv1 penalty $\lambda$, we set a grid of 25 values uniformly spaced in the interval $[1\mathrm{e}-1, 1.8\mathrm{e}4]$. For IB-ERM, we need to select the IB penalty $\gamma$, we set a grid of 25 values uniformly spaced in the interval $[1\mathrm{e}-1, 1.8\mathrm{e}4]$. For IB-IRM, we need to select both $\lambda$ and $\gamma$, we set a $5\times5$ uniform grid that searches over $[1\mathrm{e}-1, 1.8\mathrm{e}4]\times [1\mathrm{e}-1, 1.8\mathrm{e}4]$. Thus for IB-IRM, IB-ERM, and IRM, we search over 25 hyperparameter values. There are two procedures we tried to tune the hyperparameters  -- a) train-domain validation set tuning procedure \citep{gulrajani2020search} which takes samples from the same distribution as train domain and does limited model queries (we set 25 queries), b) oracle test-domain validation set hyperparameter tuning procedure \citep{gulrajani2020search}, which takes samples from the same distribution as test domain and does limited model queries (we set 25 queries). In \cite{arjovsky2019invariant}, the authors had used oracle test-domain validation set-based tuning, which is not ideal and is a limitation of all current approaches on AC-CMNIST. We used the same procedure in Table \ref{table_cmniste} ($5$ percent of the total data $50000$ follows the test environment distribution). In Section \ref{extra_experiments}, we show the results for all the methods when we use train-domain validation set tuning. For the evaluation, we reported the accuracy and standard deviations (averaged over thirty trials).

\textbf{CS-CMNIST dataset.} We use a ConvNet architecture with three convolutional layers with feature map dimensions of 64,128 and 256. Each convoluional layer is followed by a ReLU activation and  batch normalization layer. The final output layer is a linear layer with output dimension equal to the number of classes. We use  SGD optimizer for training with a learning rate set to $1\mathrm{e}-1$ and decay every 600 steps. We optimize the cross-entropy loss function without weight decay. We set the batch size to $128$. The total number of steps is set to $2000$. We use grid search to search the following hyperparameters, $\lambda$ for IRMv1 penalty, and $\gamma$ for the IB penalty.   For IRM, we need to select the IRMv1 penalty $\lambda$, we set a grid of 25 values uniformly spaced in the interval $[1\mathrm{e}-1, 1.8\mathrm{e}4]$. For IB-ERM, we need to select the IB penalty $\gamma$, we set a grid of 25 values uniformly spaced in the interval $[1\mathrm{e}-1, 1.8\mathrm{e}4]$. For IB-IRM, we need to select both $\lambda$ and $\gamma$, we set a $5\times5$ uniform grid that searches over $[1\mathrm{e}-1, 1.8\mathrm{e}4]\times [1\mathrm{e}-1, 1.8\mathrm{e}4]$. Thus for IB-IRM, IB-ERM, and IRM, we search over 25 hyperparameter values. In the paragraph above, we described that for AC-CMNIST all the procedures only work when using the oracle test-domain validation procedure. In the results of the CS-CMNIST experiment in the main manuscript, we showed results for the train domain validation procedure and found that IB-IRM and IB-ERM yield better performance. For completeness, we also carried oracle test-domain  validation procedure-based hyperparameter tuning for CS-CMNIST and the results are discussed in Section \ref{extra_experiments}. For the evaluation, we reported accuracy and standard deviations (averaged over five trials). In both CMNIST datasets, we had experimented with placing the IB penalty at the output layer (logits) and the penultimate layer (layer just before the logits), and found that it is much more effective to place the IB penalty on the penultimate layer. Thus in both the CMNIST datasets, the results presented use IB penalty on the penultimate layer. 

\textbf{Terra Incognita dataset.} We use the pretrained ResNet-50 model as a featurizer that outputs feature maps of size 2048 for a given image on top of which we add a 1 layer MLP which makes the classification $(2048\, \rightarrow \,9)$. We use a random hyper parameter sweep over 20 random hyperparameter configurations on which we look at the train-domain validation set to perform model selection, as described in \cite{gulrajani2020search}. The distribution of the hyper parameters are shown in Table \ref{table:hyperparameters}. Results shown in Table \ref{table_cmniste} are for the environment L100 as test environment, the reported accuracies are averaged over 3 random trial seed. For both the information bottleneck penalized algorithms (IB-ERM and IB-IRM), we apply the penalty on the feature map given by the featurizer, conditional on the environment.

\begin{table}[h]
    \caption{Hyperparameters distributions for random search given included penalty of the algorithm.} 
    \begin{center}
    { 
    \begin{tabular}{lll}
        \toprule
        \textbf{Penalty} & \textbf{Parameter} & \textbf{Random distribution}\\
        \midrule
        \multirow{3}{*}{All}          & dropout & $\text{RandomChoice}([0, 0.1, 0.5])$\\
                                      & learning rate & $10^{\text{Uniform}(-5, -3.5)}$\\
                                      & batch size    & $2^{\text{Uniform}(3, 5.5)}$\\
                                      & weight decay & $10^{\text{Uniform}(-6, -2)}$\\
        \midrule
        \multirow{2}{*}{IRMv1}        & penalty weight & $10^{\text{Uniform}(-1, 5)}$\\
                                      & annealing steps    & $10^{\text{Uniform}(0, 4)}$\\
        \midrule
        \multirow{2}{*}{IB}           & penalty weight & $10^{\text{Uniform}(-1, 5)}$\\
                                      & annealing steps    & $10^{\text{Uniform}(0, 4)}$\\
        \bottomrule
    \end{tabular}
    }
    \end{center}
    \label{table:hyperparameters}
\end{table}

\textbf{COCO dataset.} Other than the IB penalty, we use the exact same hyperparameters (default values) and setup as describe in Appendix B.2 of \cite{ahmed2021systematic} paper and the codebase that \cite{ahmed2021systematic} paper provides. For all experiments that involve an IB loss term component, IB penalty weighting of 1.0 is used and IB penalty weighting is linearly ramped up to 1.0 from epoch 1 to 200. For all experiments that involve an IRM loss term component, IRM penalty weighting of 1.0 is used, and IRM penalty weighting is linearly ramped up to 1.0 from epoch 1 to 200. Batch size of 64 is used for all experiments. We do not tune the hyperparameters in this experiment. Mean and standard deviation of classification accuracy are obtained via 4 seeds for each method.

\subsubsection{Supplementary experiments}
\label{extra_experiments}

\textbf{AC-CMNIST.} In the AC-CMNIST dataset, for completeness, we report the accuracy of the Oracle model, where the Oracle model at train time is fed images where the background colors do not have any correlation with the label. Oracle model achieved a test accuracy $70.39 \pm 0.47$ percent. In Table 5, we provide the supplementary experiments for AC-CMNIST carried out with train-domain validation set tuning procedure \citep{gulrajani2020search}. It can be seen that none of the methods work in this case. In Table 6, we provide the supplementary experiments for AC-CMNIST carried out with test-domain validation set tuning procedure \citep{gulrajani2020search}. In this case, both IB-IRM and IRM perform well.
\begin{table}[h!]
\centering
\begin{tabular}{c|cccc}
    \hline
    Method  & 5\% & 10\% & 15\%& 20\% \\
    \hline
     ERM &  $17.17\pm0.62$&$18.06\pm1.72$&$18.74\pm1.23$&$19.11\pm1.18$ \\
     IB-ERM & $17.69\pm0.54$&$17.80\pm1.81$&$16.27\pm1.20$&$18.18\pm1.46$\\
     IRM & $16.48\pm2.50$&$17.85\pm1.67$&$17.32\pm2.12$&$18.09\pm2.78$ \\ 
      IB-IRM & $18.37\pm1.44$&$17.83\pm0.65$&$18.54\pm1.42$&$19.24\pm1.49$\\
     \hline
\end{tabular}
\label{table:extra_expmt1}
    \caption{AC-CMNIST. Comparisons of the methods  using the train-domain validation set tuning procedure \citep{gulrajani2020search}. The percentages in the columns indicate what fraction of the total data ($50000$ points) is used for validation.}
\end{table}

\begin{table}[h!]
\centering
\begin{tabular}{c|cccc}
    \hline
    Method  & 5\% & 10\% & 15\%& 20\% \\
    \hline
     ERM & $16.84\pm0.82$&$17.01\pm0.83$&$16.79\pm0.89$&$16.27\pm0.93$ \\
     IB-ERM &  $50.24\pm0.47$ & $50.25\pm0.46$&$50.52\pm0.45$&$50.34\pm0.56$ \\ 
     IRM & $66.98\pm1.65$&$67.57\pm1.39$&$67.01\pm1.86$&$67.29\pm1.62$ \\
      IB-IRM & $67.67\pm1.78$ &$68.22\pm1.62$&$67.56\pm1.71$&$67.24\pm1.36$ \\
     \hline
\end{tabular}
    \caption{CS-CMNIST. Comparisons of the methods using the oracle test-domain validation set tuning procedure \citep{gulrajani2020search}. The percentages in the columns indicate what fraction of the total data ($50000$ points) is used for validation.}
\end{table}

\textbf{AC-CMNIST.}  In the CS-CMNIST dataset, for completeness, we report the accuracy of the Oracle model, which achieved a test accuracy of $99.03\pm 0.08$ percent. In Table 7, we provide the supplementary experiments for CS-CMNIST carried out with train-domain validation set tuning procedure \citep{gulrajani2020search}. In Table 8, we provide the supplementary experiments for CS-CMNIST carried out with test-domain validation set tuning procedure \citep{gulrajani2020search}. In both cases, both IB-IRM and IB-ERM RM perform well. Unlike AC-CMNIST, in the CS-CMNIST dataset both the validation procedures lead to a similar performance. 
\begin{table}[h!]
\centering
\begin{tabular}{c|cccc}
\hline
Method  & 5\% & 10\% & 15\%& 20\% \\
\hline
 ERM &  $60.27\pm1.21$&$61.02\pm0.59$&$60.35\pm1.01$&$58.59\pm1.67$\\
 IB-ERM &  $71.80\pm0.69$&$71.51\pm1.01$&$71.27\pm1.04$&$70.68\pm1.02$ \\
 IRM & $61.49\pm1.45$&$61.74\pm1.28$&$60.01\pm0.59$&$59.96\pm0.96$ \\
 IB-IRM & $71.79\pm0.70$ &$71.57\pm1.01$&$71.37\pm0.62$&$70.65\pm0.90$ \\
 \hline
\end{tabular}
    \caption{CS-CMNIST. Comparisons of the methods using the train-domain validation set tuning procedure \citep{gulrajani2020search}. The percentages in the columns indicate what fraction of the total data ($50000$ points) is used for validation.}
\end{table}

\begin{table}[h!]
\centering
\begin{tabular}{c|cccc}
\hline
Method  & 5\% & 10\% & 15\%& 20\% \\
\hline
 ERM & $61.27\pm1.40$&$61.02\pm1.59$&$60.35\pm1.01$&$58.59\pm1.67$ \\
 IB-ERM & $71.65\pm0.76$&$71.68\pm1.23$&$71.27\pm0.89$&$70.07\pm1.18$ \\
 IRM & $62.00\pm1.60$&$62.01\pm1.33$&$60.26\pm0.51$&$59.96\pm0.96$ \\
 IB-IRM & $71.90\pm0.78$&$71.07\pm0.95$&$71.18\pm0.80$&$70.75\pm1.00$ \\
 \hline
\end{tabular}
    \caption{CS-CMNIST. Comparisons of the methods using the oracle test-domain validation set tuning procedure \citep{gulrajani2020search}. The percentages in the columns indicate what fraction of the total data ($50000$ points) is used for validation}
\end{table}

\begin{figure}
    \centering
    \includegraphics[trim=0 3in 0 1in, width=5in]{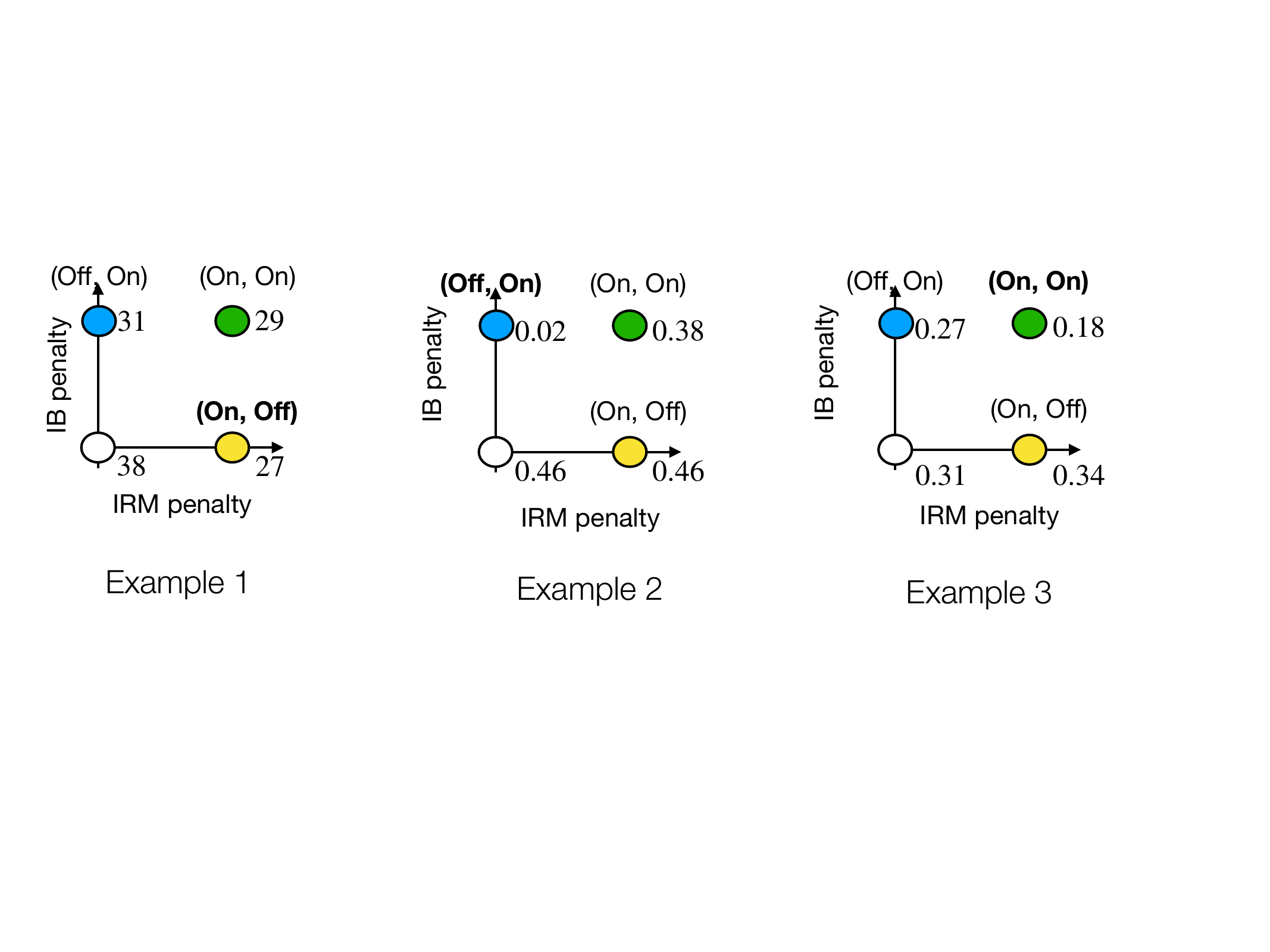}
    \caption{Illustrating the impact of the IB and IRM penalty on linear unit tests \citep{aubin2021linear}}
    \label{fig:ablation}
\end{figure}

\textbf{Ablation to understand the role of invariance penalty and information bottleneck.}  In the main body, we compared IB-IRM, IB-ERM, IRM, and ERM with the penalty of the respective methods tuned using  the validation procedures from \cite{gulrajani2020search}. In this section, we carry out an ablation analysis on linear unit tests \citep{aubin2021linear} to understand the role of the different penalties. In Figure \ref{fig:ablation}, for each example we consider the setting with six environments and show four points on a square with corresponding performance values. The bottom corner corresponds to ERM when both penalties are turned off, top corner is when both penalties are turned on, and the other two corners are when one of the penalties are on. In Example 1, which corresponds to PIIF setting, we find that IRM penalty alone helps the most. In Example 2, which corresponds to FIIF setting, we find that IB penalty helps the most. In Example 3, which again corresponds to PIIF, we find that both penalties help.


\subsubsection{Compute description}

Our computing resource is one Tesla V100-SXM2-16GB with 18 CPU cores. 

\subsubsection{Assets used and the license details}
In this work, we mainly relied on the following github repositories -- Domainbed\footnote{\url{https://github.com/facebookresearch/DomainBed} based on \cite{gulrajani2020search}}, IRM \footnote{\url{https://github.com/facebookresearch/InvariantRiskMinimization} based on \cite{arjovsky2019invariant}}, linear unit tests\footnote{\url{https://github.com/facebookresearch/InvarianceUnitTests} based on \cite{aubin2021linear}}. All the repositories mentioned above use the MIT license. We used the standard MNIST dataset \footnote{\url{http://yann.lecun.com/exdb/mnist/}} to generate the colored MNIST datasets. Other datasets we used are synthetic.
\clearpage

\subsection{Background on structural equation models} 
\label{secn:structural_equation_models}
 For completeness, we provide a more detailed background on structural equation models (SEMs), which is borrowed from \cite{arjovsky2019invariant}. 

\subsubsection{Structural equation models and assumptions on $\mathcal{E}_{all}$}
 
\begin{definition}
A structural equation model $\mathcal{C} = (\mathcal{S}, N)$ that describes the random vector $X=(X_1,\dots,X_d)$  is given as follows 
\begin{equation}
    \mathcal{S}_i : X_i \leftarrow f_i(\mathsf{Pa}(X_i), N_i),
\end{equation}
where $\mathsf{Pa}(X_i)$ are the parents of $X_i$, $N_i$ is independent noise, and $N=(N_1,\dots, N_d)$ is the noise vector. $X_j$ is said to cause $X_i$ if $X_j \in \mathsf{Pa}(X_i)$. We draw the causal graph by placing one node for each $X_i$ and drawing a directed edge from each parent to the child. The causal graphs are assumed to be acyclic.
\end{definition}

\begin{definition}
An intervention $e$ on $\mathcal{C}$ is the process of replacing one or several of its structural equations to obtain a new intervened 
SEM $\mathcal{C}^{e} = (\mathcal{S}^{e}, N^{e})$, with structural equations given as
\begin{equation}
    \mathcal{S}_{i}^{e}:X_i^{e} \leftarrow f_i^{e}(\mathsf{Pa}(X_i^{e}), N_i^{e}),
\end{equation}
where the variable $X_i^{e}$ is said to be intervened if $\mathcal{S}_i\not=\mathcal{S}_i^{e}$ or $N_i\not=N_i^{e}$
\end{definition}

The above family of interventions are used to model the environments.

\begin{definition}
\label{defn: valid_intervention}
Consider a SEM $\mathcal{C}$ that describes the random vector $(X, Y)$, where $X=(X_1, \dots, X_d),$ and the learning goal is to predict $Y$ from $X$. The set of all environments obtained using interventions $\mathcal{E}_{all}(\mathcal{C})$ indexes all the interventional distributions $\mathbb{P}^e$, where $(X^e,Y^e)\sim \mathbb{P}^e$. An intervention $e$ is  valid if the following conditions are met: i) the causal graph remains acyclic, ii) $\mathbb{E}[Y^e|\mathsf{Pa}(Y)]  = \mathbb{E}[Y|\mathsf{Pa}(Y)]$, i.e. expectation conditional on parents is invariant, and the variance $\mathsf{Var}[Y^e|\mathsf{Pa}(Y)]$ remains within a finite range.
\end{definition}

Following the above definitions it is possible to show that  a predictor that relies on causal parents only  $v:\mathbb{R}^d\rightarrow \mathcal{Y}$ and is given as $v(x) = \mathbb{E}[f_{Y}(\mathsf{Pa}(Y), N_Y)]$ solves the OOD generalization problem in equation \eqref{eqn1: min_max_ood} over the environments $\mathcal{E}_{all}(\mathcal{C})$ that form valid interventions as stated in Definition \ref{defn: valid_intervention}. Next, we provide an example to show why $v$ is OOD optimal.

\textbf{Example to illustrate why predictors that rely on causes are robust.} We reuse the toy example from \cite{arjovsky2019invariant}  to explain why models that rely on causes are more robust to valid interventions $\mathcal{E}_{all}$ discussed in the previous section. 
\begin{equation}
\begin{split}
    Y^e& \leftarrow X_{\mathsf{inv}}^e + \epsilon^e \\ 
    X_{\mathsf{spu}}^e & \leftarrow Y^e + \zeta^e
\end{split}
\end{equation}
where $X_{\mathsf{inv}}^{e} \in \mathcal{N}(0,(\sigma^e)^2)$ is the cause of $Y^e$, $N^e\in \mathcal{N}(0,(\sigma^e)^2)$ is noise, $X_{\mathsf{spu}}^e$ is the effect of $Y^e$ and $\zeta^e\in \mathcal{N}(0,1)$ is also noise. Suppose there are two training environments $\mathcal{E}_{tr}=\{e_1,e_2\}$, in the first $(\sigma^{e_1})^2=1$ and in the second $(\sigma^{e_2})^2=2$. The three possible models $w_{\mathsf{inv}}X_{\mathsf{inv}}^e + w_{\mathsf{spu}}X_{\mathsf{spu}}^e$ we could build are as follows: a) regress only on $X_{\mathsf{inv}}^e$, then in the optimal model $w_{\mathsf{inv}}=1, w_{\mathsf{spu}}=0$, b) regress only on $X_{\mathsf{spu}}^e$ and get $w_{\mathsf{inv}}=0, w_{\mathsf{spu}} = \frac{\sigma^2}{(\sigma^e)^2+\frac{1}{2}}$, c) regress on $(X_{\mathsf{inv}}^e, X_{\mathsf{spu}}^e)$ to get $w_{\mathsf{inv}}=\frac{1}{(\sigma^e)^2+1}$ and  $w_{\mathsf{inv}}=\frac{(\sigma^e)^2}{(\sigma^e)^2+1}$. Observe that the predictor that focuses on the cause only does not depend on $\sigma^2$ and is thus invariant to distribution shifts induced by change in $(\sigma^e)^2$, which is not the case with the other models.  For environment in $\mathcal{E}_{all}\setminus \mathcal{E}_{tr}$ we can change the distribution of $X_{\mathsf{inv}}^e$ and $X_{\mathsf{spu}}^e$ arbitrarily. Consider an environment $e\in \mathcal{E}_{all}$ where $X_{\mathsf{spu}}^e$ is set to a very large constant $c$, the square error of the model that relies on spurious features grows with the magnitude of $c$ but the error of the model that relies on $X_{\mathsf{inv}}^e$ does not change. Another remark we would like to make here is that in the main manuscript, we defined the notions of invariant feature map $\Phi^{*}$, and spurious feature map $\Psi^{*}$. Observe that in this example $\Phi^{*}(X^e) = X_{\mathsf{inv}}^e$, and $\Psi^{*}(X^e) =X_{\mathsf{spu}}^e$.

\subsubsection{Remark on the linear general position assumption and its implications on support overlap}
In Theorem \ref{thm9: arjovsky} that we informally stated from \cite{arjovsky2019invariant}, there is one more technical condition on that we explain below. We also explain how this assumption does not restrict the support of the latents $Z^e$ from changing arbitrarily.   

\begin{assumption}
\label{assumption 2_new}
\textbf{Linear general position.} A set of training environments $\mathcal{E}_{tr}$ lie in a linear general position of degree $r$ if $|\mathcal{E}_{tr}|>d-r+\frac{d}{r}$ for some $r \in \mathbb{N}$ and for all non-zero $x \in \mathbb{R}^d$ 
\begin{equation}
    \mathsf{dim}\Bigg(\mathsf{span}\Big(\Big\{\mathbb{E}_{X^e}[X^eX^{e \mathsf{T}}]x - \mathbb{E}_{X^e\epsilon^e}[X^e\epsilon^e]\Big\}_{e\in \mathcal{E}_{tr}}\Big)\Bigg) > d-r.
\end{equation}
\end{assumption}
The above assumption merely requires non-co-linearity of the training environments only.  The set of matrices $\mathbb{E}_{X^e}[X^eX^{e \mathsf{T}}]$ not satisfying this assumption have a zero measure (Theorem 10 \cite{arjovsky2019invariant}). Consider the case when $S$ is identity and observe that the above assumption translates to only a restriction on co-linearity of $\mathbb{E}_{Z^e}[Z^eZ^{e \mathsf{T}}]$, where $Z^e = (Z_{\mathsf{inv}}^{e}, Z_{\mathsf{spu}}^{e})$. Assume that $\mathbb{E}_{Z^e}[Z^eZ^{e \mathsf{T}}]$ is positive definite. We  explain how this Assumption \ref{assumption 2_new} does not constraint the support of the latent random variables $Z^e$.  From the set of matrices $\mathbb{E}_{Z^e}[Z^eZ^{e \mathsf{T}}]$ and $\mathbb{E}_{Z^e}[Z^e\epsilon^e]$ that satisfy the Assumption \ref{assumption 2_new}, we can construct another set of matrices with norm one that satisfy the above Assumption \ref{assumption 2_new}. Define a random variable $\tilde{Z}^e = \frac{Z^e}{c}$ and the matrices corresponding to it also satisfy the Assumption \ref{assumption 2_new}, where $c=\sqrt{\|\mathbb{E}_{Z^e}[Z^eZ^{e \mathsf{T}}]\|}$.

For all non-zero $z\in \mathbb{R}$,
\begin{equation}
\begin{split}
    \mathsf{dim}\Bigg(\mathsf{span}\Big(\Big\{\mathbb{E}_{Z^e}[Z^eZ^{e \mathsf{T}}]z - \mathbb{E}_{Z^e\epsilon^e}[Z^e\epsilon^e]\Big\}_{e\in \mathcal{E}_{tr}}\Big)\Bigg) > d-r  \implies \\ 
     \mathsf{dim}\Bigg(\mathsf{span}\Big(\Big\{\mathbb{E}_{\tilde{Z}^e}[\tilde{Z}^e\tilde{Z}^{e \mathsf{T}}]\tilde{z} - \mathbb{E}_{\tilde{Z}^e\epsilon^e}[\tilde{Z}^e\epsilon^e]\Big\}_{e\in \mathcal{E}_{tr}}\Big)\Bigg) > d-r,
    \end{split}
\end{equation}

where $\tilde{z}= zc$.
Define  $\Sigma^e = \mathbb{E}[Z^{e}Z^{e \mathsf{T}}]$ ($\tilde{\Sigma}^e = \mathbb{E}[\tilde{Z}^{e}\tilde{Z}^{e \mathsf{T}}]$) and $\rho^e = \mathbb{E}[Z^{e}\epsilon^{e}]$ ($\tilde{\rho}^e = \mathbb{E}[\tilde{Z}^{e}\epsilon^{e}]$). 
Observe that $\|\tilde{\Sigma}^e\|=1$.
So far we established that if there exist a set of matrices $\{\Sigma^e, \rho^e\}_{e\in \mathcal{E}_{tr}}$ satisfying the linear general position assumption (Assumption \ref{assumption 2_new}), then it also implies that there exist a set of matrices $\{\tilde{\Sigma}^e, \tilde{\rho}^e\}_{e\in \mathcal{E}_{tr}}$, where $\|\tilde{\Sigma}^e\|=1$, that satisfy the linear general position assumption (Assumption \ref{assumption 2_new}).  Next, we will show that the set of matrices $\{\tilde{\Sigma}^e\}_{e\in \mathcal{E}_{tr}}$, $\{\tilde{\rho}^{e}\}_{e\in \mathcal{E}_{tr}}$ can be constructed from random variables with bounded support.  We will show that  $\tilde{\Sigma}^e$ can be constructed by transforming a uniform random vector. Define a uniform random vector $K^e $, where each component $K_i^e \sim \mathsf{Uniform}[-\sqrt{3},\sqrt{3}]$. Define $\bar{Z}^e= B K^e$. Observe that 

\begin{equation}
    \mathbb{E}[\bar{Z}^e \bar{Z}^{e,\mathsf{T}}] = BB^{t}.
\end{equation}
Since every positive definite matrix can be decomposed as $BB^{t}$, we can use matrix $B$ to construct the required $\tilde{\Sigma}^e$. Since $\|\tilde{\Sigma}^e\|=1$, we get $\|BB^t\||=1 \implies \|B\|=1$. Also, $\|\bar{Z}^e\| \leq \|B\|\|K^e\| = \|K^e\|$.  Having fixed the matrix $B$ above, we use it to set the correlation $ \mathbb{E}[K^e\epsilon^e]$

\begin{equation}
B \mathbb{E}[K^e\epsilon^e] = \tilde{\rho}^e\implies \mathbb{E}[K^e\epsilon^e] = B^{-1}\tilde{\rho}^e
\end{equation}

Thus we can conclude without loss of generality that from any set of  matrices $\{\Sigma^e, \rho^e\}_{e\in \mathcal{E}_{tr}}$ satisfying the linear general position assumption, we can construct random variables with bounded support  that satisfy the linear general position assumption. 
By solving IRM (equation \eqref{eqn: IRM}) over such training environments with bounded support, we can still recover the ideal invariant predictor that solves the OOD generalization problem in equation \eqref{eqn1: min_max_ood} (i.e.,  $\nexists e\in \mathcal{E}_{all}$ for which risk $> \sigma_{\mathsf{sup}}^2$). The above conditions show that we can have the data in $\mathcal{E}_{tr}$ come from a region with bounded support, and the environments in $\mathcal{E}_{all}\setminus\mathcal{E}_{tr}$ are not required to satisfy support overlap with data from $\mathcal{E}_{tr}$, which is in stark contrast to the linear classification results that we showed. 

\clearpage

\subsection{Notations and proof of Theorem 2 (impossibility of guaranteed OOD generalization for linear classification)}
\label{secn:proof_theorem2}

\textbf{Notations for the proofs.} 
We describe the common notations used in the proofs that follow. We also remind the reader of the notation from the main manuscript for convenience.    $\circ$ is used to denote the composition of functions, $\cdot$ is used for matrix multiplication. $\mathbb{P}^e$ denotes the probability distribution over the input feature values $X^e$, and the labels $Y^e$  in environment $e$.  $Z^e$ describes the latent variables decomposed into $(Z_{\mathsf{inv}}^e, Z_{\mathsf{spu}}^e)$.
$S$ is the matrix relating $X^e$ and $Z^e$ and $X^e=S(Z^e)$.  $w$ denotes a linear classifier, $\Phi$ denotes the representation map that transforms input data into a representation, which is then fed to the classifier. $\mathsf{I}$ is the indicator function, which takes a value $1$ when the input is greater than or equal to zero, and $0$ otherwise. $\mathsf{sgn}$ is the sign function, which takes a value $1$ when the input is greater than or equal to zero, and $-1$ otherwise. In all the results, except for Theorem \ref{theorem_ib_ls}, we use $\ell$ as $0$-$1$ loss for classification, and square loss for regression.
For a discrete random variable $X\in \mathbb{R}^d$, the support is defined as $\mathcal{X} = \{x\in \mathbb{R}^d \;|\; \mathbb{P}_X(x) > 0\}$, where $\mathbb{P}_X(x)$ is the probability of $X=x$.  For a continuous random variable $X\in \mathbb{R}^d$, the support is defined as $\mathcal{X} = \{x\in \mathbb{R}^d \;|\; d\mathbb{P}_X(x) > 0\}$, where $d\mathbb{P}_X(x)$ is the Radon-Nikodym derivative of $\mathbb{P}_X$ w.r.t the Lebesgue measure over the completion of the Borel sets in $\mathbb{R}^d$ \citep{ash2000probability}.  
$\mathcal{Z}^e$, $\mathcal{Z}_{\mathsf{inv}}^e$, $\mathcal{Z}_{\mathsf{spu}}^e$, and $\mathcal{X}^e$   are the support of $Z^e$, $Z_{\mathsf{inv}}^e$, $Z_{\mathsf{spu}}^e$, and $X^e$  respectively in environment $e$.

\textbf{Remark on Assumption \ref{assumption 3_new}.} 
In all the proofs that follow, we assume that the dimension of invariant feature $m$ is greater than  or equal to $2$. Also, all the components $w^{*}_{\mathsf{inv}}$ are non-zero without loss of generality (if some component was zero, then such a latent can be a part of $Z_{\mathsf{spu}}^{e}$. $\mathcal{X}= \mathbb{R}^d$ and $\mathcal{Y}=\{0,1\}$ for classification and $\mathcal{Y}= \mathbb{R}$ for regression. Before we can prove Theorem \ref{theorem 2}, we need to prove intermediate lemmas needed as preliminary results for it. 

Define 
\begin{equation}
\mathcal{W}_{\mathsf{inv}} = \Big\{(w_{\mathsf{inv}},0)\in \mathbb{R}^{m+o}\;\big|\; \|w_{\mathsf{inv}} \|=1, \;\forall z_{\mathsf{inv}} \in \cup_{e\in \mathcal{E}_{tr}} \mathcal{Z}^{e}_{\mathsf{inv}},\; \mathsf{I}\big(w_{\mathsf{inv}}^{*}\cdot z_{\mathsf{inv}}\big) = \mathsf{I}\big(w_{\mathsf{inv}} \cdot z_{\mathsf{inv}}\big)\Big\}
\label{eqn:w_inv}
\end{equation}

This set $\mathcal{W}_{\mathsf{inv}}$ defines a family of hyperplanes equivalent to the labelling hyperplane $w_{\mathsf{inv}}^{*}$ on the training environments. Define a classifier
  $g^{*}:\mathcal{X}\rightarrow \mathcal{Y}$ as 
\begin{equation}
g^{*} = \mathsf{I}\circ \Big(\big(w_{\mathsf{inv}}^{*},0\big)\circ S^{-1}\Big) 
\label{eqn:gstar}
\end{equation}

 The classifier $g^{*}$ takes $X^{e}$ as input and outputs $\mathsf{I}(w_{\mathsf{inv}}^{*} \cdot Z_{\mathsf{inv}}^{e})$.
\begin{lemma}
\label{lemma_optimal_sol}
If we consider the set of all the environments that follow Assumption \ref{assumption 3_new}, then the classifier based on the labelling hyperplane $g^{*}$ solves equation \eqref{eqn1: min_max_ood} and achieves a risk of $q$ in each environment.
\end{lemma}

\textbf{Proof of Lemma 1.} Observe that $g^{*}$ is the classifier one would get by solving for the Bayes optimal classifier on each environment. The justification goes as follows. If $w_{\mathsf{inv}}^{*} \cdot Z_{\mathsf{inv}}^{e}\geq 0$, then $\mathbb{P}(Y^e=0|X^e) < \mathbb{P}(Y^e=1|X^e)$ (since $q<\frac{1}{2}$), which implies the prediction is $1$.  If $w_{\mathsf{inv}}^{*} \cdot Z_{\mathsf{inv}}^{e}< 0$, then $\mathbb{P}(Y^e=1|X^e) < \mathbb{P}(Y^e=0|X^e)$, which implies the prediction is $0$. We show that $g^{*}$ achieves an error of $q$ in each environment,
\begin{equation}
\begin{split}
R^{e}(g^{*}) & = \mathbb{E}\Big[Y^e \oplus \mathsf{I}(w_{\mathsf{inv}}^{*} \cdot Z_{\mathsf{inv}}^{e})\Big] \\
& = \mathbb{E}\Big[\Big(\mathsf{I}(w_{\mathsf{inv}}^{*} \cdot Z_{\mathsf{inv}}^{e}) \oplus N^e\Big) \oplus \mathsf{I}(w_{\mathsf{inv}}^{*} \cdot Z_{\mathsf{inv}}^{e})\Big] = q.
\label{eqn1: lemma_error}
\end{split}
\end{equation}

Define $\mathcal{F}$ to be the set of all the maps $\mathbb{R}^d \rightarrow \mathcal{Y}$. From the equation \eqref{eqn1: lemma_error} we get, 

\begin{equation}
\begin{split}
  &\forall e\in \mathcal{E}_{all}, \forall f \in \mathcal{F},\;  R^e(f) \geq q,  \\ 
  \implies &\forall f \in \mathcal{F}, \max_{e\in \mathcal{E}_{all}} R^{e}(f) \geq q, \\
\implies&    \min_{f\in \mathcal{F}}\max_{e\in \mathcal{E}_{all}} R^{e}(f) \geq q. \\
 \end{split}
\label{eqn1: lemma_e_all}
\end{equation}

$g^{*}$ achieves the lower bound above as it achieves an error of $q$ in each environment. This completes the proof. \hfill $\qedsymbol$

We relax the Assumption \ref{assumption 3_new} to the case where we allow for spurious features to carry extra information about the label. 
\begin{assumption}
\label{assumption 3_relaxed} \textbf{Linear classification structural equation model. (PIIF)} In each $e\in \mathcal{E}_{all}$, 
\begin{equation}
\begin{split}
  & Y^e \leftarrow \mathsf{I}\big(w_{\mathsf{inv}}^{*} \cdot Z_{\mathsf{inv}}^e\big) \oplus N^e, \;\;\;\;\;\;\; N^e \sim \mathsf{Bernoulli}(q), q<\frac{1}{2}, \;\;\;\;\;\;\;\;\;N^{e} \perp Z_{\mathsf{inv}}^{e},\\
  &  X^{e} \leftarrow S\big(Z_{\mathsf{inv}}^{e}, Z_{\mathsf{spu}}^{e}\big).
\end{split}
\end{equation}
\end{assumption}

Observe that the SEM above in Assumption \ref{assumption 3_relaxed} is analogous the the SEM in Assumption \ref{assumption 1_new}. Also, observe that in the above SEM $\exists$ $e$ such that $N^e\not\perp Z_{\mathsf{spu}}^{e}$, which makes the invariant features partially informative about the label. We show that the Lemma \ref{lemma_optimal_sol} extends to the above SEMs (Assumption \ref{assumption 3_relaxed}) as well. 

\begin{lemma}
\label{lemma_optimal_sol1}
If we consider the set of all the environments that follow Assumption \ref{assumption 3_relaxed}, then $g^{*}$ solves equation \eqref{eqn1: min_max_ood} and achieves a risk of $q$ in each environment.
\end{lemma}

\textbf{Proof of Lemma 2.}
Consider the environment $e'\in \mathcal{E}_{all}$, where $N^{e'} \perp (Z_{\mathsf{inv}}^{e'}, Z_{\mathsf{spu}}^{e'})$. Observe that in this environment $g^{*}$ is a Bayes optimal classifier and achieves a risk value of $q$. 
\begin{equation}
\begin{split} 
\forall {f} \in \mathcal{F}, R^{e'}(f) \geq q & \implies \forall f \in \mathcal{F}, \max_{e\in \mathcal{E}_{all}} R^{e}(f) \geq q, \\
  &\implies  \min_{f\in \mathcal{F}}\max_{e\in \mathcal{E}_{all}} R^{e}(f) \geq q \\
 \end{split}
\label{eqn1: lemma_e_all1}
\end{equation}
$g^{*}$ achieves the lower bound above as it achieves an error of $q$ in each environment. This completes the proof. \hfill $\qedsymbol$

\begin{lemma}
\label{lemma 1}
If Assumption \ref{assumption 3_new}, \ref{assumption 4_new}, and \ref{assumption 8_new} hold, and $m \geq 2$, then the set $\mathcal{W}_{\mathsf{inv}}$ (eq. \eqref{eqn:w_inv}) consists of infinitely many hyperplanes that are not aligned  with $w_{\mathsf{inv}}^{*}$.  
\end{lemma}

\textbf{Proof of Lemma 3.}
For each $z_{\mathsf{inv}} \in \cup_{e\in \mathcal{E}_{tr}}\mathcal{Z}_{\mathsf{inv}}^{e}$ define $y^{*} = \mathsf{sgn}(w_{\mathsf{inv}}^{*} \cdot z_{\mathsf{inv}})$.

From the definition of $\mathsf{Inv}$-$\mathsf{Margin}$ in Assumption \ref{assumption 8_new}, it follows that 
$\exists \;c>0$ such that  $\forall z_{\mathsf{inv}} \in \cup_{e\in \mathcal{E}_{tr}}\mathcal{Z}_{\mathsf{inv}}^{e}$

\begin{equation}
  y^{*} \big(w_{\mathsf{inv}}^{*} \cdot z_{\mathsf{inv}} \big) \geq c.
    \label{lemma1:eqn1}
\end{equation}
Next, we choose a $\gamma \in \mathbb{R}^{m} $ that is not in the same direction as $w_{\mathsf{inv}}^{*}$, i.e., $\nexists\; a \in \mathbb{R}$  such that $\gamma = a w_{\mathsf{inv}}^{*}$ (such a direction always exists since $m \geq 2$). Define the margin of $w_{\mathsf{inv}}^{*} + \gamma$ w.r.t labels $y^{*}$ from $w_{\mathsf{inv}}^{*}$
\begin{equation}
    y^{*}\big(w_{\mathsf{inv}}^{*}\cdot z_{\mathsf{inv}} +  \gamma \cdot z_{\mathsf{inv}} \big).
    \label{lemma1:eqn2}
\end{equation}
Using Cauchy-Schwarz inequality we get
\begin{equation}
    |y^{*} (\gamma \cdot z_{\mathsf{inv}}) | =|\gamma \cdot z_{\mathsf{inv}} | \leq \|\gamma\| \|z_{\mathsf{inv}} \|.
    \label{lemma1:eqn3}
\end{equation}

Since the support of the invariant features in training set $ \cup_{e\in \mathcal{E}_{tr}}\mathcal{Z}_{\mathsf{inv}}^{e}$ is bounded, we set the magnitude of $\gamma$ sufficiently small to control $y^{*}\big(\gamma\cdot z_{\mathsf{inv}}\big) $.  Since $ \cup_{e\in \mathcal{E}_{tr}}\mathcal{Z}_{\mathsf{inv}}^{e}$,  is bounded $\exists\; z^{\mathsf{sup}}>0$  such that   $\forall z_{\mathsf{inv}} \in \cup_{e\in \mathcal{E}_{tr}}\mathcal{Z}_{\mathsf{inv}}^{e}$, $\|z_{\mathsf{inv}}\|<z_{\mathsf{sup}}$. If $\|\gamma\| \leq \frac{c}{2z^{\mathsf{sup}}}$, then from equation \eqref{lemma1:eqn3}, we get that  for each $ z_{\mathsf{inv}} \in \cup_{e\in \mathcal{E}_{tr}}\mathcal{Z}_{\mathsf{inv}}^{e},    |y\big(\gamma \cdot z_{\mathsf{inv}}\big) | \leq \frac{c}{2}$. Using this we get for each $ z_{\mathsf{inv}} \in \cup_{e\in \mathcal{E}_{tr}}\mathcal{Z}_{\mathsf{inv}}^{e}$
\begin{equation}
  y^{*}\big(\big(w_{\mathsf{inv}}^{*} + \gamma\big)\cdot z_{\mathsf{inv}}\big)= y^{*}\big(w_{\mathsf{inv}}^{*}\cdot z_{\mathsf{inv}}\big) +  y^{*}\big(\gamma\cdot z_{\mathsf{inv}}\big) \geq y^{*}w_{\mathsf{inv}}\cdot z_{\mathsf{inv}} -     |y^{*}\gamma \cdot z_{\mathsf{inv}} | \geq \frac{c}{2}.
  \label{lemma1:eqn4}
\end{equation} 

From equation \eqref{lemma1:eqn1} and \eqref{lemma1:eqn4}, we have that  $$\mathsf{sgn}\big((w_{\mathsf{inv}}^{*} + \gamma)\cdot z_{\mathsf{inv}}\big) =\mathsf{sgn}\big(w_{\mathsf{inv}}^{*} \cdot z_{\mathsf{inv}}\big) \implies \mathsf{I}\big((w_{\mathsf{inv}}^{*} + \gamma)\cdot z_{\mathsf{inv}}\big) =\mathsf{I}\big(w_{\mathsf{inv}}^{*} \cdot z_{\mathsf{inv}}\big).$$ 

The same condition would also hold if we normalized the classifier. As a result, $$\Big(\frac{1}{\|w_{\mathsf{inv}}^{*} + \gamma\|}(w_{\mathsf{inv}}^{*} + \gamma),0\Big) \in \mathcal{W}_{\mathsf{inv}}.$$ 
Also, observe that we can construct infinite such vectors that belong to $\mathcal{W}_{\mathsf{inv}}$. A simple way to check this this is consider $\gamma^{'}= \theta\gamma$, where $\theta \in (0,1)$.  The same condition in equation \eqref{lemma1:eqn4} also holds with $\gamma$ replaced with $\gamma^{'}$. We define this set as follows 

\begin{equation}
    \mathcal{W}_{\mathsf{inv}}(\gamma) = \Big\{\Big(\frac{1}{\|w_{\mathsf{inv}}^{*} + \theta \gamma\|}(w_{\mathsf{inv}}^{*} + \theta \gamma),0\Big)\in \mathbb{R}^{m+o}\;\big|\;  \theta \in [0,1]\Big\},
    \label{lemma1:eqn5}
\end{equation}

and from the reasoning presented above it follows that $ \mathcal{W}_{\mathsf{inv}}(\gamma) \subseteq \mathcal{W}_{\mathsf{inv}}$. This completes the proof.



$\hfill$ $\qed$


We restate Theorem \ref{theorem 2} for convenience. 
\begin{theorem}
\label{theorem 2_appendix}
\textbf{Impossibility of guaranteed OOD generalization for linear classification.} Suppose  each $e \in \mathcal{E}_{all}$ follows Assumption \ref{assumption 3_new}. If for all the training environments $\mathcal{E}_{tr}$, the latent invariant features are bounded and strictly separable, i.e., Assumption \ref{assumption 4_new} and \ref{assumption 8_new} hold, then  every deterministic algorithm fails to solve the OOD generalization (eq. \eqref{eqn1: min_max_ood}), i.e.,  for the output of every algorithm $\exists \; e\in \mathcal{E}_{all}$  in which the error exceeds the minimum required value $q$ (noise level).
\end{theorem}

\textbf{Proof of Theorem \ref{theorem 2_appendix}.} 
Consider any algorithm, it takes the data from all the training environments as inputs and outputs a classifier. We write the algorithm as a map $F:\cup_{i=1}^{\infty} \Big(\mathcal{X}\times \mathcal{Y}\Big)^{i} \dots  \;|\mathcal{E}_{tr}|\; \text{times}\; \cup_{i=1}^{\infty} \Big(\mathcal{X}\times \mathcal{Y}\Big)^{i}    \rightarrow \mathcal{Y}^{\mathcal{X}}$, where $F$ takes as input data from each of the training environments and outputs a classifier, which takes as input a data point from $\mathcal{X}$ and outputs the label in $\mathcal{Y}$. For datasets $\{D^e\}_{e\in \mathcal{E}_{tr}}$ from the different training environments the output of the learner is 
$F\big(\{D^e\}_{e\in \mathcal{E}_{tr}})$. For simplicity of notation, let us denote $F\big(\{D^e\}_{e\in \mathcal{E}_{tr}})$ as $f$. 
We first show that if $f\not =g^{*}$, where $g^{*}$ is defined in equation \eqref{eqn:gstar}, then the learner cannot be OOD optimal. Take the point $x$ where the $f\not=g^{*}$. Let $z = S^{-1}(x)$. Define a test environment  where $Z^e = z$ occurs with probability $1$. In such an environment, the error achieved by 
    $f$ would be $1-q$ $(\mathbb{E}[f\oplus g^{*} \oplus N^e] = \mathbb{E}[1\oplus N^e]=1-q)$. As a result, $f$  cannot solve equation \eqref{eqn1: min_max_ood}. This observation combined with Lemma \ref{lemma_optimal_sol} leads us to the conclusion that
    $f=g^{*}$ is necessary and sufficient to solve equation \eqref{eqn1: min_max_ood} when $\mathcal{E}_{all}$ follow Assumption \ref{assumption 3_new}.

We define a family of classifiers using $\mathcal{W}_{\mathsf{inv}}$ (from eq. \eqref{eqn:w_inv}) as follows 
\begin{equation}
    \mathcal{W}_{\mathsf{inv}}^{\dagger} = \Big\{\mathsf{I}\circ\Big((w,0) \circ S^{-1}\Big)\;\Big|\; (w,0)\in \mathcal{W}_{\mathsf{inv}}\Big\}. 
    \label{eqn:w_inv_dagger}
\end{equation}



Next, we would like to show that the set $\mathcal{W}_{\mathsf{inv}}^{\dagger}$ consists of infinitely many distinct functions. 

Choose any $w_{\mathsf{inv}}^{'}$ such that $(w_{\mathsf{inv}}^{'},0) \in \mathcal{W}_{\mathsf{inv}}$ and $w_{\mathsf{inv}}^{'} \not= w_{\mathsf{inv}}^{*}$. Define  $g^{'} = \mathsf{I}\circ\Big((w_{\mathsf{inv}}^{'},0) \circ S^{-1}\Big)$. We will next show  that $g^{*}\not=g^{'}$, where $g^{*}$ was defined in equation \eqref{eqn:gstar}.

Define 
\begin{equation}
\begin{bmatrix}
w_{\mathsf{inv}}^{*} \\ 
w_{\mathsf{inv}}^{'}
\end{bmatrix} z_{\mathsf{inv}} = \begin{bmatrix}1 \\ -1 \end{bmatrix}.
\label{eqn: thm1_matrix}
\end{equation}

There are two possibilities a) $w_{\mathsf{inv}}^{'}$ is not aligned with $w_{\mathsf{inv}}^{*}$ in which case the rank of the matrix in the above equation 
\eqref{eqn: thm1_matrix} is two and as a result the range space of the matrix spans all two-dimensional vectors, b) $w_{\mathsf{inv}}^{'}$ is  aligned with $w_{\mathsf{inv}}^{*}$ but since $\|w_{\mathsf{inv}}^{'}\|=1$, $w_{\mathsf{inv}}^{'} = -w_{\mathsf{inv}}^{*}$ in which case $z_{\mathsf{inv}} = w_{\mathsf{inv}}^{*}$ solves the above equation \eqref{eqn: thm1_matrix}. In both the cases the equation \eqref{eqn: thm1_matrix} has a solution. Let  the solution of the above equation \eqref{eqn: thm1_matrix} be $\tilde{z}_{\mathsf{inv}}$. Define  $\tilde{x} = S\cdot(\tilde{z}_{\mathsf{inv}},0)$.
Therefore, from equation \eqref{eqn: thm1_matrix} it follows that 
$g^{*}(\tilde{x}) \not = g^{'}(\tilde{x})$. See the simplification below for the justification.
\begin{equation}
\begin{split}
  &  g^{*}(\tilde{x}) =  \mathsf{I}\Big((w^{*}_{\mathsf{inv}},0) \cdot S^{-1} (\tilde{x})\Big) =  \mathsf{I}(w_{\mathsf{inv}}^{*}\cdot \tilde{z}_{\mathsf{inv}}) =1 \\
  & g^{'}(\tilde{x}) =  \mathsf{I}\Big((w^{'}_{\mathsf{inv}},0) \cdot S^{-1}(\tilde{x})\Big) =  \mathsf{I}(w_{\mathsf{inv}}^{'}\cdot \tilde{z}_{\mathsf{inv}}) =0
\end{split}
\end{equation}

We showed above that $g^{*}\in \mathcal{W}_{\mathsf{inv}}^{\dagger}$ and $g^{'}\in \mathcal{W}_{\mathsf{inv}}^{\dagger}$ are two distinct functions. 
Recall in Lemma \ref{lemma 2}, we showed $\mathcal{W}_{\mathsf{inv}}$ has infinitely many distinct hyperplanes. 
We can select any pair of hyperplanes $\mathcal{W}_{\mathsf{inv}}$, for the corresponding functions in the set $ \mathcal{W}_{\mathsf{inv}}^{\dagger}$ the condition in equation \eqref{eqn: thm1_matrix} continues to hold. Thus we can conclude that there are infinitely many distinct functions in $\mathcal{W}_{\mathsf{inv}}^{\dagger}$.

Recall we described above that an algorithm can successfully solve equation \eqref{eqn1: min_max_ood}, if and only if the output $f=g^{*}$. Observe that the same exact training data $\{D^e\}_{e\in \mathcal{E}_{tr}}$ can be generated by any other labelling hyperplane $w^{'}_{\mathsf{inv}}\not=w_{\mathsf{inv}}^{*}$, where  $(w^{'}_{\mathsf{inv}},0) \in \mathcal{W}_{\mathsf{inv}}$ (this follows from the definition of $\mathcal{W}_{\mathsf{inv}}$ in equation \eqref{eqn:w_inv}). Define $g^{'}=\mathsf{I}\circ\Big((w^{'},0) \circ S^{-1}\Big)$, where  $g^{'}\in \mathcal{W}_{\mathsf{inv}}^{\dagger}$. From the justification above, we know that $g^{'}\not=g$. Since $g^{'}\not=g^{*}$ the algorithm can only be successful on one of the two labelling hyperplanes $w^{'}_{\mathsf{inv}}$ or $w_{\mathsf{inv}}^{*}$. In fact, since we showed that there are infinitely many possible distinct hyperplanes in $\mathcal{W}_{\mathsf{inv}}$, the algorithm can only succeed on one of them.  To summarize, the algorithm fails almost everywhere on the entire set, $\mathcal{W}_{\mathsf{inv}}$, of equivalent generating models. This completes the proof.\hfill $\qedsymbol$




\textbf{Remark on extension under partially informative invariant features, i.e., Assumption \ref{assumption 3_relaxed}.} The impossibility result extends to the case when the environments follow Assumption \ref{assumption 3_relaxed}. The first thing to note is that from Lemma \ref{lemma_optimal_sol1}, $g^{*}$ continues to be the OOD optimal solution hyperplane.  In the above proof, we had shown the construction of how there are infinitely many possible equally good hyperplanes that could have generated the data. To arrive at those hyperplanes, we relied on Lemma \ref{lemma 1}, where we showed that there are multiple candidate hyperplanes that could have generated the same training data. In the lemma, we only exploited the separability of latent invariant features and boundedness. If we continue to assume separability and boundedness for invariant features, then the result from Lemma \ref{lemma 1} can be used in this case as well. As a result, we can continue to use the claim that there are multiple equally good candidate hyperplanes that the algorithm cannot distinguish. Thus the impossibility result extends to this setup too.

\textbf{Remark on inveribility of $S$.} The entire proof only requires us to assume to be able to have invertibility on the latent invariant features, i.e., we should be able to recover $Z_{\mathsf{inv}}^e$ from $X^e$. Therefore, Theorem \ref{theorem 2} extends to matrices $S$ that are only invertible upto the $Z_{\mathsf{inv}}^{e}$.

\textbf{Remark on impossibility under continuous random variable assumption. } In the proof, we showed that if the test environment $e$ places all the mass on the solution of equation \eqref{eqn: thm1_matrix}, then the algorithm fails. In the setting, where we are only allowed to work with continuous random variables, can we continue to claim impossibility? The answer is yes. The reason is quite simple, we can instead of using the solution to equation \eqref{eqn: thm1_matrix} construct a small ball around that region. Since the solution to equation \eqref{eqn: thm1_matrix} that we constructed is in the interior of the half-spaces such an argument works.

\textbf{Remark on multi-class classification.} We describe a natural extension of the model in Assumption \ref{assumption 3_new} to $k$-class classification.

\begin{assumption}
\label{multi-class}
 \textbf{Linear classification structural equation model (FIIF) for multi-class classification.} 
 In each $e\in \mathcal{E}_{all}$ 
\begin{equation}
\begin{split}
  & Y^e \leftarrow \arg \max (W_{\mathsf{inv}}^{*} \cdot Z_{\mathsf{inv}}^e)  \\ 
  &  X^{e} \leftarrow S\big(Z_{\mathsf{inv}}^{e}, Z_{\mathsf{spu}}^{e}\big),
\end{split}
\end{equation}
where $W_{\mathsf{inv}}^{*} \in \mathbb{R}^{k \times m}$, $\arg \max $ is taken over the $k$ rows to generate the label $Y^e$, $S\in \mathbb{R}^{d\times d}$.
\end{assumption}
We can add noise as well in the above SEM, which uniformly at random switches the class.  The key geometric intuition for the impossibility result that we proved above, which was illustrated in Figure \ref{fig:2d_impossible}, carries over to this case provided the label generating hyperplane separates the supports of adjacent classes with a finite margin. Following the same geometric intuition, we can generalize the formal impossibility proof to this case as well for the SEM in Assumption \ref{multi-class}.



\clearpage 

\subsection{Proof of Theorem \ref{theorem 3}: sufficiency and insufficiency of ERM and IRM}
\label{secn:proof_theorem3}

\begin{lemma}
\label{lemma 2}
If Assumptions \ref{assumption 3_new},  \ref{assumption 5_new}, \ref{assumption 8_new}  hold, then there exists a classifier which puts a non-zero weight on the spurious feature and continues to be Bayes optimal in all the training environments. 

\end{lemma}

\textbf{Proof of Lemma 4.}
We will follow the construction based on Lemma \ref{lemma 1}'s proof.  

Choose an arbitrary non-zero vector $\gamma \in \mathbb{R}^o $. We will derive a bound on the margin of $(w_{\mathsf{inv}}^{*}, \gamma)$. Consider a $z_{\mathsf{inv}} \in \cup_{e\in \mathcal{E}_{tr}}\mathcal{Z}_{\mathsf{inv}}^{e}$ and  a $z_{\mathsf{spu}} \in \cup_{e\in \mathcal{E}_{tr}}\mathcal{Z}_{\mathsf{spu}}^{e}$. Define $y^{*} = \mathsf{sgn}(w_{\mathsf{inv}}^{*} \cdot z_{\mathsf{inv}})$.
The margin $(w_{\mathsf{inv}}^{*}, \gamma)$ at this point $(z_{\mathsf{inv}}, z_{\mathsf{spu}})$ with respect to $y^{*}$ is defined as
\begin{equation}
    y^{*}\big(w_{\mathsf{inv}}^{*} \cdot z_{\mathsf{inv}}\big) +  y^{*}\big(\gamma \cdot z_{\mathsf{spu}} \big).
\end{equation}

Using Cauchy-Schwarz inequality, we get 
\begin{equation}
    |y^{*}\big(\gamma \cdot z_{\mathsf{spu}}\big) | =|\gamma \cdot z_{\mathsf{spu}} | \leq \|\gamma\| \|z_{\mathsf{spu}} \|.
\end{equation}

Since the train support of spurious feature is bounded we can set the magnitude of $\gamma$ sufficiently small to control $y^{*}\big(\gamma \cdot z_{\mathsf{spu}}\big) $. 
If $\|\gamma\| \leq \frac{c}{2 z^{\mathsf{sup}}}$, then    $    |\gamma \cdot z_{\mathsf{spu}} | \leq \frac{c}{2}$, where $z^{\mathsf{sup}}$ satisfies the following condition -- for each $z \in \cup_{e\in \mathcal{E}_{tr}}\mathcal{Z}_{\mathsf{spu}}^{e}$ and $\|z\|\leq z^{\mathsf{sup}} $. 
We can use this to find a bound on the margin as follows.
Recall from equation \eqref{lemma1:eqn1} we have 

\begin{equation}
     y^{*}\big(w_{\mathsf{inv}}^{*} \cdot z_{\mathsf{inv}}\big) \geq c.
\end{equation}

We use the condition $|\gamma \cdot z_{\mathsf{spu}} | \leq \frac{c}{2}$ in the simplification below 

\begin{equation}
 y^{*}\big(w_{\mathsf{inv}}^{*} \cdot z_{\mathsf{inv}}\big) +  y^{*}\big(\gamma \cdot z_{\mathsf{spu}}\big) \geq c - |\gamma \cdot z_{\mathsf{spu}} | \geq  \frac{c}{2}.
\end{equation}

From the above equation it follows that $\mathsf{sgn}\big((w_{\mathsf{inv}}^{*}, \gamma)\cdot(z_{\mathsf{inv}}, z_{\mathsf{spu}}) \big) = \mathsf{sgn} \big((w_{\mathsf{inv}}^{*},0)\cdot(z_{\mathsf{inv}}, z_{\mathsf{spu}}) \big)  \implies \mathsf{I}\big((w_{\mathsf{inv}}^{*}, \gamma)\cdot(z_{\mathsf{inv}}, z_{\mathsf{spu}}) \big) = \mathsf{I} \big((w_{\mathsf{inv}}^{*},0)\cdot(z_{\mathsf{inv}}, z_{\mathsf{spu}}) \big)$. This condition holds for each $z_{\mathsf{inv}} \in \cup_{e\in \mathcal{E}_{tr}}\mathcal{Z}_{\mathsf{inv}}^{e}$ and  a $z_{\mathsf{spu}} \in \cup_{e\in \mathcal{E}_{tr}}\mathcal{Z}_{\mathsf{spu}}^{e}$. We use this condition to compute the error of a classifier based on $(w_{\mathsf{inv}}^{*}, \gamma)$  below. Define $g_{\mathsf{spu}}^{*} = \mathsf{I} \circ (w_{\mathsf{inv}}^{*}, \gamma) \circ S^{-1}$. The error achieved by $g_{\mathsf{spu}}^{*}$ is 

\begin{equation}
\begin{split}
    R^{e}(g_{\mathsf{spu}}^{*}) &= \mathbb{E}\Big[Y^{e} \oplus \mathsf{I}\big((w_{\mathsf{inv}}^{*}, \gamma) \cdot ( z_{\mathsf{inv}},  z_{\mathsf{spu}}) \big)\Big] \\ 
    &= \mathbb{E}\Big[\mathsf{I} \big((w_{\mathsf{inv}}^{*},0)\cdot(z_{\mathsf{inv}}, z_{\mathsf{spu}}) \big) \oplus N^{e}  \oplus \mathsf{I}\big((w_{\mathsf{inv}}^{*}, \gamma) \cdot ( z_{\mathsf{inv}},  z_{\mathsf{spu}}) \big)\Big] = \mathbb{E}\big[N^{e}\big] = q.
    \label{eqn:g_spu_error}
\end{split}
\end{equation}
The same calculation as above equation \eqref{eqn:g_spu_error} holds in all the training environments.
Thus $g_{\mathsf{spu}}^{*} $ achieves the minimum error possible $q$ for all the training environments $e\in \mathcal{E}_{tr}$. \hfill $\qedsymbol$

We restate Theorem \ref{theorem 3} for convenience. 
\begin{theorem}
\label{theorem 3_appendix}
\textbf{Sufficiency and Insufficiency of ERM and IRM.} Suppose each $e \in \mathcal{E}_{all}$ follows Assumption \ref{assumption 3_new}. Assume that a) the invariant features are strictly separable, bounded, and satisfy support overlap, b) the spurious features are bounded (Assumptions \ref{assumption 4_new}-\ref{assumption 6_new}, \ref{assumption 8_new} hold).

$\bullet$ \textbf{Sufficiency:} If the joint features satisfy support overlap (Assumption \ref{assumption joint_new} holds), then both ERM and IRM  solve the OOD generalization problem (eq. \eqref{eqn1: min_max_ood}). Also,  there exist ERM and IRM solutions that  rely on the spurious features and still achieve OOD generalization. 

$\bullet$ \textbf{Insufficiency:} If spurious features do not satisfy support overlap (Assumption \ref{assumption 7_new} is violated), then both ERM and IRM  fail at solving the OOD generalization problem (eq. \eqref{eqn1: min_max_ood}).  Also,  there exist no such classifiers that  rely on the spurious features and still achieve OOD generalization. 

\end{theorem}

\textbf{Proof of Theorem \ref{theorem 3_appendix}.} Let us begin with the first part of the Theorem. We first show that there exist solutions to ERM and IRM that rely on spurious features that also achieve OOD generalization (that is solve \eqref{eqn1: min_max_ood}). Since Assumptions \ref{assumption 3_new},  \ref{assumption 5_new}, \ref{assumption 8_new}, hold we can use Lemma \ref{lemma 2}. From Lemma \ref{lemma 2}, it  follows  that for each $z_{\mathsf{inv}} \in \cup_{e\in \mathcal{E}_{tr}}\mathcal{Z}_{\mathsf{inv}}^{e}$ and  for each $z_{\mathsf{spu}} \in \cup_{e\in \mathcal{E}_{tr}}\mathcal{Z}_{\mathsf{inv}}^{e}$:

\begin{equation}
\mathsf{I}\big((w_{\mathsf{inv}}^{*}, \gamma)\cdot(z_{\mathsf{inv}}, z_{\mathsf{spu}}) \big) = \mathsf{I} \big((w_{\mathsf{inv}}^{*},0)\cdot(z_{\mathsf{inv}}, z_{\mathsf{spu}}) \big).
\end{equation}
From Assumption \ref{assumption 6_new} and  \ref{assumption 7_new} it follows that for each $z_{\mathsf{inv}} \in \cup_{e\in \mathcal{E}_{all}}\mathcal{Z}_{\mathsf{inv}}^{e}$ and  for each $z_{\mathsf{spu}} \in \cup_{e\in \mathcal{E}_{all}}\mathcal{Z}_{\mathsf{inv}}^{e}$.
\begin{equation}
\mathsf{I}\big((w_{\mathsf{inv}}^{*}, \gamma)\cdot(z_{\mathsf{inv}}, z_{\mathsf{spu}}) \big) = \mathsf{I} \big((w_{\mathsf{inv}}^{*},0)\cdot(z_{\mathsf{inv}}, z_{\mathsf{spu}}) \big)
\end{equation}
Therefore, the error of the classifier $g_{\mathsf{spu}}^{*} = \mathsf{I} \circ (w_{\mathsf{inv}}^{*}, \gamma) \circ S^{-1}$ in each environment $e \in \mathcal{E}_{all}$ is 

\begin{equation}
\begin{split}
    R^e(g^{*}_{\mathsf{spu}}) &= \mathbb{E}\Big[Y^{e} \oplus \mathsf{I}\big((w_{\mathsf{inv}}^{*}, \gamma) \cdot ( z_{\mathsf{inv}},  z_{\mathsf{spu}}) \big)\Big] \\ 
    &= \mathbb{E}\Big[\mathsf{I} \big((w_{\mathsf{inv}}^{*},0)\cdot(z_{\mathsf{inv}}, z_{\mathsf{spu}}) \big) \oplus N^{e}  \oplus \mathsf{I}\big((w_{\mathsf{inv}}^{*}, \gamma) \cdot ( z_{\mathsf{inv}},  z_{\mathsf{spu}}) \big)\Big] = \mathbb{E}\big[N^{e}\big] = q.
\end{split}
\end{equation}
 $g^{*}_{\mathsf{spu}}$ is Bayes optimal on each environment $e\in \mathcal{E}_{all}$. Therefore, $g^{*}_{\mathsf{spu}}$ also solves equation \eqref{eqn1: min_max_ood}. Since $g^{*}_{\mathsf{spu}}$ is optimal in all the environments, it also solves ERM  as it also minimizes the sum of risks across training environments.  $g^{*}_{\mathsf{spu}}$ is also a valid invariant predictor since it is simultaneously optimal across all the environments. Since $g^{*}_{\mathsf{spu}}$ achieves an average error of $q$ across training environments, each solution to ERM and IRM has to achieve an error of $q$ in all the training environments as well. Since the solution to ERM and IRM achieves an error of $q$ it cannot differ from $g^{*}$ (defined in equation \eqref{eqn:gstar}) at any point in the training support. This argument holds in a pointwise sense when $Z_{\mathsf{inv}}^e$ is a discrete random variable, otherwise, say when $Z_{\mathsf{inv}}^e$ is a continuous random variable this argument can only be violated over a set of measure zero.\footnote{The continuous random variable case can give rise to some pathological shifts. We show later in the proof of Theorem \ref{theorem4} as to why we do not need to worry about these pathological shifts.} Owing to the joint feature support overlap between $\mathcal{E}_{tr}$ and $\mathcal{E}_{all}$, each solution to ERM and IRM continues to succeed in $\mathcal{E}_{all}$.  This completes the first part of the proof.  

We now move to the next part of the theorem, where the spurious features do not satisfy support overlap assumption (Assumption \ref{assumption 7_new}). Consider a linear classifier that the method learns  $\mathsf{I} \circ w$, where $\mathsf{I}$ is composed with a linear function. The classifier operates on $x$, and we get $\mathsf{I}( w \cdot x)$ and since $x=Sz$ (from Assumption \ref{assumption 3_new}) we can write this as $\mathsf{I}(w\cdot S(z))$. To simplify notation, we call  $\mathsf{I} \circ w \circ S =  \mathsf{I} \circ \tilde{w}$. Our goal is to show that if $\tilde{w}$ assigns a non-zero weight to the spurious features, then $\mathsf{I} \circ w \circ S$ cannot solve the OOD generalization problem (eq. \eqref{eqn1: min_max_ood}).  We write $\tilde{w} = (\tilde{w}_{\mathsf{inv}}, \tilde{w}_{\mathsf{spu}})$. Suppose $\tilde{w}_{\mathsf{spu}}\not=0$ and yet the classifier solves the problem in equation \eqref{eqn1: min_max_ood}. Consider the classifier that generates the data $(w_{\mathsf{inv}}^{*},0)$. Pick any point $z_{\mathsf{inv}}  \in \cup_{e\in \mathcal{E}_{all}}\mathcal{Z}^{e}_{\mathsf{inv}}$ and pick any non-zero $ z_{\mathsf{spu}}^{e}\in \mathbb{R}^o$. Call $z=(z_{\mathsf{inv}}, z_{\mathsf{spu}})$ We divide the analysis into two cases.

Case 1: $\mathsf{I}\big((\tilde{w}_{\mathsf{inv}}, \tilde{w}_{\mathsf{spu}}) \cdot z \big) \not= \mathsf{I}\big((w_{\mathsf{inv}}^{*}, 0) \cdot z \big) $.  In this case, $(\tilde{w}_{\mathsf{inv}}, \tilde{w}_{\mathsf{spu}})$ cannot solve equation \eqref{eqn1: min_max_ood} as there exists a test environment where we have all the mass on $z$.

Case 2: $\mathsf{I}\big((\tilde{w}_{\mathsf{inv}}, \tilde{w}_{\mathsf{spu}}) \cdot z \big) = \mathsf{I}\big((w_{\mathsf{inv}}^{*}, 0) \cdot z \big) $. Observe that since $\tilde{w}_{\mathsf{spu}}\not=0$, we can increase or decrease one of the components of $z_{\mathsf{spu}}$ corresponding to a non-zero $\tilde{w}_{\mathsf{spu}}$ until the two classifiers disagree in which case we get Case 1. Note that since Assumption \ref{assumption 7_new} does not hold, we are allowed to change $z_{\mathsf{spu}}$ arbitrarily. 

Thus we have established that a classifier cannot be OOD optimal if it assigns a non-zero weight to the spurious feature. As a result, the classifier from the first part  $g^{*}_{\mathsf{spu}}$ which assigned non-zero weight to spurious features cannot be OOD optimal without the Assumption \ref{assumption 7_new}. However, $g^{*}_{\mathsf{spu}}$ continues to be in the solution space of both ERM and IRM as it is still
Bayes optimal across all the train environments, which is why both ERM and IRM fail. At this point the proof of the statement of theorem is complete. However, we give a characterization of optimal solutions in the next paragraph.

Now let us consider any classifier in $w \in \mathcal{W}_{\mathsf{inv}}$ (from equation \eqref{eqn:w_inv}) written as $w= (w_{\mathsf{inv}},0)$. For such a classifier  by definition it is true that for each $ z_{\mathsf{inv}} \in \cup_{e\in \mathcal{E}_{tr}}\mathcal{Z}_{\mathsf{inv}}^{e}$,
$\mathsf{I}\big(w_{\mathsf{inv}} \cdot z_{\mathsf{inv}} \big) = \mathsf{I}\big(w_{\mathsf{inv}}^{*} \cdot z_{\mathsf{inv}} \big) $. From Assumption \ref{assumption 6_new} it follows that for each $ z_{\mathsf{inv}} \in \cup_{e\in \mathcal{E}_{all}}\mathcal{Z}_{\mathsf{inv}}^{e}$, $\mathsf{I}\big(w_{\mathsf{inv}} \cdot z_{\mathsf{inv}} \big) = \mathsf{I}\big(w_{\mathsf{inv}}^{*} \cdot z_{\mathsf{inv}} \big) $ and thus the classifier continues to achieve an error of $q$ on all the test environments. Thus we can conclude that $\mathsf{I}\circ w \circ S^{-1}$ is OOD optimal.  Therefore, all the elements in the set 
$\mathcal{W}_{\mathsf{inv}}^{\dagger}$ (from eq. \eqref{eqn:w_inv_dagger}) are OOD optimal.


\hfill $\qedsymbol$

\textbf{Remark on invertibility of $S$.}  The proof extends to the case when we can invert and recover entire $Z_{\mathsf{inv}}^e$  and also recover at least one component of the spurious features $Z_{\mathsf{spu}}^e$.

\textbf{Remark on failure of ERM and IRM under continuous random variable assumption.} In the proof, we showed that if the test environment $e$ places all the mass on the solution to Case 1, then the algorithm fails. In the setting, where we are only allowed to work with 
continuous random variables, can we continue to make the claim for impossibility? The answer is yes. The reason is quite simple, we can instead of using the solution to Case 1 construct a small ball around that region, where the classifiers continue to disagree.

\textbf{Remark on multi-class classification.} We extend the result to the above SEM in Assumption \ref{multi-class}. The reason ERM and IRM fail in this case is two fold -- a) there exists a hyperplane that perfectly separates the support of the invariant features with a finite margin and b) support of spurious features are allowed to change. In the multi-class case, we can use the same reasoning -- if there is a hyperplane that perfectly separates for adjacent classes, ERM and IRM continue to fail as long as the support of spurious features is allowed to change.

\clearpage 

\subsection{Proof of Theorem 4: IB-IRM and IB-ERM vs. IRM and ERM}
\label{secn:proof_theorem4}
We now lay down some properties of the entropy of discrete random variables and in parallel also lay down the properties of differential entropy of continuous random variables. Recall that a discrete random variable has a non-zero probability at each point in its support and a continuous random variable has a zero probability (and a positive density) at each point in the support.

The entropy or the Shannon entropy of a discrete random variable $X\sim \mathbb{P}_X$ with support $\mathcal{X}$ is defined as 
\begin{equation}
H(X) = -\sum_{x \in \mathcal{X}} \mathbb{P}_X(X=x) \log \big(\mathbb{P}_X(X=x)\big).
\end{equation}

The differential entropy of a continuous random variable $X \sim \mathbb{P}_X$ with support $\mathcal{X}$  is given as follows

\begin{equation}
    h(X) = -\int_{x\in \mathcal{X}}  \log\big(d\mathbb{P}_X(x)\big)d\mathbb{P}_X(x),
\end{equation}
where $d\mathbb{P}_X(x)$ is the Radon-Nikodym derivative of $\mathbb{P}_{X}$ w.r.t the Lesbegue measure.

\begin{lemma} \label{lemma 3} If $X$ and $Y$ are  discrete scalar valued random variables that are independent, then 
$$H(X+Y) \geq \max \Big\{ H(X), H(Y)\Big\}.$$
\end{lemma}

\textbf{Proof of Lemma 5.} Define $Z=X+Y$.

\begin{equation}
\begin{split}
     H(Z|X) &= -\sum_{x\in \mathcal{X}} \mathbb{P}_{X}(x)\sum_{z\in \mathcal{Z}} \mathbb{P}_{Z|X}(Z=z|X=x) \log\Big(\mathbb{P}_{Z|X}(Z=z|X=x)\Big) \\ 
 & = -\sum_{x\in \mathcal{X}} \mathbb{P}_{X}(x)\sum_{z\in \mathcal{Z}} \mathbb{P}_{Y|X}(Y=z-x|X=x) \log\Big(\mathbb{P}_{Y|X}(Y=z-x|X=x)\Big) \\
  & = -\sum_{x\in \mathcal{X}}\mathbb{P}_{X}(x) \sum_{z\in \mathcal{Z}} \mathbb{P}_{Y|X}(Y=z-x|X=x) \log\Big(\mathbb{P}_{Y|X}(Y=z-x|X=x)\Big) \; (\text{use}\; X\perp Y)\\ 
  & = -\sum_{x\in \mathcal{X}} \mathbb{P}_{X}(x)\sum_{z\in \mathcal{Z}} \mathbb{P}_{Y}(Y=z-x) \log\Big(\mathbb{P}_{Y}(Y=z-x)\Big) \\
  & = H(Y)
\end{split}
\label{eqn1:prelem-lemma1}
\end{equation}
\begin{equation}
\begin{split}
    & I(Z;X) = H(Z) - H(Z|X) = H(X+Y)- H(Y) \\ 
    & I(Z;Y) = H(Z) - H(Z|Y) = H(X+Y)- H(X)
\end{split}
\label{eqn2:prelem-lemma1}
\end{equation}

From equation \eqref{eqn2:prelem-lemma1} and the property of mutual information that $I(Z;X)\geq 0, I(Z;Y)\geq 0$ it follows that 
\begin{equation}
    \begin{split}
        H(X+Y) \geq H(Y), \; H(X+Y) \geq H(X) \implies H(X+Y) \geq \max\{H(X),H(Y)\}.
    \end{split}
\end{equation}

This completes the proof. \hfill $\qedsymbol$

\begin{lemma} \label{lemma 4} If $X$ and $Y$ are  continuous scalar valued random variables that are independent, then
$$h(X+Y) \geq \max \Big\{ h(X), h(Y)\Big\}. $$
\end{lemma}

\textbf{Proof of Lemma 6.} Define $Z=X+Y$.

\begin{equation}
\begin{split}
     h(Z|X) &= \mathbb{E}_{\mathbb{P}_{X}}\Big[\mathbb{E}_{\mathbb{P}_{Z|X}}\Big[\log\Big(d\mathbb{P}_{Z|X}(Z=z|X=x)\Big)\Big] \Big]\\ 
 & = \mathbb{E}_{\mathbb{P}_{X}}\Big[\mathbb{E}_{\mathbb{P}_{Y|X}}\Big[\log\Big(d\mathbb{P}_{Y|X}(Y=z-x|X=x)\Big)\Big] \Big]\; (\text{use}\; X\perp Y)\\ \\ 
  & = h(Y)
\end{split}
\label{eqn1:prelem-lemma1}
\end{equation}
Note that $I(Z;X) \geq 0 \implies h(Z) \geq h(Z|X)$. Combining this with the above equation \eqref{eqn1:prelem-lemma1} we get 
\begin{equation}
    h(X+Y) \geq h(Y).
\end{equation}
From symmetry it follows that $h(X+Y)\geq h(X)$. This completes the proof. \hfill $\qedsymbol$

\begin{lemma} 
\label{lemma 5} If $X$ and $Y$ are  discrete scalar valued random variables that are independent with the supports satisfying $2\leq |\mathcal{X}|<\infty$, $2\leq |\mathcal{Y}|<\infty$, then 
$$H(X+Y) > \max \Big\{ H(X), H(Y)\Big\}. $$
\end{lemma}

\textbf{Proof of Lemma 7.} Suppose $|\mathcal{X}|=\{x_{\mathsf{min}},\dots, x_{\mathsf{max}}\}$ and $\mathcal{Y}=\{y_{\mathsf{min}},\dots, y_{\mathsf{max}}\}$. The smallest value of $X+Y$ is $x_{\mathsf{min}} + y_{\mathsf{min}}$ and the largest value is $x_{\mathsf{max}} + y_{\mathsf{max}}$. Suppose that the inequality in the claim is not true in which case from Lemma \ref{lemma 3} it follows  $H(X+Y) = H(X)$ or $H(X+Y)=H(Y)$. Suppose $H(X+Y) = H(X)$, then from equation \eqref{eqn2:prelem-lemma1} it follows that $I(X+Y; Y) = 0 \implies X+Y \perp Y $. Observe that if $Z=x_{\mathsf{max}} + y_{\mathsf{max}} \implies Y = y_{\mathsf{max}}$. Therefore, $\mathbb{P}(Y = y_{\mathsf{max}}|Z=x_{\mathsf{max}} + y_{\mathsf{max}})=1$. However, $\mathbb{P}(Y = y_{\mathsf{max}})\not=1$ as the support of $Y$ has at least two elements. This contradicts $X+Y \perp Y$. As a result, $H(X+Y) \not=H(X)$. We can symmetrically show that $H(X+Y) \not= H(Y)$. Combining this with  Lemma \ref{lemma 3}, it follows that $H(X+Y) > \max\{H(X), H(Y)\}$. \hfill $\qedsymbol$

\begin{lemma}
\label{lemma 6} If $X$ and $Y$ are continuous scalar valued random variables that are independent and have a bounded support, then 
$$h(X+Y) > \max \Big\{ h(X), h(Y)\Big\} $$
\end{lemma}
\textbf{Proof of Lemma 8.} The steps of the proof are similar to Lemma \ref{lemma 5}. Suppose the inequality in the claim is not true in which case from Lemma \ref{lemma 4} it follows that either $h(X+Y)=h(X)$ or $h(X+Y)=h(Y)$. Suppose $h(X+Y)=h(X)$ which implies $I(X+Y;Y)=0 \implies X+Y \perp Y$. The support of $X$ can be written in the form of union of intervals. Suppose we consider the rightmost interval and we write it as 
$[x_{\mathsf{max}}-\Delta, x_{\mathsf{max}}]$. Similarly for $Y$, we write the rightmost interval as $[y_{\mathsf{max}}-\Delta, y_{\mathsf{max}}]$. \footnote{We use same $\Delta$ for both $X$ and $Y$ because can take the smaller of the rightmost intervals for $X$ and $Y$.} Define an event $\mathcal{M}: x_{\mathsf{max}} + y_{\mathsf{max}} -\delta \leq X+Y\leq x_{\mathsf{max}} + y_{\mathsf{max}}$. 
If $\mathcal{M}$ occurs, then $Y\geq y_{\mathsf{max}}-\delta$ and $X\geq x_{\mathsf{max}}-\delta$. 
\begin{equation}
\begin{split}
    \mathbb{P}_X(X \leq x_{\mathsf{max}}-\delta | \mathcal{M}) = 0 \\
    \mathbb{P}_Y(Y \leq y_{\mathsf{max}}-\delta | \mathcal{M}) = 0
\end{split}
\label{lemma6:eqn1}
\end{equation}
If $\delta<\Delta$ we know that 
\begin{equation}
\begin{split}
    \mathbb{P}_X(X \leq x_{\mathsf{max}}-\delta ) >0 \\
    \mathbb{P}_Y(Y \leq y_{\mathsf{max}}-\delta ) > 0
\end{split}
\label{lemma6:eqn2}
\end{equation}
If $X+Y \perp Y$ then $\mathbb{P}_Y(Y \leq y_{\mathsf{max}}-\delta ) = \mathbb{P}_Y(Y \leq y_{\mathsf{max}}-\delta | \mathcal{M})$, which is not the case from the above equations \eqref{lemma6:eqn1} and \eqref{lemma6:eqn2}. 
Thus $X+Y \not\perp Y \implies I(X+Y;Y)>0 \implies h(X+Y) > h(X)$. We can say the same for  $Y$ and conclude that $h(X+Y) > h(Y)$. This completes the proof. \hfill $\qedsymbol$

Theorem \ref{theorem4} has two versions -- one for discrete random variables, and the other for continuous random variables. We discuss the discrete random variable case first as its easier to understand and then move to the continuous random variable case. 

\subsubsection{Discrete random variables}
In this section, we assume that  in each $e\in \mathcal{E}_{all}$, random variables $Z_{\mathsf{inv}}^{e},Z_{\mathsf{spu}}^{e}, N^e, W^e $ in Assumption \ref{assumption 2_new} are discrete.  We formulate the optimization in terms of Shannon entropy as follows. 

\begin{equation}
\begin{split}
&     \min_{w\in \mathbb{R}^{k\times r}, \Phi\in \mathbb{R}^{r\times d}} \frac{1}{|\mathcal{E}_{tr}|}\sum_{e}H^e\big(w\cdot \Phi)  \\  
 & \text{s.t.} \;\; \frac{1}{|\mathcal{E}_{tr}|}\sum_{e}R^{e}\big(w\cdot \Phi \big) \leq r^{*}  \\ 
 & \;\;\;\;\;\;w \in \arg\min_{\tilde{w}\in \mathbb{R}^{k\times r}} R^e(\tilde{w}\cdot \Phi) 
\end{split}
\label{eqn: entropy_risk_min_1_appendix_disc}
\end{equation}

Note that the only difference between equation \eqref{eqn: entropy_risk_min_1_appendix_disc} and the equation \eqref{eqn: entropy_risk_min_1} is that the objective here is Shannnon entropy, while the objective in the other case is the differential entropy.

\begin{theorem}
\label{theorem4_appendix1}
 \textbf{IB-IRM and IB-ERM vs IRM and ERM}

\textbf{Fully informative invariant features (FIIF).} Suppose each $e \in \mathcal{E}_{all}$ follows 
Assumption \ref{assumption 3_new}. Assume that the invariant features are strictly separable, bounded, and satisfy support overlap (Assumptions \ref{assumption 4_new},\ref{assumption 6_new} and \ref{assumption 8_new} hold). Also, for each $e\in \mathcal{E}_{tr}$ $Z_{\mathsf{spu}}^e \leftarrow AZ_{\mathsf{inv}}^e + W^e$, where $A \in \mathbb{R}^{o\times m}$, $W^e\in \mathbb{R}^{o}$ is discrete,  bounded noise, with zero mean (and each component takes at least two distinct values).  Each solution to IB-IRM (eq. \eqref{eqn: entropy_risk_min_1}, with $\ell$ as $0$-$1$ loss, and $r^{\mathsf{th}}=q$), and IB-ERM  solves the OOD generalization (eq. \eqref{eqn1: min_max_ood}) but ERM and IRM (eq.\eqref{eqn: IRM}) fail.


\end{theorem}

In the above Theorem \ref{theorem4_appendix1}, we only state the first part of the Theorem \ref{theorem4}, the reason is that the proof of the second part proof is exactly the same in both discrete and continuous random variable case and we describe the proof for the second part in the continuous random variable section next. 

\textbf{Proof of Theorem 8.}
First, let us discuss why IRM and ERM fail in the above setting. We argue that the failure, in this case, follows directly from the second part of Theorem \ref{theorem 3}. To directly use the second part of Theorem \ref{theorem 3}, we need Assumptions \ref{assumption 3_new}-\ref{assumption 6_new} and \ref{assumption 8_new} to hold. In the statement of the above theorem, Assumption \ref{assumption 3_new}, \ref{assumption 4_new}, \ref{assumption 6_new}, and \ref{assumption 8_new} already hold. We are only required to show that Assumption \ref{assumption 5_new} holds.
 Since $Z_{\mathsf{inv}}^{e}$ and $W^{e}$ are bounded on training environments we can argue that $Z_{\mathsf{spu}}^e$ is also bounded in training environments ($\|Z_{\mathsf{spu}}^e\| \leq \|A\|Z_{\mathsf{inv}}^{e}\| + \|W^{e}\| $). We can now directly use the second part of Theorem \ref{theorem 3} because Assumptions \ref{assumption 3_new}-\ref{assumption 6_new} and \ref{assumption 8_new} hold. Since Assumption \ref{assumption 7_new} is not required to hold, both ERM and IRM will fail as their solution space continue to contain classifiers that rely on spurious features. 
 To further elaborate on why ERM and IRM fail,  recall that in the second part of Theorem \ref{theorem 3}, we relied on Lemma \ref{lemma 2}. In Lemma \ref{lemma 2}, we had shown that if latent invariant features are strictly separable, and latent spurious features are bounded, then there exist classifiers that rely on spurious features and yet are Bayes optimal on all the training environments. In this case, we have latent invariant features that are strictly separable and spurious features that are bounded, which is why we can use Theorem \ref{theorem 3}. 
 We now move to the part, where we establish why IB-IRM and IB-ERM succeed.

Consider a solution to equation \eqref{eqn: entropy_risk_min_1_appendix_disc} and call it $\Phi^{\dagger}$.  Consider the prediction made by this model 

\begin{equation}
\begin{split}
    & \Phi^{\dagger} \cdot X^e = \Phi^{\dagger} \cdot S (Z_{\mathsf{inv}}^{e}, Z_{\mathsf{spu}}^{e}) = \Phi_{\mathsf{inv}}\cdot Z_{\mathsf{inv}}^{e} + \Phi_{\mathsf{spu}}\cdot Z_{\mathsf{spu}}^{e}. 
\end{split}
\label{theorem 6 proof: eqn0}
\end{equation}

We first show that $\Phi_{\mathsf{spu}}$ is zero. We prove this by contradiction. Assume  $\Phi_{\mathsf{spu}} \not=0$ and use the condition in the theorem to simplify the expression for the prediction as follows

\begin{equation}
\begin{split}
    & \Phi_{\mathsf{inv}}\cdot Z_{\mathsf{inv}}^{e} + \Phi_{\mathsf{spu}}\cdot Z_{\mathsf{spu}}^{e}  \\ 
    & = \Phi_{\mathsf{inv}}\cdot Z_{\mathsf{inv}}^{e} + \Phi_{\mathsf{spu}}\cdot(AZ_{\mathsf{inv}}^{e} + W^{e})  \\ 
    & = \Phi_{\mathsf{inv}}\cdot Z_{\mathsf{inv}}^{e} + \Phi_{\mathsf{spu}}\cdot (AZ_{\mathsf{inv}}^{e} + W^e)  \\ 
    & = \Big[\Phi_{\mathsf{inv}} +\Phi_{\mathsf{spu}}\cdot A\Big]\cdot Z_{\mathsf{inv}}^{e} + \Phi_{\mathsf{spu}}\cdot W^e.
    \end{split}
\end{equation}
We will show that $\Phi^{+} = \Big(\Big[\Phi_{\mathsf{inv}} +\Phi_{\mathsf{spu}}\cdot A \Big], 0\Big)S^{-1} = \Big[\Phi_{\mathsf{inv}} +\Phi_{\mathsf{spu}}\cdot A \Big] S^{\dagger}_{\mathsf{inv}}$, where  $S^{\dagger}_{\mathsf{inv}}$ corresponds to the first $m$ rows of the matrix $S^{-1}$, can continue to achieve an error of $q$ and has a lower entropy than $\Phi^{\dagger}$. Recall that $\Phi^{\dagger}$ achieves an average error  across the training environments of $q$ (because $r^{\mathsf{th}}=q$ the average cannot fall below $q$ as in that case at least one environment would have a lower error than $q$ which is not possible), which implies each environment also achieves an error  of $q$.

Consider an environment $e \in \mathcal{E}_{tr}$.
Since the error  $\Phi^{\dagger}$ is $q$ it implies that  for each training environment $e$
\begin{equation}
\begin{split}
\mathsf{I}(w_{\mathsf{inv}}^{*}\cdot Z_{\mathsf{inv}}^{e})= \mathsf{I}(\Phi_{\mathsf{inv}}\cdot Z_{\mathsf{inv}}^{e} + \Phi_{\mathsf{spu}}\cdot Z_{\mathsf{spu}}^{e} )
\end{split}
\label{theorem 6 proof2: eqn1}
\end{equation}
holds over all the points in the support of environment $e$. Suppose the above claim was not true, i.e. suppose the set 
$\mathsf{I}(w_{\mathsf{inv}}^{*}\cdot Z_{\mathsf{inv}}^{e})\not= \mathsf{I}(\Phi_{\mathsf{inv}}\cdot Z_{\mathsf{inv}}^{e} + \Phi_{\mathsf{spu}}\cdot Z_{\mathsf{spu}}^{e} )$ occurs with a for some point in the support (suppose that point occurs with probability $\theta$). Let us compute the error 
\begin{equation}
\begin{split}
& R^{e}(\Phi^{\dagger}) = \mathbb{E}\Big[\big(\mathsf{I}(w_{\mathsf{inv}}^{*}\cdot Z_{\mathsf{inv}}^{e}) \oplus N^e \oplus \mathsf{I}(\Phi_{\mathsf{inv}}\cdot Z_{\mathsf{inv}}^{e} + \Phi_{\mathsf{spu}}\cdot Z_{\mathsf{spu}}^{e} )\big) \Big]  \\ 
& = \theta \mathbb{E}[1\oplus N^e] + (1-\theta)\mathbb{E}[N^e] > q
\end{split}
\end{equation}

If the above is true, then that contradicts the claim that $\Phi^{\dagger}$ achieves an error of $q$. Thus the statement in equation \eqref{theorem 6 proof2: eqn1} has to hold at all points in the training support of the invariant features. Let $\mathcal{W}^e$ be the support of $W^e$. In each training environment, if we consider a $ z_{\mathsf{inv}}^{e} \in \mathcal{Z}_{\mathsf{inv}}^{e}$, then $\forall w^{e} \in   \mathcal{W}^e$,  the following holds -- if
$\mathsf{I}(w_{\mathsf{inv}}^{*}\cdot z_{\mathsf{inv}}^{e})=1$, then

\begin{equation}
\begin{split}
&\Phi_{\mathsf{inv}}\cdot z_{\mathsf{inv}}^{e} + \Phi_{\mathsf{spu}}\cdot (Az_{\mathsf{inv}}^{e}+w^e) \geq 0 \\
\implies &\Phi_{\mathsf{inv}}\cdot z_{\mathsf{inv}}^{e} + \Phi_{\mathsf{spu}}\cdot (Az_{\mathsf{inv}}^{e}) \geq -\Phi_{\mathsf{spu}}\cdot w^e \\ 
\implies &\big(\Phi_{\mathsf{inv}} +\Phi_{\mathsf{spu}}\cdot A\big) \cdot z_{\mathsf{inv}}^{e} \geq \max_{w^{e}\in\tilde{\mathcal{W}^e}}-\Phi_{\mathsf{spu}}\cdot w^e \\ 
\implies& \big(\Phi_{\mathsf{inv}} + \Phi_{\mathsf{spu}} \cdot A\big) \cdot z_{\mathsf{inv}}^{e} \geq 0 \\ 
\implies & \Phi^{+}X^e \geq 0.
\end{split}
\label{theorem 6 proof1: eqn2}
\end{equation}

Similarly, we can argue that if $\mathsf{I}(w_{\mathsf{inv}}^{*}\cdot z_{\mathsf{inv}}^{e})=0$, then  
\begin{equation}
\begin{split}
  & \big(\Phi_{\mathsf{inv}} + \Phi_{\mathsf{spu}} \cdot A\big) \cdot z_{\mathsf{inv}}^{e} < 0 \\
    & \Phi^{+}X^e <0.
\label{theorem 6 proof1: eqn3}
\end{split}
\end{equation}

In the above simplification equation \eqref{theorem 6 proof1: eqn2}, we use $\max_{w^{e}}-\Phi_{\mathsf{spu}}\cdot w^e \geq 0$. Consider any component of $-\Phi_{\mathsf{spu}}$; if the sign of the component is positive (negative), then set the corresponding component of $w^e$ to be positive (negative). As a result, $-\Phi_{\mathsf{spu}}\cdot w^e \geq 0$. In this argument, we only relied on the assumption that $w^e$ can take both signs in the set $\mathcal{W}^e$. Suppose $\mathcal{W}^e$ had  either positive or negative values only then this would imply that the mean of $w^e$ is strictly positive or negative, which cannot be true because $W^e$ is zero mean. From equation \eqref{theorem 6 proof1: eqn2} and \eqref{theorem 6 proof1: eqn3}, we can conclude that $\Phi^{+}$ achieves the same error of $q$ in all the training environments.


Observe that we can write $\Phi^{\dagger}\cdot X^{e} = \Phi^{+}\cdot X^e + \Phi_{\mathsf{spu}}\cdot W^{e}$.
We state two properties that we use to show that entropy $\Phi^{+}$ is smaller than $\Phi^{\dagger}$: 

a) $\Phi_{\mathsf{spu}}\cdot W^{e} \perp \Phi^{+}\cdot X^e $ ($\Phi^{+}\cdot X^e=\Big[\Phi_{\mathsf{inv}} +\Phi_{\mathsf{spu}}\cdot A\Big]\cdot Z_{\mathsf{inv}}^{e}$  and $Z_{\mathsf{inv}}^{e} \perp W^e$),

b) $\Phi^{+}\cdot X$, $\Phi_{\mathsf{spu}}\cdot W^{e}$
are discrete random variables with finite support of size at least two.

We justify why b) is true in the above. $\Phi^{+}\cdot X^e$ is a bounded random variable ($Z_{\mathsf{spu}}^e$ is bounded as $Z_{\mathsf{inv}}^e$ and $W^e$ are  bounded. Thus $X^e$ is also bounded).  $\Phi^{+}\cdot X^e$ has at least two elements in its support this follows from equation \eqref{theorem 6 proof1: eqn2} and \eqref{theorem 6 proof1: eqn3}. $\Phi_{\mathsf{spu}}\cdot W^{e}$ is bounded since $W^e$ is bounded and takes at least two values because  each component of $W^e$ takes at least two distinct values.

From a), b), and Lemma \ref{lemma 5} it follows that $\Phi^{+}\cdot X^{e}$ is a classifier with lower entropy. We already established that $\Phi^{+}$  achieves the same error as $\Phi^{\dagger}$ for all the training environments. $\Phi^{+}$ achieves an error  of $q$ for all the training environments simultaneously. Since $q$ is the smallest value for the error that is achievable, the invariance constraint in equation \eqref{eqn: entropy_risk_min_2} is automatically satisfied. Therefore, $\Phi^{+}$ is strictly preferable to $\Phi^{\dagger}$. Thus the solution $\Phi^{\dagger}$ cannot rely on the spurious features and $\Phi_{\mathsf{spu}}=0$.

Thus any solution $\Phi^{\dagger}$ to  equation \eqref{eqn: entropy_risk_min_1_appendix_disc} has to satisfy $\Phi^{\dagger}\cdot S = (\Phi_{\mathsf{inv}}, 0)$ and  $\Phi^{\dagger}\cdot S$ also satisfies 
\begin{equation}
    \mathsf{I}(w_{\mathsf{inv}}^{*}\cdot Z_{\mathsf{inv}}^{e})= \mathsf{I}(\Phi_{\mathsf{inv}}\cdot Z_{\mathsf{inv}}^{e}).
    \label{theorem 6 proof: eqn5}
\end{equation}
Recall that in the second part of Theorem \ref{theorem 3}'s proof we showed that if a solution does not rely on spurious features and satisfies equation \eqref{theorem 6 proof: eqn5} for all the points in the support, then under the support overlap assumptions such a solution is OOD optimal as well. Since we assume support overlap assumption holds for the invariant features, we use the same argument from the second part of Theorem \ref{theorem 3} and  it follows that the solution to equation \eqref{eqn: entropy_risk_min_1_appendix_disc} also solves equation \eqref{eqn1: min_max_ood}. \hfill $\qedsymbol$




\subsubsection{Continuous random variables}
In this section, we assume that in each $e\in \mathcal{E}_{all}$, the random variables $Z_{\mathsf{inv}}^{e},Z_{\mathsf{spu}}^{e}, N^e, W^e $ in Assumption \ref{assumption 3_new} are continuous. 

\textbf{Lower bounding the differential entropy objective:} In general, the differential entropy can be unbounded below. Following the work of \cite{kirsch2020unpacking}, we add an independent noise term to the predictor to ensure that the entropy is lower bounded.
Suppose $w\cdot \Phi$ is the output of the predictor and the entropy of the predictor for the data in environment $e$ as $h^e(w\cdot \Phi)$. Consider a prediction made by the classifier $w\cdot \Phi(X^e)$; we add noise $\kappa^e$ (continuous, bounded random variable with a finite entropy) to this prediction to get $w \cdot \Phi(X) + \kappa^e$. The  differential entropy after noise addition as $ h^{e}(w\cdot \Phi(X^e) + \kappa^e)$. Observe that $h^{e}(w\cdot \Phi(X^e) + \kappa^e) \geq h(\kappa^e)$. In the rest of the discussion, we just write $h^{e}(w\cdot \Phi(X^e) + \kappa^e)$ as $h^{e}(w\cdot \Phi)$ to make the notation less cumbersome. We constrain $\mathcal{H}_{\Phi}$ ($\mathcal{H}_{w}$) in the optimization in equation \eqref{eqn: entropy_risk_min_1} to a set $\tilde{\mathcal{H}}_{\Phi} = \{\Phi \in \mathbb{R}^{r \times d}\;|\; 0<\phi_{\mathsf{inf}}\leq \|\Phi\| \leq \phi_{\mathsf{sup}}\}$ ($\tilde{\mathcal{H}}_{w} = \{w \in \mathbb{R}^{k \times r}\;|\; 0<w_{\mathsf{inf}}\leq \|w\| \leq w_{\mathsf{sup}}\}$) instead of $\mathcal{H}_{\Phi} = \mathbb{R}^{r \times d}$ ($\mathcal{H}_{w} = \mathbb{R}^{k \times r}$). The reason to do this is that while the $0$-$1$ loss does not change with scaling of the predictor but the entropy can change a lot. The lower bound on the norm of the classifier ensures that the optimization does not shrink it to zero in trying to minimize the entropy. We restate the optimization in equation \eqref{eqn: entropy_risk_min_1}  after accounting for the pathologies of differential entropy that we described above:   
\begin{equation}
\begin{split}
&     \min_{w\in \tilde{\mathcal{H}}_{w}, \Phi \in \tilde{\mathcal{H}}_{\Phi}} \frac{1}{|\mathcal{E}_{tr}|}\sum_{e}h^e\big(w\cdot \Phi)  \\  
 & \text{s.t.} \;\; \frac{1}{|\mathcal{E}_{tr}|}\sum_{e}R^{e}\big(w\cdot \Phi \big) \leq r^{\mathsf{th}}  \\ 
 & \;\;\;\;\;\;w \in \arg\min_{\tilde{w}\in \tilde{\mathcal{H}}_{w}} R^e(\tilde{w}\cdot \Phi) 
\end{split}
\label{eqn: entropy_risk_min_1_appendix}
\end{equation}
 
  We restate Theorem \ref{theorem4} for convenience.




\begin{theorem}
\label{theorem4_appendix}
 \textbf{IB-IRM and IB-ERM vs IRM and ERM}
 
 $\bullet$ \textbf{Fully informative invariant features (FIIF).} Suppose each $e \in \mathcal{E}_{all}$ follows 
Assumption \ref{assumption 3_new}. Assume that the invariant features are strictly separable, bounded, and satisfy support overlap (Assumptions \ref{assumption 4_new},\ref{assumption 6_new} and \ref{assumption 8_new} hold). Also, for each $e\in \mathcal{E}_{tr}$ $Z_{\mathsf{spu}}^e \leftarrow AZ_{\mathsf{inv}}^e + W^e$, where $A \in \mathbb{R}^{o\times m}$, $W^e\in \mathbb{R}^{o}$ is continuous, bounded, and zero mean noise.  Each solution to IB-IRM (eq. \eqref{eqn: entropy_risk_min_1}, with $\ell$ as $0$-$1$ loss, and $r^{\mathsf{th}}=q$), and IB-ERM  solves the OOD generalization (eq. \eqref{eqn1: min_max_ood}) but ERM and IRM (eq.\eqref{eqn: IRM}) fail.

$\bullet$ \textbf{Partially informative invariant features (PIIF).} Suppose each $e \in \mathcal{E}_{all}$ follows 
Assumption \ref{assumption 1_new} and $\exists\; e \in \mathcal{E}_{tr}$ such that  $ \mathbb{E}[\epsilon^eZ_{\mathsf{spu}}^e]\not=0$. If $|\mathcal{E}_{tr}|>2d$ and the set $\mathcal{E}_{tr}$ lies in a linear general position (a mild condition defined in the Appendix), then each solution to IB-IRM (eq. \eqref{eqn: entropy_risk_min_1}, with $\ell$ as square loss, $\sigma_{\epsilon}^{2} <r^{\mathsf{th}} \leq \sigma_{Y}^{2}$, where $\sigma_{Y}^{2}$ and $\sigma_{\epsilon}^2$ are the variance in the label and noise across $\mathcal{E}_{tr}$) and IRM (eq.\eqref{eqn: IRM}) solves  OOD generalization  (eq. \eqref{eqn1: min_max_ood}) but IB-ERM and ERM fail. 

\end{theorem}

\textbf{Proof of Theorem 9.}  First, let us discuss why IRM and ERM fail in the above setting. We argue that the failure, in this case, follows directly from the second part of Theorem \ref{theorem 3}. To directly use the second part of Theorem \ref{theorem 3}, we need Assumptions \ref{assumption 3_new}-\ref{assumption 6_new} and \ref{assumption 8_new} to hold. In the statement of the above theorem, Assumption \ref{assumption 3_new}, \ref{assumption 4_new}, \ref{assumption 6_new}, and \ref{assumption 8_new} already hold. We are only required to show that Assumption \ref{assumption 5_new} holds.
 Since $Z_{\mathsf{inv}}^{e}$ and $W^{e}$ are bounded on training environments we can argue that $Z_{\mathsf{spu}}^e$ is also bounded in training environments ($\|Z_{\mathsf{spu}}^e\| \leq \|A\|Z_{\mathsf{inv}}^{e}\| + \|W^{e}\| $). We can now directly use the second part of Theorem \ref{theorem 3} because Assumptions \ref{assumption 3_new}-\ref{assumption 6_new} and \ref{assumption 8_new} hold. Since Assumption \ref{assumption 7_new} is not required to hold, both ERM and IRM will fail as their solution space continue to contain classifiers that rely on spurious features. \footnote{In the remark following the proof of Theorem \ref{theorem 3}, we had discussed the failure of ERM and IRM continues to hold even when we are restricted to use continuous random variables.}


Consider a solution to IB-IRM (eq. \eqref{eqn: entropy_risk_min_1_appendix}) and call it $\Phi^{\dagger}$.  Consider the prediction made by this model 

\begin{equation}
\begin{split}
    & \Phi^{\dagger} \cdot X^e = \Phi^{\dagger} \cdot S (Z_{\mathsf{inv}}^{e}, Z_{\mathsf{spu}}^{e}) = \Phi_{\mathsf{inv}}\cdot Z_{\mathsf{inv}}^{e} + \Phi_{\mathsf{spu}}\cdot Z_{\mathsf{spu}}^{e}. 
\end{split}
\label{theorem 6 proof: eqn0}
\end{equation}

We first show that $\Phi_{\mathsf{spu}}$ is zero. We prove this by contradiction. Assume  $\Phi_{\mathsf{spu}} \not=0$ and use the condition in the theorem to simplify the expression for the prediction as follows.

\begin{equation}
\begin{split}
    & \Phi_{\mathsf{inv}}\cdot Z_{\mathsf{inv}}^{e} + \Phi_{\mathsf{spu}}\cdot Z_{\mathsf{spu}}^{e}  \\ 
    & = \Phi_{\mathsf{inv}}\cdot Z_{\mathsf{inv}}^{e} + \Phi_{\mathsf{spu}}\cdot(AZ_{\mathsf{inv}}^{e} + W^{e})  \\ 
    & = \Phi_{\mathsf{inv}}\cdot Z_{\mathsf{inv}}^{e} + \Phi_{\mathsf{spu}}\cdot (AZ_{\mathsf{inv}}^{e} + W^e)  \\ 
    & = \Big[\Phi_{\mathsf{inv}} +\Phi_{\mathsf{spu}}\cdot A\Big]\cdot Z_{\mathsf{inv}}^{e} + \Phi_{\mathsf{spu}}\cdot W^e.
    \end{split}
\end{equation}
We will show that $\Phi^{+} = \Big(\Big[\Phi_{\mathsf{inv}} +\Phi_{\mathsf{spu}}\cdot A \Big], 0\Big)S^{-1} = \Big[\Phi_{\mathsf{inv}} +\Phi_{\mathsf{spu}}\cdot A \Big] S^{\dagger}_{\mathsf{inv}}$, where  $S^{\dagger}_{\mathsf{inv}}$ corresponds to the first $m$ rows of the matrix $S^{-1}$, can continue to achieve an error of $q$ and has a lower entropy than $\Phi^{\dagger}$. Recall that $\Phi^{\dagger}$ achieves an average error across the training environments of $q$ (because $r^{\mathsf{th}}=q$ the average cannot fall below $q$ as in that case at least one environment would have a lower error than $q$ which is not possible), which implies each environment also achieves an error  of $q$.

Consider an environment $e \in \mathcal{E}_{tr}$.
Since the error $\Phi^{\dagger}$ is $q$ it implies that  for each training environment
\begin{equation}
\begin{split}
\mathsf{I}(w_{\mathsf{inv}}^{*}\cdot Z_{\mathsf{inv}}^{e})= \mathsf{I}(\Phi_{\mathsf{inv}}\cdot Z_{\mathsf{inv}}^{e} + \Phi_{\mathsf{spu}}\cdot Z_{\mathsf{spu}}^{e} ),
\end{split}
\label{theorem 6 proof: eqn1}
\end{equation}
holds with probability $1$. 
Suppose the above claim was not true, i.e. suppose the set 
$\mathsf{I}(w_{\mathsf{inv}}^{*}\cdot Z_{\mathsf{inv}}^{e})\not= \mathsf{I}(\Phi_{\mathsf{inv}}\cdot Z_{\mathsf{inv}}^{e} + \Phi_{\mathsf{spu}}\cdot Z_{\mathsf{spu}}^{e} )$ occurs with a non-zero probability say $\theta$. Let us compute the error 
\begin{equation}
\begin{split}
& R^{e}(\Phi^{\dagger}) = \mathbb{E}\Big[\big(\mathsf{I}(w_{\mathsf{inv}}^{*}\cdot Z_{\mathsf{inv}}^{e}) \oplus N^e \oplus \mathsf{I}(\Phi_{\mathsf{inv}}\cdot Z_{\mathsf{inv}}^{e} + \Phi_{\mathsf{spu}}\cdot Z_{\mathsf{spu}}^{e} )\big) \Big]  \\ 
& = \theta \mathbb{E}[1\oplus N^e] + (1-\theta)\mathbb{E}[N^e] > q
\end{split}
\end{equation}
If the above is true, then that contradicts the claim that $\Phi^{\dagger}$ achieves an error of $q$. Thus the statement in equation \eqref{theorem 6 proof: eqn1} has to hold with probability $1$. Let $\mathcal{W}^{e}$ denote the support of $W^e$ in environment $e$. 
We can restate the above observation as -- there exists sets $\tilde{\mathcal{Z}}_{\mathsf{inv}}^{e} \subseteq \mathcal{Z}_{\mathsf{inv}}^{e} $ and a set $\tilde{\mathcal{W}}^{e} \subseteq \mathcal{W}^{e}$ such that $\mathbb{P}(\tilde{\mathcal{Z}}_{\mathsf{inv}}^{e} \times \tilde{\mathcal{W}}^{e})=1$ \footnote{Owing to the independence of the noise we also have $\mathbb{P}(\tilde{\mathcal{Z}}_{\mathsf{inv}}^{e})=1$, $\mathbb{P}(\tilde{\mathcal{W}}^{e})=1$.} and for each element in $\tilde{\mathcal{Z}}_{\mathsf{inv}}^{e}\times \tilde{\mathcal{W}}^{e}$ 
\begin{equation}
    \mathsf{I}(w_{\mathsf{inv}}^{*}\cdot Z_{\mathsf{inv}}^{e})= \mathsf{I}(\Phi_{\mathsf{inv}}\cdot Z_{\mathsf{inv}}^{e} + \Phi_{\mathsf{spu}}\cdot Z_{\mathsf{spu}}^{e} )
\end{equation}

Consider a training environment $e\in \mathcal{E}_{tr}$. For each $ z_{\mathsf{inv}}^{e} \in \tilde{\mathcal{Z}}_{\mathsf{inv}}^{e}$, the following conditions hold $\forall w^{e} \in   \tilde{\mathcal{W}}^e$  -- 
if $\mathsf{I}(w_{\mathsf{inv}}^{*}\cdot z_{\mathsf{inv}}^{e})=1$, then

\begin{equation}
\begin{split}
&\Phi_{\mathsf{inv}}\cdot z_{\mathsf{inv}}^{e} + \Phi_{\mathsf{spu}}\cdot (Az_{\mathsf{inv}}^{e}+w^e) \geq 0 \\
\implies &\Phi_{\mathsf{inv}}\cdot z_{\mathsf{inv}}^{e} + \Phi_{\mathsf{spu}}\cdot (Az_{\mathsf{inv}}^{e}) \geq -\Phi_{\mathsf{spu}}\cdot w^e \\ 
\implies &\big(\Phi_{\mathsf{inv}} +\Phi_{\mathsf{spu}}\cdot A\big) \cdot z_{\mathsf{inv}}^{e} \geq \max_{w^{e}\in\tilde{\mathcal{W}^e}}-\Phi_{\mathsf{spu}}\cdot w^e \\ 
\implies& \big(\Phi_{\mathsf{inv}} + \Phi_{\mathsf{spu}} \cdot A\big) \cdot z_{\mathsf{inv}}^{e} \geq 0 \\ 
\implies & \Phi^{+}X^e \geq 0. 
\end{split}
\label{theorem 6 proof: eqn2}
\end{equation}

Similarly, we can argue that if $\mathsf{I}(w_{\mathsf{inv}}^{*}\cdot z_{\mathsf{inv}}^{e})=0$, then  
\begin{equation}
\begin{split}
  & \big(\Phi_{\mathsf{inv}} + \Phi_{\mathsf{spu}} \cdot A\big) \cdot z_{\mathsf{inv}}^{e} < 0 \\
    & \Phi^{+}X^e <0.
\label{theorem 6 proof: eqn3}
\end{split}
\end{equation}

In the above simplification in equation \eqref{theorem 6 proof: eqn2}, we use $\max_{w^{e}}-\Phi_{\mathsf{spu}}\cdot w^e \geq 0$. Consider any component of $-\Phi_{\mathsf{spu}}$; if the sign of the component is positive (negative), then set the corresponding component of $w^e$ to be positive (negative). As a result, $-\Phi_{\mathsf{spu}}\cdot w^e \geq 0$. In this argument, we only relied on the assumption that $w^e$ can take both signs in the set $\tilde{\mathcal{W}^e}$. Suppose $w^e$ can only take either positive or negative values in $\tilde{\mathcal{W}^e}$ this would imply that the mean of $w^e$ is strictly positive or negative, which cannot be true because $W^e$ is zero mean. From equation \eqref{theorem 6 proof: eqn2}, \eqref{theorem 6 proof: eqn3}, and $\mathbb{P}(\tilde{\mathcal{Z}}_{\mathsf{inv}}^{e} \times \tilde{\mathcal{W}}^{e})=1$, we can conclude that $\Phi^{+}$ achieves the same error of $q$ in all the training environments.

Observe that we can write $\Phi^{\dagger}\cdot X^{e} = \Phi^{+}\cdot X^e + \Phi_{\mathsf{spu}}\cdot W^{e}$.
We state two properties that we use to show that entropy $\Phi^{+}$ is smaller than $\Phi^{\dagger}$:

a) $\Phi_{\mathsf{spu}}\cdot W^{e} \perp \Phi^{+}\cdot X^e $ ($\Phi^{+}\cdot X^e=\Big[\Phi_{\mathsf{inv}} +\Phi_{\mathsf{spu}}\cdot A\Big]\cdot Z_{\mathsf{inv}}^{e}$  and $Z_{\mathsf{inv}}^{e} \perp W^e$),

b) $\Phi^{+}_{\mathsf{inv}}\cdot X$, $\Phi_{\mathsf{spu}}\cdot W^{e}$
are continuous bounded random variables, 

We justify why b) is true in the above. $\Phi^{+}_{\mathsf{inv}}\cdot X^e$ is a bounded random variable ($Z_{\mathsf{spu}}^e$ is bounded as $Z_{\mathsf{inv}}^e$ is bounded and as a result $X^e$ is bounded as well).  
Observe that $\Phi^{+}_{\mathsf{inv}}\not=0$, this follows from equation \eqref{theorem 6 proof: eqn2} and \eqref{theorem 6 proof: eqn3}. $\Phi^{+}_{\mathsf{inv}}\cdot X^e$ is a continuous random variable as well. Suppose $\Phi^{+}_{\mathsf{inv}}\cdot X^e$ was not continuous, which implies for some constant $b$, $\Phi^{+}_{\mathsf{inv}}\cdot X^e=b$ with a finite probability. If $\Phi^{+}_{\mathsf{inv}}\cdot X^e=b$ with a finite probability, then $X$ cannot be a continuous random vector (as there exists a hyperplane which occurs with a non-zero probability).

From a), b), and Lemma \ref{lemma 6} it follows that 
\begin{equation}
h^{e}(\Phi^{+}\cdot X^{e}) < h^{e}(\Phi^{\dagger}\cdot X^{e}) 
\label{theorem 6 proof: eqn4}
\end{equation}

Note that the above equation \eqref{theorem 6 proof: eqn4} is true independent of whether we added a bounded noise to keep the entropy bounded from below. Therefore, so far we have established that $\Phi^{+}$ is a classifier with lower entropy and the same error as $\Phi^{\dagger}$. Observe that $\Phi^{+}$ achieves an error of $q$ for all the training environments simultaneously. Since $q$ is the smallest value for the error  that is achievable, the invariance constraint in equation \eqref{eqn: entropy_risk_min_2} is automatically satisfied with $\Phi^{\dagger}$ as the classifier and the representation as the identity. Thus $\Phi^{+}$ is a strictly preferable solution $\Phi^{\dagger}$, which contradicts the optimality of $\Phi^{\dagger}$. Therefore, it follows that  $\Phi_{\mathsf{spu}}=0$

Thus any solution $\Phi^{\dagger}$ to  equation \eqref{eqn: entropy_risk_min_1_appendix} has to satisfy $\Phi^{\dagger}\cdot S = (\Phi_{\mathsf{inv}}, 0)$ and  $\Phi^{\dagger}\cdot S$ also satisfies 
\begin{equation}
    \mathsf{I}(w_{\mathsf{inv}}^{*}\cdot Z_{\mathsf{inv}}^{e})= \mathsf{I}(\Phi_{\mathsf{inv}}\cdot Z_{\mathsf{inv}}^{e})
    \label{theorem 6 proof: eqn5}
\end{equation}
with probability one.  From the second part of Theorem \ref{theorem 3}'s proof we know if a solution satisfies two properties a) does not rely on spurious features, and  b) satisfies equation \eqref{theorem 6 proof: eqn5} for all the points in the support, then under the support overlap of invariant features such a solution is OOD optimal (solves equation \eqref{eqn1: min_max_ood}) as well. In this case, we have also assumed support overlap assumption holds for the invariant features. We have established that the solution does not rely on spurious features. Also, we have shown that equation \eqref{theorem 6 proof: eqn5} holds not pointwise but with probability one.
We can still use the same argument from the second part of Theorem \ref{theorem 3} and  it follows that the solution to equation \eqref{eqn: entropy_risk_min_1_appendix} also solves equation \eqref{eqn1: min_max_ood}.   Next, we show why it suffices for the equation \eqref{theorem 6 proof: eqn5} to hold with probability one.

Since the equation \eqref{theorem 6 proof: eqn5} does not hold pointwise at all the points in the support and can be violated over a set of probability zero we need to be careful about some pathological shifts at test time that place a finite mass in the region where equation \eqref{eqn1: min_max_ood} is violated. We now argue using arguments based on standard measure theory \citep{ash2000probability} that such pathological shifts cannot occur under the assumptions made in this setting. 

Recall that we defined $\tilde{\mathcal{Z}}_{\mathsf{inv}}^{e}\times \tilde{\mathcal{W}}^{e}$ to be the set where equation \eqref{theorem 6 proof: eqn5} holds pointwise.  $\mathbb{P}(\tilde{\mathcal{Z}}_{\mathsf{inv}}^{e} \times \tilde{\mathcal{W}}^{e})=1$. Owing to the independence $Z^e \perp W^e$, we have $\mathbb{P}(\tilde{\mathcal{Z}}_{\mathsf{inv}}^{e})=1$, $\mathbb{P}(\tilde{\mathcal{W}}^{e})=1$.    It can be shown that the Lebesgue measure $\mu$ of the set $\mathcal{Z}_{\mathsf{inv}}^{e}\setminus\tilde{\mathcal{Z}}_{\mathsf{inv}}^{e}$ is zero, i.e., $\mu(\mathcal{Z}_{\mathsf{inv}}^{e}\setminus\tilde{\mathcal{Z}}_{\mathsf{inv}}^{e})=0$. If the Lebesgue measure was positive, i.e.,  $\mu(\mathcal{Z}_{\mathsf{inv}}^{e}\setminus\tilde{\mathcal{Z}}_{\mathsf{inv}}^{e})>0$, then the probability of this set would also be non-zero, i.e., $\mathbb{P}(\mathcal{Z}_{\mathsf{inv}}^{e}\setminus\tilde{\mathcal{Z}}_{\mathsf{inv}}^{e})>0$. The main insight to show this follows from the observation that the probability density is positive on the set $\mathcal{Z}_{\mathsf{inv}}^{e}\setminus\tilde{\mathcal{Z}}_{\mathsf{inv}}^{e}$ since the set is part of the support of $Z_{\mathsf{inv}}^{e}$. 

A formal argument to show $\mu(\mathcal{Z}_{\mathsf{inv}}^{e}\setminus\tilde{\mathcal{Z}}_{\mathsf{inv}}^{e})>0 \implies \mathbb{P}(\mathcal{Z}_{\mathsf{inv}}^{e}\setminus\tilde{\mathcal{Z}}_{\mathsf{inv}}^{e})>0$ goes as follows. 

Assume the contrary, i.e., $\mathbb{P}(\mathcal{Z}_{\mathsf{inv}}^{e}\setminus\tilde{\mathcal{Z}}_{\mathsf{inv}}^{e})=0$. Let the density be denoted as  $f_{Z_{\mathsf{inv}}^e}$.  Define the set $\mathcal{P}_k = \{z_{\mathsf{inv}} \in \mathcal{Z}_{\mathsf{inv}}^{e}\setminus\tilde{\mathcal{Z}}_{\mathsf{inv}}^{e}\;|\;f_{Z_{\mathsf{inv}}^e}(z) > \frac{1}{k} \}$.
\begin{equation}
    \mathcal{Z}_{\mathsf{inv}}^{e}\setminus\tilde{\mathcal{Z}}_{\mathsf{inv}}^{e} = \cup_{k=1}^{\infty} \mathcal{P}_k
\end{equation}
 $\mathcal{P}_k \uparrow  \mathcal{Z}_{\mathsf{inv}}^{e}\setminus\tilde{\mathcal{Z}}_{\mathsf{inv}}^{e} \implies \mu(\mathcal{P}_k) \rightarrow \mu(\mathcal{Z}_{\mathsf{inv}}^{e}\setminus\tilde{\mathcal{Z}}_{\mathsf{inv}}^{e})$. Since $\mu(\mathcal{Z}_{\mathsf{inv}}^{e}\setminus\tilde{\mathcal{Z}}_{\mathsf{inv}}^{e})>0$, $\exists $ some $s$ for which $\mu(\mathcal{P}_s)>0$.

Define $g_s$ 
\begin{equation}
g_s(x) =     \begin{cases}
    \frac{1}{s} \; \text{if} \;x \in \mathcal{P}_k \\ 
    0 \; \text{otherwise}
    \end{cases}
\end{equation}
\begin{equation}
    \mathbb{P}(\mathcal{Z}_{\mathsf{inv}}^{e}\setminus\tilde{\mathcal{Z}}_{\mathsf{inv}}^{e}) = \int_{\mathcal{Z}_{\mathsf{inv}}^{e}\setminus\tilde{\mathcal{Z}}_{\mathsf{inv}}^{e}}  f_{Z_{\mathsf{inv}}^e} d\mu \geq \int_{\mathcal{Z}_{\mathsf{inv}}^{e}\setminus\tilde{\mathcal{Z}}_{\mathsf{inv}}^{e}} g_s d\mu \geq \frac{1}{s}\mu(\mathcal{P}_s)>0
\end{equation}

 $\mu(\mathcal{Z}_{\mathsf{inv}}^{e}\setminus\tilde{\mathcal{Z}}_{\mathsf{inv}}^{e})>0 \implies \mathbb{P}(\mathcal{Z}_{\mathsf{inv}}^{e}\setminus\tilde{\mathcal{Z}}_{\mathsf{inv}}^{e})>0 \implies \mathbb{P}(\tilde{\mathcal{Z}}_{\mathsf{inv}}^{e})<1$ which is a contradiction.  Therefore, $\mu(\mathcal{Z}_{\mathsf{inv}}^{e}\setminus\tilde{\mathcal{Z}}_{\mathsf{inv}}^{e})=0$.


We now describe how our assumptions already eliminate the possibility of distribution shifts that happen in such a way that the a finite mass of the distribution resides in the region $\mathcal{Z}_{\mathsf{inv}}^{e}\setminus\tilde{\mathcal{Z}}_{\mathsf{inv}}^{e}$. Recall we assume that $\forall e \in \mathcal{E}_{all}$, $Z_{\mathsf{inv}}^{e}$ is a continuous random variable. Since the probability of continuous random is absolutely continuous w.r.t the Lebesgue measure it follows that for each $e\in \mathcal{E}_{all}$,  $\mu(\mathcal{Z}_{\mathsf{inv}}^{e}\setminus\tilde{\mathcal{Z}}_{\mathsf{inv}}^{e})=0 \implies \mathbb{P}(\mathcal{Z}_{\mathsf{inv}}^{e}\setminus\tilde{\mathcal{Z}}_{\mathsf{inv}}^{e})=0 $. Thus all distribution shifts would place a zero mass in the region of disagreement.




This completes the first part of the proof. 


The second part of the theorem follows directly from the analysis of linear regression SEM in \cite{arjovsky2019invariant}.  The conditions in the second part of the theorem cover the conditions that are required in Theorem \ref{thm9: arjovsky}. Under those conditions there can be two invariant predictors one is the trivial invariant predictor that maps every input to zero. The other is the ideal invariant predictor that focuses on the causes. The constraint $r^{\mathsf{th}}$ is set to a low enough value such that only the ideal invariant predictor gets selected. 
Observe that the risk achieved by the trivial zero invariant predictor is $\frac{1}{|\mathcal{E}_{tr}|}E[(Y^e)^2]=\sigma_Y^2$ and the risk achieved by the ideal $\frac{1}{|\mathcal{E}_{tr}|}E[(N^e)^2]=\sigma_N^2$.  If $\sigma_N^2<r^{\mathsf{th}}<\sigma_Y^2$, then the only predictor that is selected is the ideal invariant predictor. 

We now describe why ERM fails in this case. In the theorem, we assume that $\exists\; e$ where $ v= \mathbb{E}[\epsilon^eZ_{\mathsf{spu}}^e] \not=0$, which implies $ \mathbb{E}[\epsilon^eX^e]\not=0$. We show why this is the case next.  
\begin{equation}
\begin{split}
    \mathbb{E}[\epsilon^eX^e] =     \mathbb{E}[\epsilon^eS (Z_{\mathsf{inv}}^e, Z_{\mathsf{spu}}^e )] =  \mathbb{E}[S\epsilon^e (Z_{\mathsf{inv}}^e, Z_{\mathsf{spu}}^e )] = S(0,v)  \not=0 ; \text{since} \;S\; \text{is invertible}
\end{split}
\end{equation}
The rest of the proof follows from Proposition 17 in \citep{ahuja2020empirical}. If $r^{\mathsf{th}}$ is set low enough to assume the same risk achieved by ERM, then IB-ERM and ERM are identical and IB-ERM also fails. 

\hfill $\qedsymbol$

\textbf{Remark on invertibility of $S$.} The entire proof extends to the case when $S$ is not invertible but $Z_{\mathsf{inv}}^{e}$ can still be recovered. Note that at no point in the proof we required 
to have full $S$ to be invertible. 

\textbf{Remark on regularized ERM, IRM.} Note that while we showed that the ERM and IRM fail, the failures extend to $\ell_1$ or $\ell_2$ regularized models as well. 
We would like to also mention that it may seem that information bottleneck and sparsity constraints such as $\ell_1$ have similarity. We want to point out that there is a major difference between the two.
In our model, we observe scrambled data. As a result, even if there is sparsity in the latent space, that does not translate to the observed space. $\ell_1$ constraints operate in the input space and that is why they cannot fetch the same outcome as information bottleneck constraints.

\textbf{Remark on multi-class classification.} The proof presented in this section extends to multi-class setting described in Assumption \ref{multi-class}. The simplification in equation \eqref{theorem 6 proof1: eqn2} along with the lemmas (Lemma \ref{lemma 4}, Lemma \ref{lemma 5})  help establish why low-entropy representation based classifier discourages the use of spurious features. We can adapt the analysis in equation \eqref{theorem 6 proof1: eqn2} to the multi-class case (Assumption \ref{multi-class}) and follow the same line of reasoning to justify why IB-IRM and IB-ERM succeed.

\clearpage
\subsection{Derivation of the final objective in equation \eqref{eqn:finalobjective}}
\label{secn:loss_eqn7}
In this section, we give a step-by-step description of derivation of the objective in equation \eqref{eqn:finalobjective}. 
We rewrite the IB-IRM optimization below in equation \eqref{eqn: entropy_risk_min_2_uc}. 

\begin{equation}
\begin{split}
&     \min_{\Phi \in \mathbb{R}^{k}} \frac{1}{|\mathcal{E}_{tr}|}\sum_{e}h^e\big( w\cdot \Phi\big)  \\  
 & \text{s.t.} \;\; \frac{1}{|\mathcal{E}_{tr}|}\sum_{e}R^{e}\big(w\cdot \Phi \big) \leq r^{\mathsf{th}},  \\ 
 & \;\;\;\;\;\;1 \in \arg\min_{\tilde{w}\in \mathbb{R}} R^e(\tilde{w}\cdot \Phi).
\end{split}
\label{eqn: entropy_risk_min_2_uc}
\end{equation}

In the above we assumed that the classifiers are scalar. We state a new  optimization that we show is equivalent to the optimization in equation \eqref{eqn: entropy_risk_min_2_uc}.

\begin{equation}
\begin{split}
&     \min_{\Phi \in \mathbb{R}^{k}} \frac{1}{|\mathcal{E}_{tr}|}\sum_{e}h^e\big( \Phi\big)  \\  
 & \text{s.t.} \;\; \frac{1}{|\mathcal{E}_{tr}|}\sum_{e}R^{e}\big( \Phi \big) \leq r^{\mathsf{th}},  \\ 
 & \;\;\;\;\;\;1 \in \arg\min_{\tilde{w}\in \mathbb{R}} R^e(\tilde{w}\cdot \Phi). 
\end{split}
\label{eqn: entropy_risk_min_2}
\end{equation}

It can be shown that the two forms of optimization in equation \eqref{eqn: entropy_risk_min_2_uc} and equation \eqref{eqn: entropy_risk_min_2} are equivalent. First, we would like to show that the set of feasible classifiers $w\cdot \Phi$ for the first optimization in equation \eqref{eqn: entropy_risk_min_2} and $\Phi$ in the second optimization
in equation \eqref{eqn: entropy_risk_min_2} are the same. 

Suppose $w^{*}, \Phi^{*}$ is a feasible solution to the constraints in equation \eqref{eqn: entropy_risk_min_2_uc}. Construct $\Phi^{\dagger} = w^{*}\cdot \Phi^{*}$. $\Phi^{\dagger}$ satisfies the constraint $\frac{1}{|\mathcal{E}_{tr}|}\sum_{e}R^{e}\big( \Phi^{\dagger} \big) \leq r^{\mathsf{th}}$. Suppose for some environment $e$, $1\not \in \arg\min_{\tilde{w}} R^e(\tilde{w}\cdot \Phi^{\dagger}) \implies \exists\; w \not=1$ such that $R^e(w\cdot \Phi^{\dagger}) < R^e(\Phi^{\dagger})$. If this is the case, then $w\times w^{*}$ improves over $w^{*}$ and contradicts the optimality of $w^{*}$ in equation \eqref{eqn: entropy_risk_min_2_uc}. This establishes that $\Phi^{\dagger}$ satisfies the constraints in equation \eqref{eqn: entropy_risk_min_2_uc}. 
This shows that the set of feasible classifiers for the first optimization in equation \eqref{eqn: entropy_risk_min_2_uc} are a subset of the feasible classifiers in the second optimization \eqref{eqn: entropy_risk_min_2}.

Suppose $\Phi^{*}$ is a feasible solution to the constraints in equation \eqref{eqn: entropy_risk_min_2}. Take any scalar $w$ and corresponding representation $\Phi^{*}/w$. The combined classifier $w\cdot (\Phi^{*}/w)$ satisfies the first constraint. Suppose $w \not \in \arg\min_{\tilde{w}\in \mathbb{R}} R^e(\tilde{w}\cdot \frac{\Phi}{w})$, this implies that $\exists \; w^{+}\not=w $ such that $R^e({\frac{w^{+}}{w} \cdot \Phi^{*}})<R^{e}(\Phi^{*})$. If this was true, then that contradicts the optimality of 1 in equation \eqref{eqn: entropy_risk_min_2}. 
This shows that the set of feasible classifiers for the second optimization in equation \eqref{eqn: entropy_risk_min_2} are a subset of the feasible classifiers in the first optimization \eqref{eqn: entropy_risk_min_2_uc}. 

From the above discussion, it is clear that the two formulations result in the same set of feasible $w\cdot \Phi$, which are finally fed into the same entropy minimization objective. Thus the two optimizations are equivalent.  To get to the penalized objective in equation \eqref{eqn:finalobjective} from the equation \eqref{eqn: entropy_risk_min_2} there are two key steps:  i) converting the invariance constraint into the gradient-based penalty, i.e., the IRMv1 penalty from \citep{arjovsky2019invariant}, ii) converting the differential entropy term into a constraint on the variance. For ii), as we explained in the manuscript, minimization of variance is equivalent to minimizing an upper bound on the entropy. Also, note that since variance has a lower bound, we can directly work with $\Phi$ and do not need to add a noise term like earlier, which was done to ensure that differential entropy is lower bounded. Below we break down the steps to arrive at the objective. We first start with a weighted combination of the terms in equation \eqref{eqn: entropy_risk_min_1}.

\begin{equation}
    \sum_{e}\Big(R^{e}(\Phi) + \lambda \|\nabla_{w, w=1.0} R^{e}(w \cdot \Phi)\|^2 + \nu h^e(\Phi) \Big). \label{eqn:finalobjective1}
\end{equation}

where $ \|\nabla_{w, w=1.0} R^{e}(w \cdot \Phi)\|^2$ is the norm of the gradient computed w.r.t scalar classifier $w$ at $1.0$. Note that in general the gradient can be computed w.r.t a fixed vector as well.  In our experiments, we found that using entropy conditioned on the environment or entropy unconditioned on the environment works equally well. Thus, we introduce the unconditional entropy $h(\Phi)$. We assume that all the environments occur with an equal probability.  

\begin{equation}
    h(\Phi) =  -\mathbb{E}_{X\sim \mathbb{P}}[\log(d\mathbb{P}(\Phi(X))] 
\end{equation}
where $d\mathbb{P}(\Phi(X))$ is the probability density of predictions (unconditional on the environment), $\mathbb{P} = \frac{1}{|\mathcal{E}_{tr}|}\sum_{e\in \mathcal{E}_{tr}}\mathbb{P}^e$ is the uniform mixture of data from all environments. Note here $X$ denotes an input sample and we do not know the environment it comes from unlike the sample $X^e$. The entropy of predictions computed in environment $e$ is given as

\begin{equation}
    h^{e}(\Phi) =  -\mathbb{E}_{X^{e}\sim \mathbb{P}^e}[\log(d\mathbb{P}^e(\Phi(X^e))],
\end{equation}
where $d\mathbb{P}^e$ is the probability density of the predictions in environment $e$. The conditional entropy over predictions conditioned on a random environment is given as 

\begin{equation}
    h(\Phi|E) =  -\frac{1}{|\mathcal{E}_{tr}|}\sum_{e\in \mathcal{E}_{tr}}\mathbb{E}[\log(d\mathbb{P}^e(\Phi(X^e))].
\end{equation}
Conditioning reduces entropy $ h(\Phi)\geq  h(\Phi|E)$ and thus we propose an upper bound on the objective in equation \eqref{eqn:finalobjective1} below

\begin{equation}
    \sum_{e}\Big(R^{e}(\Phi) + \lambda \|\nabla_{w, w=1.0} R^{e}(w \cdot \Phi)\|^2 + \nu h(\Phi) \Big). \label{eqn:finalobjective2}
\end{equation}

Finally, instead of $h(\Phi)$ we use variance in predictions $\Phi$ denoted as $\mathsf{Var}(\Phi) = \mathbb{E}_{X\sim \mathbb{P}}[(\Phi(X)-\mathbb{E}[\Phi(X)])^2]$ to get 

\begin{equation}
    \sum_{e}\Big(R^{e}(\Phi) + \lambda \|\nabla_{w, w=1.0} R^{e}(w \cdot \Phi)\|^2 + \gamma \mathsf{Var}(\Phi) \Big). \label{eqn:finalobjective3}
\end{equation}
\clearpage 
\subsection{Proof of Theorem \ref{theorem_ib_ls}: impact of IB on the learning speed}
\label{secn:2d_example_details}

In this section, we present a detailed analysis of 2D case in equation \eqref{eqn:2d_toy_example} leading up to the proof of Theorem \ref{theorem_ib_ls}. For convenience, we will restate the  equation \eqref{eqn:2d_toy_example}. Also, instead of assuming the binary values are from the set $\{0,1\}$ we would shift them to $\{-1,1\}$; we do this purely for making notation clearer.
\begin{equation}
    \begin{split}
    &  Y^e \leftarrow \mathsf{sgn}\Big(X_{\mathsf{inv}}^e\Big), \; \text{where}\; X_{\mathsf{inv}}^e \in\{-1,1\}\; \text{is}\; \mathsf{Bernoulli}\Big(\frac{1}{2}\Big), \\
    & X_{\mathsf{spu}}^e \leftarrow X_{\mathsf{inv}}^eW^e,\; \text{where}\; W^e \in \{-1,1\} \; \text{is} \; \mathsf{Bernoulli}\big(1-p^e\big)\; \text{with selection bias} \; p^{e}>\frac{1}{2},
    \end{split}
    \label{eqn:2d_toy_example_appendix}
\end{equation}

\textbf{Connection between the discrete and the continuous case.} Before discussing the proof of Theorem \ref{theorem_ib_ls}, we provide an explanation as to why can we use the variance penalty as a proxy for the 2D example (eq. \eqref{eqn:2d_toy_example_appendix}), where the random variables are discrete (recall that variance is monotonically related to upper bound on the differential entropy of continuous random variables). We present a variation of equation \eqref{eqn:2d_toy_example_appendix}, where the input feature values are continuous. For each $e\in \mathcal{E}_{tr}$ we have
\begin{equation}
    \begin{split}
      X_{\mathsf{inv}}^{e} \leftarrow C^e + U^{e}, \\ 
      Y^{e}\leftarrow \mathsf{sgn}(X_{\mathsf{inv}}^{e}),
    \end{split}
\end{equation}
where $C^{e}\in \{-1,1\}$ with equal probability for $-1$ and $1$ and $U^{e}$ is a uniform random variable with range $[-\delta, \delta]$ with  $\delta<\frac{1}{2}$. Similarly, with probability $1-p^{e}$, $$X_{\mathsf{spu}}^{e} \leftarrow C^{e} + M^{e},$$ and with probability $p^e$, $$X_{\mathsf{spu}}^{e} \leftarrow -C^{e} + M^{e},$$ where $M^{e}$ is a uniform random variable with range $[-\delta, \delta]$. 

Suppose $\ell$ is exponential loss and the predictor has two dimensions $w_{\mathsf{inv}}$ and $w_{\mathsf{spu}}$. For the above problem description, we write the ERM objective  ($\lambda=0, \gamma=0$ in equation \eqref{eqn:finalobjective}) and we get the following

\begin{equation}
\begin{split}
 &   R_{\mathsf{ERM}}(w_{\mathsf{inv}}, w_{\mathsf{spu}}) = \\ &
 \frac{1}{|\mathcal{E}_{tr}|}\sum_{e\in \mathcal{E}_{tr}}\Big( p^e e^{-(w_{\mathsf{inv}}+w_{\mathsf{spu}})}\mathbb{E}[e^{-w_{\mathsf{inv}}U^e} e^{-w_{\mathsf{spu}}M^e}] +  (1-p^e)e^{-(w_{\mathsf{inv}}-w_{\mathsf{spu}})}  \mathbb{E}[e^{-w_{\mathsf{inv}}U^e} e^{w_{\mathsf{spu}}M^e}] \Big)\\ 
 &\mathbb{E}[e^{-w_{\mathsf{inv}}U^e} e^{-w_{\mathsf{spu}}M^e}] = \mathbb{E}[e^{-w_{\mathsf{inv}}U^e}]\mathbb{E}[e^{-w_{\mathsf{spu}}M^e}] \\
 &  \mathbb{E}[e^{-w_{\mathsf{inv}}U^e}] = \Big(\int_{-\delta}^{\delta} e^{-w_{\mathsf{inv}}u} du \Big)\frac{1}{2 \delta} = \frac{e^{w_{\mathsf{inv}}\delta}- e^{-w_{\mathsf{inv}}\delta}}{2w_{\mathsf{inv}}\delta} \approx  \frac{(1+w_{\mathsf{inv}}\delta) - (1-w_{\mathsf{inv}}\delta)}{2w_{\mathsf{inv}}\delta} =1
\end{split}
\end{equation}
If $\delta$ is small, then we can approximate the loss as if the each of the feature values were discrete and only assumed one of the four possible values in $\{-1,1\}\times \{-1,1\}$.
\begin{equation}
R_{\mathsf{ERM}}(w_{\mathsf{inv}}, w_{\mathsf{spu}})  \approx p e^{-(w_{\mathsf{inv}}+w_{\mathsf{spu}})}+  (1-p)e^{-(w_{\mathsf{inv}}-w_{\mathsf{spu}})} 
\end{equation}
where $p = \frac{1}{|\mathcal{E}_{tr}|}p^e$. On the same lines, we expand the IB-ERM objective as follows

\begin{equation}
R_{\mathsf{IB-ERM}}(w_{\mathsf{inv}}, w_{\mathsf{spu}})  \approx p e^{-(w_{\mathsf{inv}}+w_{\mathsf{spu}})}+  (1-p)e^{-(w_{\mathsf{inv}}-w_{\mathsf{spu}})}+  \gamma[w_{\mathsf{inv}},  w_{\mathsf{spu}}]\Sigma [w_{\mathsf{inv}},  w_{\mathsf{spu}}]^{\mathsf{T}}
\end{equation}

where $\Sigma = \begin{pmatrix}
  1 + \delta^2 & 2p-1\\ 
  2p-1 & 1 +\delta^2
\end{pmatrix} $. Since $\delta$ is small, we approximate $\Sigma$ as $ \begin{pmatrix}
  1 & 2p-1\\ 
  2p-1 & 1 
\end{pmatrix}$.

\textbf{Theorem on impact of information bottleneck.} We would compare the rate of convergence of continuous-time gradient descent for $ R_{\mathsf{IB-ERM}}$ and $R_{\mathsf{ERM}}$.


\begin{theorem}\label{theorem 10_appendix}
Suppose each $e\in \mathcal{E}_{tr}$ follows the 2D case from equation \eqref{eqn:2d_toy_example}. Set $\lambda=0$, $\gamma>0$ in equation \eqref{eqn:finalobjective} to get the IB-ERM objective with $\ell$ as exponential loss.  Continuous-time gradient descent on this IB-ERM  objective achieves $|\frac{w_{\mathsf{spu}}(t)}{w_{\mathsf{inv}}(t)}|\leq \epsilon$ in time less than $\frac{W_{0}(\frac{1}{2\gamma})}{2(1-p)\epsilon}$ ($W_0(\cdot)$ denotes the principal branch of the Lambert $W$ function), while in the same time the ratio for ERM  $|\frac{w_{\mathsf{spu}}(t)}{w_{\mathsf{inv}}(t)}| \geq 
{\ln(\frac{1+2p}{3-2p})}
/{\ln\big(1+\frac{W_{0}(\frac{1}{2\gamma})}{2(1-p)\epsilon}\big)}
$, where $p=\frac{1}{|\mathcal{E}_{tr}|}\sum_{e\in \mathcal{E}_{tr}} p^e$ .
\end{theorem}

\textbf{Proof of Theorem 10.}  We simplify the ERM and the IB-ERM objective in equation \eqref{eqn:finalobjective} for the 2D case. 
$$R_{\mathsf{ERM}}(w_{\mathsf{inv}}, w_{\mathsf{spu}})  = p e^{-(w_{\mathsf{inv}}+w_{\mathsf{spu}})}+  (1-p)e^{-(w_{\mathsf{inv}}-w_{\mathsf{spu}})} $$

$$R_{\mathsf{IB-ERM}}(w_{\mathsf{inv}}, w_{\mathsf{spu}}) = p e^{-(w_{\mathsf{inv}}+w_{\mathsf{spu}})}+  (1-p)e^{-(w_{\mathsf{inv}}-w_{\mathsf{spu}})}+  \gamma[w_{\mathsf{inv}},  w_{\mathsf{spu}}]\Sigma [w_{\mathsf{inv}},  w_{\mathsf{spu}}]^{\mathsf{T}}$$
where $w_{\mathsf{inv}}, w_{\mathsf{spu}} \in \mathbb{R}$ are the weights for invariant and spurious features,  $p=\frac{1}{|\mathcal{E}_{tr}|}\sum_{e\in \mathcal{E}_{tr}}p^e$ $\Sigma$ as $ \begin{pmatrix}
  1 & 2p-1\\ 
  2p-1 & 1 
\end{pmatrix}$.
We first find the equilibrium point of the continuous-time gradient descent for $R_{\mathsf{IB-ERM}}$.
\begin{equation} 
\begin{split}
  &  \frac{\partial  R_{\mathsf{IB-ERM}}(w_{\mathsf{inv}}, w_{\mathsf{spu}})}{\partial w_{\mathsf{inv}}} = -pe^{-(w_{\mathsf{inv}}+w_{\mathsf{spu}})}-(1-p)e^{-(w_{\mathsf{inv}}-w_{\mathsf{spu}})} + 2\gamma(w_{\mathsf{inv}} + (2p-1)w_{\mathsf{spu}}) \\ 
  &  \frac{\partial   R_{\mathsf{IB-ERM}}(w_{\mathsf{inv}}, w_{\mathsf{spu}})}{\partial w_{\mathsf{spu}}} = -pe^{-(w_{\mathsf{inv}}+w_{\mathsf{spu}})}+(1-p)e^{-(w_{\mathsf{inv}}-w_{\mathsf{spu}})} + 2\gamma((2p-1)w_{\mathsf{inv}} + w_{\mathsf{spu}}) 
\end{split}
\end{equation}

\begin{equation}
    \begin{split}
      & \frac{\partial  R_{\mathsf{IB-ERM}}(w_{\mathsf{inv}},w_{\mathsf{spu}})}{\partial w_{\mathsf{inv}}} + \frac{\partial  R_{\mathsf{IB-ERM}}(w_{\mathsf{inv}}, w_{\mathsf{spu}})}{\partial w_{\mathsf{spu}}} =  -2pe^{-(w_{\mathsf{inv}}+w_{\mathsf{spu}})} +4\gamma p(w_{\mathsf{inv}} + w_{\mathsf{spu}}) =0  \\ 
        &\implies\frac{1}{2\gamma}e^{-(w_{\mathsf{inv}}+w_{\mathsf{spu}})} = w_{\mathsf{inv}} + w_{\mathsf{spu}} \\
        & \implies w_{\mathsf{inv}} + w_{\mathsf{spu}} = W_{0}\Big(\frac{1}{2\gamma}\Big)
    \end{split}
\end{equation}

\begin{equation}
    \begin{split}
      & \frac{\partial  R_{\mathsf{IB-ERM}}(w_{\mathsf{inv}}, w_{\mathsf{spu}})}{\partial w_{\mathsf{inv}}} - \frac{\partial   R_{\mathsf{IB-ERM}}(w_{\mathsf{inv}}, w_{\mathsf{spu}})}{\partial w_{\mathsf{spu}}} = -2(1-p)pe^{-(w_{\mathsf{inv}}-w_{\mathsf{spu}})} +4\gamma (1-p)(w_{\mathsf{inv}} - w_{\mathsf{spu}}) =0  \\ 
        &\implies \frac{1}{2\gamma}e^{-(w_{\mathsf{inv}}-w_{\mathsf{spu}})} =w_{\mathsf{inv}} - w_{\mathsf{spu}} \\
        &\implies w_{\mathsf{inv}} - w_{\mathsf{spu}} = W_{0}\Big(\frac{1}{2\gamma}\Big)
    \end{split}
\end{equation}
Therefore, the equilibrium point is $w_{\mathsf{inv}} =W_{0}\big(\frac{1}{2\gamma}\big) $ and $w_{\mathsf{spu}}=0$. Having established that the equilibrium point of the differential equation coincides with ideal predictor, we now analyze the convergence of the trajectory. Let $w_{\mathsf{inv}} + w_{\mathsf{spu}} = x$
and $w_{\mathsf{inv}} - w_{\mathsf{spu}} =y$. 
\begin{equation}
    \frac{\partial x}{\partial t} =  -\Big( \frac{\partial   R_{\mathsf{IB-ERM}}(w_{\mathsf{inv}}, w_{\mathsf{spu}})}{\partial w_{\mathsf{inv}}} + \frac{\partial   R_{\mathsf{IB-ERM}}(w_{\mathsf{inv}}, w_{\mathsf{spu}})}{\partial w_{\mathsf{spu}}}\Big)= 2p(e^{-x} -2\gamma x)
\end{equation}

\begin{equation}
    \frac{\partial y}{\partial t} = 2(1-p)(e^{-y} -2\gamma y)
\end{equation}
Let us call $x^{*} = W_{0}\big(\frac{1}{2\gamma}\big)$; $x^{*}$ is equilibrium point for both $x(t)$ and $y(t)$. Denote $w_{\mathsf{inv}}(t) = \frac{x(t)+y(t)}{2}$ and $w_{\mathsf{spu}}(t) = \frac{x(t)-sy(t)}{2}$. Let us assume that $x(0)=0$ and $y(0)=0$. We would first like to argue that the solution to the above differential equations exist and are unique given the initial conditions $x(0)=0$ and $y(0)=0$. Since $(e^{-x} -2\gamma x)$ is Lipschitz continuous in $x$ on $\mathbb{R}$ the solution to the differential equation exists and is unique for any finite interval $t \in [0, T]$ \citep{simmons2016differential}. With $T$ set to a sufficiently large value, we now show that the solution to the ODE converges to $x^{*}$.

Define an energy function $V(z) = z^2 $ and define $V(x-x^{*}) = (x-x^{*})^2$
\begin{equation}
    \frac{\partial V(x-x^{*})}{\partial t}  = 2(x-x^{*})\frac{\partial x}{\partial t} = 4p(x-x^{*})(e^{-x} -2\gamma x)
\end{equation}
Observe that $\frac{\partial V(x-x^{*})}{\partial t} <0$ for all $x\not=x^{*}$ and $ \frac{\partial V(x-x^{*})}{\partial t} =0$ when $x=x^{*}$. Therefore, from Lyapunov's asymptotic global stability theorem \citep{khalil2009lyapunov} we obtain that  $x(t)$ would converge to $x^{*}$.

Observe that for $x<x^{*}$, $\frac{\partial x}{\partial t }>0$ and moreover $2p(e^{-x} -2\gamma x)$ is a monotonically decreasing function. For all $x<x^{*}-\epsilon$, we can bound the rate at which $x$ increases is bounded below by $2p(e^{-x^{*}+\epsilon} -2\gamma (x^{*}-\epsilon)) \approx 2p (e^{-x^{*}}(1+\epsilon)-2\gamma x^{*} + 2\gamma \epsilon) = 2p\epsilon(e^{-x^{*}} +2\gamma)$. Let us call $\gamma^{*} = \epsilon(e^{-x^{*}} +2\lambda)$. The rate at which $x$ increases is greater than $ 2p\epsilon \gamma^{*}$ and the rate at which $y$ increases is greater than $2(1-p)\epsilon \gamma^{*}$. Thus the time to convergence for $x $ is atmost $\frac{x^{*}}{2p\epsilon} $. Similarly, the time to convergence for $y$ is atmost $\frac{x^{*}}{2(1-p)\epsilon} $. Since $p>\frac{1}{2}$ the time to convergence for $y(t)$ is more than the time taken for the convergence of $x(t)$. 

If $|x(t)-x^{*}|\leq \epsilon$ and $|y(t)-x^{*}|\leq \epsilon$, then $|w_{\mathsf{spu}}(t)| = |\frac{x(t)-y(t)}{2}| =|\frac{x(t)-x^{*} + x^{*}-y(t)}{2}| \leq \frac{|x(t)-x^{*}|}{2} + \frac{|y(t)-x^{*}|}{2} \leq \epsilon$.

If $|x(t)-x^{*}|\leq \epsilon$ and $|y(t)-x^{*}|\leq \epsilon$, then $|w_{\mathsf{inv}}(t)-x^{*}| = |\frac{x(t)+y(t)}{2}-x^{*}| =|\frac{x(t)-x^{*} + y(t)-x^{*}}{2}| \leq \frac{|x(t)-x^{*}|}{2} + \frac{|y(t)-x^{*}|}{2} \leq \epsilon$.

As a result, if $|x(t)-x^{*}|\leq \epsilon$ and $|y(t)-x^{*}|\leq \epsilon$, then 
\begin{equation}
\frac{|w_{\mathsf{spu}}(t)|}{|w_{\mathsf{inv}}(t)|}     \leq \frac{\epsilon}{x^{*}-\epsilon} \approx \frac{\epsilon}{x^{*}}
\end{equation}

Therefore, to get the ratio $\frac{|w_{\mathsf{spu}}(t)|}{|w_{\mathsf{inv}}(t)|} \leq \frac{\epsilon}{x^{*}}$ the time taken is at most $ \frac{x^{*}}{2(1-p)\epsilon}$.

In comparison in the same amount of time the ratio $|\frac{w_{\mathsf{spu}}(t)}{w_{\mathsf{inv}}(t)}|$  achieved by gradient descent on $R_{\mathsf{ERM}}$ is at least $\frac{\ln(\frac{1+2p}{3-2p})}{\ln(1+\frac{x^{*}}{2(1-p)\epsilon})}$. The expression for lower bound on the ratio $|\frac{w_{\mathsf{spu}}(t)}{w_{\mathsf{inv}(t)}}|$ is derived by substituting the time taken, i.e., $ \frac{x^{*}}{2(1-p)\epsilon}$, in the expression for the lower bound derived in Section B.3 in \cite{nagarajan2020understanding}). \hfill $\qedsymbol$

\textbf{Remark on max-margin classifiers.} In the 2D example, the max-margin classifier seems to solve the problem. In general max-margin classifier would not work. In the more general setting, if there is noise in the labels, which is allowed by the SEM in Assumption \ref{assumption 2_new}, and the data is scrambled, which is also the case in Assumption \ref{assumption 2_new}, there is no guarantee that max-margin classifier would not rely on the spurious features.

\clearpage 
\subsection{Illustrating both invariance and information bottleneck acting in conjunction} 
\label{secn:ib_irm_conjunction}
In this section, we present a case to illustrate why the invariance principle and the information bottleneck are needed simultaneously. The model we present follows a DAG that combines the DAGs in Figure \ref{figure:dag_comparison}a) and Figure \ref{figure:dag_comparison}b).

\textbf{Example extending the 2D case from equation \eqref{eqn:2d_toy_example}.} For all the environments $e\in \mathcal{E}_{tr}$

\begin{equation}
    \begin{split}
        Y^e \leftarrow X_{\mathsf{inv}}^{e} \oplus N^e \\ 
        X_{\mathsf{spu}}^{1,e} \leftarrow Y^e \oplus W^e \\ 
        X_{\mathsf{spu}}^{2,e} \leftarrow X_{\mathsf{inv}}^{e} \oplus V^e \\
    \end{split}
\end{equation}

where all the variables in the above SEM are binary $\{0,1\}$ random variables. $N^{e} \sim \mathsf{Bernoulli}(q)$, $V^e \sim \mathsf{Bernoulli}(a)$; the distribution of noise $N^e$ and $V^e$ are the same across the environments.  $W^e \sim \mathsf{Bernoulli}(u^e)$ where $u^e$  is an environment dependent probability. For all the environments $e \in \mathcal{E}_{all}$, we assume that the distribution of $X_{\mathsf{inv}}^{e}$, $N^e$, and $V^e$ does not change. The labelling function to generate $Y^e$ is also the same. The distribution of $X_{\mathsf{spu}}^{1,e}$ can change arbitrarily.  In this example, observe that $\mathbb{E}[Y^e|X^e]$ varies across the training environments. We show the simplification below.
\begin{equation}
    \begin{split}
        & \mathbb{E}[Y^e|X^e] = \mathbb{E}\Big[X_{\mathsf{inv}}^{e} \oplus N^e \big|(X_{\mathsf{inv}}^{e}, X_{\mathsf{spu}}^{1,e}, X_{\mathsf{spu}}^{2,e})\Big] 
        \end{split}
\end{equation}

If $X_{\mathsf{inv}}^{e}=0, X_{\mathsf{spu}}^{e}=0$, then $\mathbb{E}[Y^e|X^e] = \mathbb{P}(N^e=1|X_{\mathsf{inv}}^{e}=0, X_{\mathsf{spu}}^{1,e}=0)$. We show that $\mathbb{P}(N^e=1|X_{\mathsf{inv}}^{e}=0, X_{\mathsf{spu}}^{1,e}=0)$ varies across the environments.

\begin{equation}
\begin{split}
&    \mathbb{P}(N^e=1|X_{\mathsf{inv}}^{e}=0, X_{\mathsf{spu}}^{1,e}=0) = \frac{\mathbb{P}(N^e=1,X_{\mathsf{inv}}^{e}=0, X_{\mathsf{spu}}^{1,e}=0) }{\mathbb{P}(N^e=1,X_{\mathsf{inv}}^{e}=0, X_{\mathsf{spu}}^{1,e}=0) + \mathbb{P}(N^e=0,X_{\mathsf{inv}}^{e}=0, X_{\mathsf{spu}}^{1,e}=0)} \\
 &   = \frac{\mathbb{P}(N^e=1,X_{\mathsf{inv}}^{e}=0) u^e}{\mathbb{P}(N^e=1,X_{\mathsf{inv}}^{e}=0) u^e+ \mathbb{P}(N^e=0,X_{\mathsf{inv}}^{e}=0) (1-u^e)} 
\end{split}
\label{eqn:example_2d_ib_irm_conj_1}
\end{equation}
Note that the above equation \eqref{eqn:example_2d_ib_irm_conj_1} describes the probability computed by the Bayes optimal classifier that relies on input feature dimensions are used.  Observe that the above probability in equation \eqref{eqn:example_2d_ib_irm_conj_1} can only be equal across two environments if $u^e$ was the same. Therefore,  if $|\mathcal{E}_{tr}|\geq 2$ and the probability $u^e$ varies across the environments, then the invariance constraint restrict us from using the identity representation. However, $\mathbb{E}[Y^e|X_{\mathsf{inv}}^{e}, X_{\mathsf{spu}}^{2,e}]$ is invariant and so is  $\mathbb{E}[Y^e|X_{\mathsf{inv}}^{e}]$.  Based on the same arguments that we discussed in the main manuscript, we can show that one can construct classifiers that output probability distributions that minimize cross-entropy (maximize likelihood)  and continue to depend on $X_{\mathsf{spu}}^{2,e}$ as follows
\begin{equation}
\begin{split}
&\hat{\mathbb{P}}(Y^e=1|X_{\mathsf{inv}}^{e}, X_{\mathsf{spu}}^{2,e}) = (1-q) \mathsf{I}\Big(w_{\mathsf{inv}}X_{\mathsf{inv}}^e + w_{\mathsf{spu}}X_{\mathsf{spu}}^e - \frac{(w_{\mathsf{inv}} + w_{\mathsf{spu}})}{2}\Big) +\\& q \Big(1-\mathsf{I}\Big(w_{\mathsf{inv}}X_{\mathsf{inv}}^e + w_{\mathsf{spu}}X_{\mathsf{spu}}^e - \frac{(w_{\mathsf{inv}} + w_{\mathsf{spu}})}{2}\Big)\Big).
\end{split}
\label{eqn:fail_irm}
\end{equation}

If $w_{\mathsf{inv}} > |w_{\mathsf{spu}}|$, then above classifier $\hat{\mathbb{P}}(Y^e=1|X_{\mathsf{inv}}^{e}, X_{\mathsf{spu}}^{2,e})$ matches the true probability distribution conditional on the invariant feature $  \mathbb{P}(Y^e=1|X_{\mathsf{inv}}^{e})$ on all the training environments and it thus forms a valid invariant predictor with representation that focuses on $X_{\mathsf{inv}}^{e}, X_{\mathsf{spu}}^{2,e}$. Since the classifier relies on 
$X_{\mathsf{spu}}^{2,e}$, the classifier fails as the support of spurious features can change.  If we place an entropy constraint, then the representation that focuses only on $X_{\mathsf{inv}}^{e}$ is strictly prefered to one that focuses on both $X_{\mathsf{inv}}^{e}, X_{\mathsf{spu}}^{2,e}$ and continues to achieve the same cross-entropy loss. Thus in this example, IRM fails as its solution space contains classifiers that rely on spurious features but IB-IRM would succeed.  In the above example, ERM and IB-ERM  (with $r^{\mathsf{th}}$ set to match the loss of ERM) will rely on $X_{\mathsf{spu}}^{1,e}$ on top of $X_{\mathsf{inv}}^{e}$ as conditioning on $X_{\mathsf{spu}}^{1,e}$ in addition to $X_{\mathsf{inv}}^{e}$ further reduces the conditional entropy thus reducing the cross-entropy loss.

Let us consider a generalization of the above example. 

\begin{assumption}
\label{assumption 9_new}
 Each environment $e\in \mathcal{E}_{all}$ follows 
\begin{equation}
\begin{split}
  Y^e \leftarrow \mathsf{I}\big(w_{\mathsf{inv}}^{*} \cdot X_{\mathsf{inv}}^e\big) \oplus N^e \\ 
\end{split}
\end{equation}
$N^{e}$ is binary noise, and $X_{\mathsf{inv}}^e$ are binary features. Both $N^e$ and $X_{\mathsf{inv}}^e$ have identical distributions across all the environments $\mathcal{E}_{all}$
\end{assumption}
Divide the spurious features into two parts $X_{\mathsf{spu}}^{e} = (X_{\mathsf{spu}}^{1,e}, X_{\mathsf{spu}}^{2,e})$. 
\begin{assumption}
\label{assumption 10_new}
 Each environment $e\in \mathcal{E}_{tr}$ follows 
\begin{equation}
\begin{split}
  &  X_{\mathsf{spu}}^{1,e} \leftarrow  Y^e \boldsymbol{1} \oplus W^e \\
  &  X_{\mathsf{spu}}^{2,e} \leftarrow  X_{\mathsf{inv}}^e\oplus V^e
 \end{split}
 \end{equation}
 where $\boldsymbol{1}\in \mathbb{R}^{o'}$ is a vector of ones, $W^e\in \mathbb{R}^{o'}$ is a binary $0$-$1$ vector with each component drawn i.i.d. from  $\mathsf{Bernoulli}(u^e)$ vector, $V^e$ is also a binary $0$-$1$ vector with each component drawn i.i.d. from $\mathsf{Bernoulli}(a)$ vector. The distribution of $W^e$ changes across environments and no two training environments have the same $u^e$.  The distribution of $V^e$ is identical across all the training environments.  Also, assume that there are at least two training environments, i.e., $|\mathcal{E}_{tr}|\geq 2$. 
\end{assumption}

\begin{assumption}
\label{assumption 11_new}
$\mathcal{H}_{\Phi}$ is a set of diagonal matrices, where each element in the matrix is $0$ or $1$ ($\mathcal{H}_{\Phi}$  act as matrices that seletct subset of input features).  $\mathcal{H}_{w}$ is set of all probability distributions on $\mathbb{R}^d$. $\ell$ is the cross-entropy loss.
\end{assumption}

 

We use the Shannon entropy formulation of IB-IRM in this case as all the random variables involved are discrete.  Moreover, we carry out entropy minimization for the representation directly and not the predictor. The IB-IRM optimization is given as follows. 

\begin{equation}
\begin{split}
&     \min_{ \Phi\in \mathcal{H}_{\Phi}} \frac{1}{|\mathcal{E}_{tr}|}\sum_{e}H^e(\Phi)  \\  
 & \text{s.t.} \;\; \frac{1}{|\mathcal{E}_{tr}|}\sum_{e}R^{e}\big(w\circ \Phi \big) \leq r^{\mathsf{th}}  \\ 
 & \;\;\;\;\;\;w \in \arg\min_{\tilde{w}\in \mathcal{H}_w} R^e(\tilde{w}\circ \Phi) 
\end{split}
\label{eqn: entropy_risk_min_2_appendix_disc}
\end{equation}


\begin{theorem}
Suppose the data follows Assumption \ref{assumption 9_new}, Assumption \ref{assumption 10_new}. Suppose $\mathcal{H}_w$ and $\mathcal{H}_{\Phi}$ follow Assumption \ref{assumption 11_new}. If invariant features are strictly separable, i.e., Assumption \ref{assumption 8_new} holds, then IRM fails but IB-IRM succeeds.
\end{theorem}

\textbf{Proof of Theorem 11.} We carry out the analysis for different types of representations separately.

Case 1: Consider a representation that selects a subset $\tilde{X}_1^e$ of $(X_{\mathsf{inv}}^e, X_{\mathsf{spu}}^{2,e})$ and a subset $\tilde{X}_2^e$ of $X_{\mathsf{spu}}^{1,e} $. 


\begin{equation}
\begin{split}
&    \mathbb{P}(Y^e=1|\tilde{X}_{1}^{e}=0, \tilde{X}_{2}^{e}=0) = \frac{\mathbb{P}(Y^e=1,\tilde{X}_{1}^{e}=0, \tilde{X}_{2}^{e}=0) }{\mathbb{P}(Y^e=1,\tilde{X}_{1}^{e}=0, \tilde{X}_{2}^{e}=0)+ \mathbb{P}(Y^e=0,\tilde{X}_{1}^{e}=0, \tilde{X}_{2}^{e}=0)} \\
 & = \frac{\mathbb{P}(Y^e=1,\tilde{X}_{1}^{e}=0)(u^e)^{o'}}{\mathbb{P}(Y^e=1,\tilde{X}_{1}^{e}=0)(u^e)^{o'} + \mathbb{P}(Y^e=1,\tilde{X}_{1}^{e}=0)(1-u^e)^{o'}}
\end{split}
\label{ib_irm_comb_eq1}
\end{equation}

Since $\mathbb{P}(Y^e=1|\tilde{X}_{1}^{e}=0, \tilde{X}_{2}^{e}=0)$ is strictly monotonic in $u^e$, this probability cannot be same across two environments. Hence, any $\tilde{X}_{1}^{e}, \tilde{X}_{2}^{e}$ cannot lead to an invariant predictor across the two environments. 



Case 2: Consider a representation that selects a subset $\tilde{X}^e$ of $X_{\mathsf{spu}}^{1,e} $.
 
    \begin{equation}
\begin{split}
&    \mathbb{P}(Y^e=1|\tilde{X}^{e}=0) = \frac{\mathbb{P}(Y^e=1, \tilde{X}^{1,e}=0) }{\mathbb{P}(Y^e=1, \tilde{X}^{1,e}=0) + \mathbb{P}(Y^e=0, \tilde{X}^{1,e}=0)} \\
 & = \frac{\mathbb{P}(Y^e=1)(u^e)^{o'}}{\mathbb{P}(Y^e=1)(u^e)^{o'} + \mathbb{P}(Y^e=0)(1-u^e)^{o'}}
\end{split}
\label{ib_irm_comb_eq2}
\end{equation}

For the above class of representations also, we can use the same argument as the one discussed in Case 1 and show that the above probability cannot be the same across two environments. 


Case 3: At this point, our only option is to consider representations that select subsets of  $(X_{\mathsf{inv}}^e, X_{\mathsf{spu}}^{2,e})$. 
Each subset of $(X_{\mathsf{inv}}^e, X_{\mathsf{spu}}^{2,e})$ satisfies invariance. Among this set all the subsets that lead to lowest cross-entropy are selected by IRM. Among those sets IRM does not exclude the inclusion of spurious covariates $X_{\mathsf{spu}}^{2,e}$. However, when we impose entropy minimization objective, then $X_{\mathsf{spu}}^{2,e}$ will never be selected as entropy can be strictly reduced by not including these covariates in the set without sacrificing invariance or cross-entropy. To explicitly show a construction of the failure of IRM in this case, we can use the same construction as equation \eqref{eqn:fail_irm} but replacing the hyperplane in the indicator function with hyperplane constructed in Lemma \ref{lemma 2}.



\hfill $\qedsymbol$

\clearpage 
\subsection{Related works}
\label{secn:related_works}

\subsubsection{Invariance principles in causality}
The foundations of invariance principles are rooted in the theory of causality \citep{pearl1995causal}. There are several different forms in which the invariance principles or principles similar to it appear in the literature on causality. Modularity condition states that a variable $Y$ is caused by a set of variables  $X_{\mathrm{Pa}(Y)}$if and only if under all interventions other than those on $Y$  the conditional probability $\mathbb{P}(Y|X_{\mathrm{Pa}(Y)})$ remains invariant.  Related and similar notions are \emph{stability} \citep{pearl2009causality}, \emph{autonomy} \citep{scholkopf2012causal}, \emph{invariant causal prediction principle} \citep{peters2016causal,heinze2018invariant}. These principles lead to a powerful insight -- if we model all the environments (train and test) using interventions, then as long as these interventions do not affect the causal mechanism that generates the target variable $Y$, a classifier trained only on the transformation that extracts causal variables  ($\Phi(X)= X_{\mathrm{Pa}(y)}$) to predict $Y$  is invariant under interventions. 

\subsubsection{Invariance principles in OOD generalization}
In recent years, there has been a surge in the works inspired from causality, examples of some notable works are \citep{peters2016causal, arjovsky2019invariant}, which seek to address OOD generalization failures. The invariance principle is at the heart of many of these works. For a better understanding, we divide these works into two categories -- theory and methods, though some works belong to both. 

\textbf{Theory.} In  \cite{rojas2018invariant} it was shown that the predictors trained on the causes are min-max optimal under a large class of distribution shifts modeled by the interventions. These findings were generalized in \cite{koyama2020out}. Given that we know that predictors that focus on the causes are min-max optimal under many distribution shifts, the central question then is -- can we learn these predictors from a finite set of training distributions/environments? \cite{arjovsky2019invariant} showed how to achieve such causal predictors that generalize OOD from a finite set of training environments for linear regression tasks under very general assumptions.  \cite{rosenfeld2020risks} considered linear classification tasks where invariant features were partially informative w.r.t the label and showed that under assumptions of support overlap for invariant and spurious features, it is possible to learn predictors that generalize OOD. In this work, we analyze classification tasks but different from \cite{rosenfeld2020risks} we consider both fully and partially informative features. We showed that support overlap of invariant features is necessarily needed for OOD generalization in classification tasks else OOD generalization, in general, is impossible. On the other hand, we showed that support overlap for spurious features is not necessary but in its absence standard methods such as ERM and IRM can fail.

Recent works \citep{rosenfeld2020risks,rosenfeld2021online,kamath2021does, gulrajani2020search} have also pointed to several limitations of invariance based approaches for addressing OOD generalization failures.   
In \cite{rosenfeld2020risks}, the authors showed that if we use the IRMv1 objective, then for non-linear tasks the solutions from IRMv1 are no better than ERM in generalizing OOD. In \cite{lu2021nonlinear}, the authors present a two-phased approach to addressing the difficulties faced by IRM in the non-linear regime. In the first phase, an identifiable variational autoencoder \citep{khemakhem2020variational} is used to extract the latent representations from the raw input data. In the second phase, causal discovery-based approaches are used to identify the causal parents of the label and then learn predictors based on the causal parents only.  The entire analysis in \cite{lu2021nonlinear} is for the setting when the invariant features are partially informative about the label. Also, the analysis assumes that we have access to side information (possibly in the form of environment index) that can help disentangle all the latent features, i.e., all the latent features are independent conditioned on this side information. Having access to such information, in general, is a strong assumption.   In \cite{kamath2021does}, the authors show that if the label and feature space is finite and if the distribution shifts are captured by analytic functions, then the set of invariant predictors found from two environments exactly capture all the invariant predictors described by the analytic function. While this is a very interesting and important result,  we would like to point out that the distribution shifts captured using analytic functions represent a small family of interventions that are otherwise allowed when learning predictors that focus on causes.  

In this work, we focused on linear SEMs unlike the non-linear SEMs described above. The setting that we considered in this work has three salient features -- a) classification when invariant features are fully informative, b) spurious features are correlated with invariant features, and c) arbitrary shifts are allowed on the spurious feature distribution.  This setting is important as many of the existing failures correspond to this setting. We are the first to give provable OOD generalization guarantees for this setting. Considering non-linear models is a natural next step. On this note, we would like to mention that we believe several of our results can be generalized to the setting when the mapping from the latents to the raw data is piecewise linear.



\textbf{Methods.} 
Following the original works ICP \citep{peters2016causal} and IRM  \citep{arjovsky2019invariant}, there have been several interesting works ---  \citep{teney2020unshuffling, krueger2020out, ahuja2020invariant, jin2020enforcing, chang2020invariant, ahuja2021linear, mahajan2020domain,koyama2020out,muller2020learning,parascandolo2020learning, ahmed2021systematic,robey2021model, Zhang2021CanSS} is an incomplete representative list --- that build new methods inspired from IRM to address the OOD generalization problem.  We would not go into the details of these different works. However, we believe it is important to talk about works that use conditional independence-based criterion to achieve invariance \citep{koyama2020out,ferenc_blog} as those objectives also involve mutual information.  Invariance can be enforced using conditional independence as follows. Suppose the environment is given as a random variable $E$. In this case, if we can learn a representation $\Phi(X)$ such that $Y \perp E | \Phi(X)$, then the predictors learned on $\Phi$ are invariant predictors. This conditional independence constraint is formulated in the form of mutual information-based criterion in \citep{koyama2020out, ferenc_blog}.
In this work, we argue that often in classification tasks, there are many representations $\Phi$ that satisfy $Y \perp E | \Phi(X)$ and we have to learn the one that has the least entropy or otherwise OOD generalization is not possible. 


\subsubsection{Theory of domain adaptation and domain generalization}

In the previous section, we discussed works that were directly based on causality/invariance or inspired from it. We now briefly review other relevant works on domain adaptation and domain generalization that are not based on invariance principle from causality.  Starting with the seminal works \citep{ben2007analysis,ben2010theory}, there have been many other interesting works in the area of domain adaptation and domain generalization. \citep{muandet2013domain, zhao2019learning, albuquerque2019adversarial, piratla2020efficient, matsuura2020domain, deng2020representation, pagnoni2018pac, greenfeld2020robust, garg2021learn} is an incomplete representative list of works that build the theory of domain adaptation and generalization and construct new methods based on it. We recommend the reader to \cite{redko2019advances} for further references. 

In the case of domain adaptation, many of these works develop bounds on the loss over the target domain using train data and unlabeled target data. In the case of domain generalization, these works develop bounds on the loss over the target domains using training data from multiple domains. Other works  \citep{ben2012hardness,david2010impossibility}  analyze the minimal conditions under which domain adaptation is possible. In \cite{david2010impossibility}, the authors showed that the two most common assumptions,  a) covariate shifts, and b) the presence of a classifier that achieves close to ideal performance simultaneously in train and test domains, are not sufficient for guaranteed domain adaptation. In this work, we established the necessary and sufficient conditions for domain generalization in linear classification tasks. We showed that under a) covariate shift assumption (SEM in Assumption \ref{assumption 3_new} satisfies the covariate shift), and b) the presence of a common labelling function across all the domains (a much stronger condition than assuming the existence of a classifier that achieves low error across the train and test domains), domain generalization in linear classification is impossible. We showed that adding the requirement that the invariant features satisfy support overlap is both necessary and sufficient (our approach IB-IRM succeeds while IRM and ERM fail) in many cases to guarantee domain generalization.

There has been a long line of research focused on learning domain invariant feature representations \citep{ganin2016domain,li2018deep, zhao2020domain}. In these works, the common assumption is that the there exist highly predictive representations whose distributions $\mathbb{P}(\Phi(X^e))$(or distributions conditional on the labels $\mathbb{P}(\Phi(X^e)|Y^e)$) do not change across environments. Note that this is a much stronger assumption than the one typically made in works based on invariance principle \citep{arjovsky2019invariant}, where the labelling function ($\mathbb{P}(Y^e|\Phi(X^e))$ does not change.  For a detailed analysis of why the assumptions made in these works are too strong and can often fail refer to \cite{arjovsky2019invariant, zhao2019learning}.



\subsubsection{Other works on OOD generalization} In \cite{nagarajan2020understanding} the authors explained why ERM based models trained with gradient descent based approaches fail to generalize OOD  in terms of two failure modes -- a) gradient descent during training early on relies on shortcut features, b) overparametrized models exhibit geometric biases that cause the models to rely on spurious features. We now describe the line of work based on domain adaptation. For failure mode described in a), we showed in Theorem \ref{theorem_ib_ls} how information bottleneck penalty can help.
\cite{sagawa2019distributionally} studied how overparametrized models can exacerbate the impact of selection biases, \cite{xie2021innout} studied the role of auxilliary information  and how it can help OOD generalization.

\subsubsection{Information bottleneck penalties and impact on generalization}
Information bottleneck principle \citep{tishby2000information} has been used to explain the success of deep learning models; the principle has also been used to build regularizers that can help build models that achieve better in-distribution generalization. We refer the reader to \cite{kirsch2020unpacking}, which presents an excellent summary of the existing works on information bottleneck in deep learning. \cite{kirsch2020unpacking} also present a unified framework to view many of the information bottleneck objectives in the literature such as the deterministic information bottleneck \citep{strouse2017deterministic} and the standard information bottleneck. Other works \citep{alemi2016deep,arpit2019entropy} have argued for how information bottleneck can help achieve robustness to adversarial examples, and also to OOD generalization failures. In \cite{arpit2019entropy}, the authors argued that 
information bottleneck constraints help filter out features that are less correlated with the label. However, the principle of invariance argues for selecting the invariant features even if they have small but invariant correlation with the label over features that maybe strongly correlated but have a varying correlation.  As we showed, considering both the principles of invariance and information bottleneck in conjunction is important to achieve OOD generalization (eq. \eqref{eqn1: min_max_ood}) in a wide range of settings -- when the invariant features are fully informative about the label and also when they are partially informative about the label.

\end{document}